\documentclass{article}

\usepackage[final]{corl_2021} 

\usepackage{amsmath}
\usepackage{amsfonts}
\usepackage{bbm}
\usepackage{graphicx}
\usepackage{subcaption}

\usepackage[numbers]{natbib}
\usepackage{multicol}
\usepackage{makecell}

\usepackage{marginnote}
\usepackage{paralist}
\usepackage{wrapfig}
\usepackage{xspace}
\usepackage{setspace}

\usepackage{enumitem}

\usepackage{floatrow}
\newfloatcommand{capbtabbox}{table}[][\FBwidth]
\usepackage{caption}

\newcommand{\X}{\mathcal{X}}
\newcommand{\reals}{{\mathbb R}}                    
\newcommand{\nnreals}{\mathbb R^{\geq 0}}

\newcommand\TBDRAFT[1]{\color{black}{#1}}
\newcommand\TMPDRAFT[1]{\textcolor{black}{#1}}
\newcommand\DRAFT[1]{\textcolor{black}{#1}}

\newcommand{\densitygain}{density concentration function\xspace}

\newcommand{\suppref}[1]{Appendix #1}
\newcommand\CAPRBT{. The gray dots are sampled points and blue / green / red / colored regions are reachability results}

\title{Learning Density Distribution of Reachable States for Autonomous Systems}

\author{
  Yue Meng \\
  MIT\\
  United States\\
  \texttt{mengyue@mit.edu} \\
  
  \And Dawei Sun \\
  UIUC\\
  United States\\
  \texttt{daweis2@illinois.edu} \\
  
  \And Zeng Qiu \\
  Ford\\
  United States\\
  \texttt{cqiu1@ford.com}\\
  
  \And Md Tawhid Bin Waez \\
  Ford\\
  United States\\
  \texttt{mwaez@ford.com}\\
  
  \And Chuchu Fan \\
  MIT\\
  United States\\
  \texttt{chuchu@mit.edu} \\
  
}

\begin{document}
\maketitle
\begin{abstract}
    State density distribution, in contrast to worst-case reachability, can be leveraged for safety-related problems to better quantify the likelihood of the risk for potentially hazardous situations. In this work, we propose a data-driven method to compute the density distribution of reachable states for nonlinear and even black-box systems. Our semi-supervised approach learns system dynamics and the state density jointly from trajectory data, guided by the fact that the state density evolution follows the Liouville partial  differential equation. With the help of neural network reachability tools, our approach can estimate the set of all possible future states as well as their density. Moreover, we could perform online safety verification with probability ranges for unsafe behaviors to occur. We use an extensive set of experiments to show that our learned solution can produce a much more accurate estimate on density distribution, and can quantify risks \DRAFT{less conservatively} and flexibly comparing with worst-case analysis.
\end{abstract}
\keywords{Reachability Density Distribution, Learning Density Distribution, Liouville Theorem} 
\section{Introduction}
	\begin{wrapfigure}{r}{0.5\textwidth}
	\vspace{-1.0em}
	\begingroup
\setlength{\columnsep}{0pt}
\setlength{\intextsep}{0pt}
\begin{subfigure}[b]{0.32\textwidth}
\includegraphics[width=0.92\textwidth]{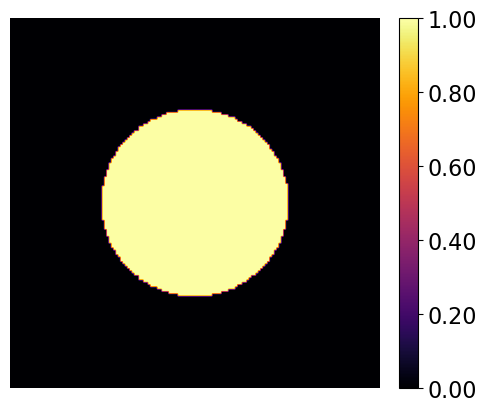}
\caption{\vspace{-0.5em}$t=0s$.}
\end{subfigure}
\begin{subfigure}[b]{0.32\textwidth}
\includegraphics[width=0.94\textwidth]{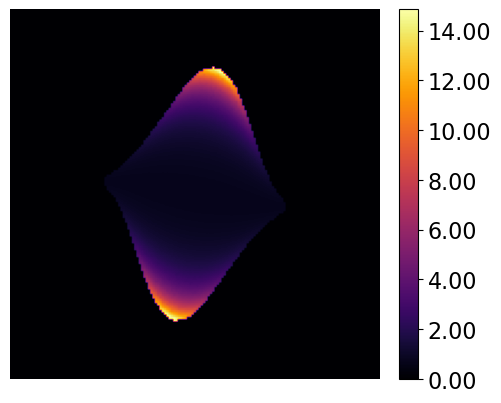}
\caption{\vspace{-0.5em}$t=0.35s$.}
\end{subfigure}
\begin{subfigure}[b]{0.32\textwidth}
\includegraphics[width=0.99\textwidth]{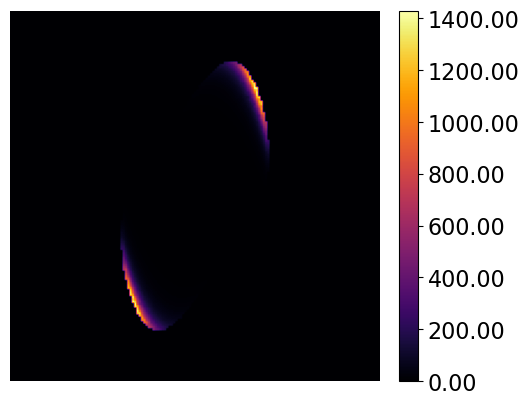}
\caption{\vspace{-0.5em}$t=0.80s$.}
\end{subfigure}
\centering
\vspace{-0.5em}
\caption{\small Reachability density of Van der Pol. \vspace{-1.0em}}
\label{fig:van-der-pol}
\endgroup
\end{wrapfigure}
Reachability analysis has been a central topic and the key for the verification of safety-critical autonomous systems. The majority of existing reachability approaches either compute worst-case reachable sets~\citep{chen2018hamilton,devonport2020data,chen2013flow,tran2020nnv,bansal2020deepreach,liu2019algorithms}, or use Monte Carlo simulation to estimate the reachable states with probabilistic guarantees~\citep{devonport2020estimating,liebenwein2018sampling,lew2020sampling,devonport2020data}. There are several obvious disadvantages of such methods. 1). Worst-case methods, \DRAFT{especially those for nonlinear systems}, often over-approximate the reachable sets and produce very conservative results (due to convex set representations~\citep{kurzhanski2000ellipsoidal,girard2005reachability,duggirala2016parsimonious,meyer2019tira}, wrap effects~\citep{bak2014reducing,althoff2008reachability}, low-order approximation~\citep{chen2013flow,cyranka2017lagrangian}, etc.), not to say they are usually very computationally expensive (also known as the curse of dimensionality). 2). Existing methods do not care about reachable state concentration and produce the same reachable state estimations for different distributions of initial states if those distributions share the same support~\footnote{The probabilistic guarantees of the sampling-based methods do rely on the form of the initial states distribution. However, the final reachable sets estimate is the same for different distributions with the same support.}. That is, current approaches do not compute which states are more likely to be reached. 
For example, in Fig.~\ref{fig:van-der-pol}, the state density distribution of a Van der Pol oscillator evolves over time and concentrates on certain states (the highlighted part in the figure), even if the initial states are uniformly distributed. \DRAFT{If one uses an existing worst-case reachability algorithm, most likely the results in Fig.~\ref{fig:van-der-pol} (b)(c) will show almost the entire black region inside the highlighted part is reachable, as those methods often use convex sets to represent the reachable sets.}


In this paper, we aim to tackle the above problems and propose a learning-based method that can compute the density distribution of the reachable states from given initial distributions. The state density is much more powerful than worst-case reachability and can better quantify risks.
Our proposed method is based on the Liouville theorem~\cite{ehrendorfer1994liouville,nakamura2019density,chen2020optimal}, which is from classical Hamiltonian mechanics and asserts that the state density distribution function (which is the measurement of state concentration) is constant along the trajectories of the system. Given an autonomous system $\dot{x}=f(x)$ that is locally Lipschitz continuous, the evolution of the state density $\rho(x,t)$ (i.e. the density of state $x$ at time $t$) is characterized by a Liouville partial differential equation (PDE). We learn the state density as a neural network (NN) while respecting the laws of physics described by the nonlinear Liouville PDE, in a way that is similar to the physics-informed NN~\cite{raissi2019physics}. 

    Furthermore, we make two major improvements to make the learned density NN suitable for verification. 1). Instead of learning $\rho$ for a fixed initial distribution $\rho(x_0,0)$, we learn the {\densitygain} which specifies the multiplicative change of density from any $\rho(x_0,0)$. 2). We use a single NN to jointly learn both the reachable state $\Phi(x_0,t)$ and its corresponding density. Moreover, we use Reachable Polyhedral Marching (RPM)~\citep{vincent2020reachable}---an exact ReLU NN reachability tool---to parse our learned NN as linear mappings from input polyhedrons to output polyhedrons. Using such parsed polyhedrons, we can perform online forward and backward reachability analysis and get the range of density bounds $[\rho_{\min}, \rho_{\max}]$ for each output polyhedron. Together, our method can perform  online safety verification by computing the probability of safety (instead of a single yes or no answer) under various initial conditions, \TMPDRAFT{even with unknown system dynamics $f$} (i.e. a black-box system where we only have access to a simulator of $x(t)$). In this way, it also has the potential to collect environmental data in run-time and update its distribution for online safety verification.  


We conduct experiments on 10 different benchmarks covering systems from low dimension academic examples to high dimension black-box simulators equipped with either hand-crafted or NN controllers. Surprisingly, without using any ground-truth density data in the learning process, our approach can achieve up to 99.89\% of error reduction in KL divergence with respect to the ground-truth value, \DRAFT{when compared to sampling-based methods like kernel density estimation and Gaussian Processes}. Moreover, our learned \densitygain can also be used for reachability distribution analysis. We show that in several systems, more than 90\% of the states can actually just reside in a small region (less than 10\% of the volume of the convex hull for those states) in the state space, which also points out the conservativeness of the worst-case reachability analysis methods in terms of quantifying risks. We also show that different initial distributions can lead to a drastic change in the safety probability, which can help in cases when unsafe is inevitable but can be designed to happen with a very low probability.

Our major contributions are: (1) \DRAFT{we are the first to provide an explicit probability density function for reachable states of dynamical systems characterized as ordinary differential equations; this density function can be used for online safety analysis}, (2) we propose the first data-driven method to learn the state density evolution and give accurate state density estimation \DRAFT{for arbitrary (bounded support) initial density conditions without retraining or fine-tuning,} and (3) we use a variety of examples to show the necessity to perform reachability distribution analysis instead of pure worst-case reachability, to flexibly and \DRAFT{less conservatively} quantify  safety risks.

\paragraph{Related work}









Reachability analysis, especially worst-case reachability using sampling-based methods or set propagation, has been studied for decades. The literature on reachability has been extensively studied in many surveys~\cite{chen2018hamilton,liu2019algorithms,arch,agha2018survey}.  Here we only discuss a few closely related works. 

Hamilton Jacobian PDE has been used to derive the exact reachable sets in~\citep{mitchell2005time,chen2018hamilton,bansal2020deepreach}. However, the HJ-PDE does not provide the density information.
Many data-driven approaches can compute probabilistic reachable sets using scenario optimization~\citep{devonport2020estimating,xue2020pac}, convex shapes ~\citep{liebenwein2018sampling,lew2020sampling,berndt2021data}, support vector machines~\citep{allen2014machine,rasmussen2017approximation}, kernel embedding~\citep{thorpe2020data}, active learning~\citep{chakrabarty2020active}, and Gaussian process~\citep{devonport2020data}. However, the probabilistic guarantees they provide are usually in the form that  $\mathbf{Prob}(x(t)\in \mathtt{estimated~set})>1-\epsilon$ with enough samples, instead of the state density distribution. \DRAFT{\citep{fridovich2020confidence} estimates human state distribution using a probabilistic model but requires state discretization. \citep{majumdar2014convex} uses the Liouville equation to maximize the backward reachable set for only polynomial system dynamics}.  \DRAFT{In \citep{abate2007probabilistic} the authors compute the stochastic reachability by discretizing stochastic hybrid systems to Markov Chains (MC), then perform probabilistic analysis on the discretized MC. The closed-form expression of the probability requires integrals over the whole state space hence is computation-heavy and cannot be used for online safety check.} The closest to ours is~\citep{matavalam2020data} where Perron-Frobenius and Koopman operators are learned from samples of trajectories. Then, the learned operators can be used to transform the moments of the distribution over time. The distribution at time $t$ is then recovered from the transformed moments. However, as they use moments up to a specific order to represent a distribution, even if the learned operators are perfect, the estimation error of the distribution might not be zero. Also, such a moment-based method is hard to scale to large-dimensional systems.
Recently, there is also a growing interest in studying the (worst-case) reachability of NN~\citep{katz2017reluplex,katz2019marabou,xiang2018output,yang2020reachability,vincent2020reachable} or systems with NN controllers~\citep{ivanov2019verisig,tran2020nnv,fan2020reachnn,hu2020reach,everett2021efficient}. 
In this paper, we use~\citep{vincent2020reachable} to parse our learned NN as a set of linear mappings between polyhedrons for online reachability computation, but this step can be replaced with other NN reachability tools.

To measure the probability distribution in the reachable sets, the most na\"{i}ve approach is to use histograms or kernel density estimation~\citep{chen2020optimal}, similar to the Monte-Carlo method used in dispersion analysis~\citep{spencer1996mars}. But this could lead to poor accuracy and computational scalability~\citep{niederreiter1992random}. Another approach propagates the uncertainty by approximating the solution of the probability density function (PDF) transport equation \citep{pantano2007least}, which could still be time-consuming due to the optimization process performed at each time step. 
Our approach finds the PDF transport equation by solving the Liouville PDE \footnote{In the absence of process noise, this PDF transport equation reduces to stochastic Liouville equation~\citep{ehrendorfer1994liouville}} using NN. Similar ideas have been explored in~\citep{nakamura2019density}. However, the density NN in~\citep{nakamura2019density} was learned solely for a fixed initial states distribution and therefore, cannot be used for online prediction. Instead, we jointly learn the reachable states and density changes, and can perform online reachability density computation for any initial states distribution.


The idea of using deep learning to solve PDEs can be traced back to the 1990s \citep{lee1990neural,lagaris1998artificial,uchiyama1993solving}, where the solutions of the PDE on a priori fixed mesh are approximated by NN. Recently, there is a growing interest in the related sub-fields including: mesh-free methods in strong form~\citep{raissi2019physics,sirignano2018dgm,berg2018unified} and weak form~\citep{weinan2018deep,he2018relu}, solving high-dimension PDEs via BSDE method~\citep{han2020algorithms}, solving parametric PDE~\citep{khoo2021solving,kutyniok2021theoretical} and stochastic differential equations~\citep{zhang2019quantifying,yang2018physics}, learning differential operators\citep{li2020fourier,long2018pde,lu2019deeponet}, and developing more advanced toolboxes \citep{koryagin2019pydens,lu2021deepxde,chen2020neurodiffeq}. Our idea for solving Liouville PDE along trajectories is similar to~\citep{han2020algorithms} but without the stochastic term.

\section{Preliminaries}
\label{sec:prelim}
We denote by $\reals$, $\nnreals$, $\reals^n$, $\reals^{n \times n}$ the sets of real numbers, non-negative real numbers, $n$-dimensional real vectors, and $n\times n$ real matrices. We consider autonomous dynamical systems of the form
$\dot{x}(t) = f(x(t))$,
where for all $t \in \reals$, $x(t) \in \X \subseteq \reals^d$ is the state and $\X$ is a compact set that represents the state space. We assume that $f: \reals^d \mapsto \reals^d$ is locally Lipschitz continuous. The solution of the above differential equation exists and is unique for a given initial condition $x_0$. We define the flow map $\Phi: \X \times \reals \mapsto \X$ such that $\Phi(x_0, t)$ (also written as $x(t)$ for brevity) is the state at time $t$ starting from $x_0$ at time $0$. Note that system parameters can be easily incorporated as additional state variables with time derivative to be $0$.

We analyze the evolution of the dynamical system by equipping it with a density function $\rho:\mathcal{X} \times \reals \to \nnreals$, which measures how states distribute in the state space at a specific time instant. A larger density $\rho(x,t)$ means the state is more likely to reside around $x$ at time $t$, and vice versa. The density function is completely determined by the underlying dynamics (i.e., function $f$) and the initial density map $\rho_0: \X \mapsto \nnreals$. Specifically, given a $\rho_0$, the density function $\rho$ solves the following boundary value problem of the Liouville PDE~\citep{ehrendorfer1994liouville}:
\begin{equation}
    \frac{\partial \rho}{\partial t}+\nabla \cdot (\rho \cdot f)=0,~~~
    \rho(x,0)=\rho_0(x),
\label{eq:liouville-pde}
\end{equation}
where $\nabla\cdot (\rho\cdot f)=\sum_{i=1}^d\frac{\partial [\rho\cdot f]_i}{\partial [x]_i}$ is the divergence for the vector field $\rho\cdot f$, and $[\cdot]_i$ takes the $i$-th coordinate of a vector.
\footnote{A general form of Liouville PDE can have a non-zero term on the right-hand side indicating how many (new) states appear or exit from the system during the run time~\cite{chen2020optimal}. In all the systems we discuss here, there is no state entering (other than the initial states) or leaving the system, so the right-hand side of Eq.~\eqref{eq:liouville-pde} is zero. The total density of the systems we consider is invariant over time.} Intuitively, as shown in \citep{ehrendorfer1994liouville}, Liouville PDE is analogous to the mass conservation in fluid mechanics, where the change of density $\frac{\partial \rho}{\partial t}$ at one point is balanced by the total flux traversing the surface of a small volume surrounding that point.
It is hard to solve the Liouville PDE for a closed from of the density function $\rho$. However, it is relatively easy to evaluate the density along a trajectory of the system. To do that, we first convert the PDE into an ODE as follows. Considering a trajectory $\Phi(x_0, t)$, the density along this trajectory is an univariate function of $t$, i.e., $\rho(t) = \rho(\Phi(x_0, t), t)$. If we consider the augmented system with states $[x, \rho]$, from Eq.~\eqref{eq:liouville-pde} we can easily get the dynamics of the augmented system~\citep{chen2020optimal}:
\begin{equation}
    \begin{bmatrix}\dot{x}\\\dot{\rho}\end{bmatrix}=\begin{bmatrix} f(x)\\  - \nabla \cdot f(x)\rho \end{bmatrix}.
    \label{liouville-ode}
\end{equation}
Therefore, to compute the density at an arbitrary point $x_T \in \X$ at time $T$, one can simply proceed as follows: First, find the initial state $x_0=\Phi(x_T, -T)$ using the inverse dynamics $-f$. Then, solve the Eq.~\eqref{liouville-ode} with initial condition $[x_0, \rho_0(x_0)]$. The solution at time $T$ just gives the desired density value. However, such a procedure only gives the density at a single point, therefore, cannot be used in reachability analysis.
Instead, we need to compute the solution of Eq.~\eqref{liouville-ode} for a set of initial conditions and use that to compute the reachable sets.
To achieve this, we use an NN  with ReLU (Rectified Linear Unit) activation functions~\citep{nair2010rectified} to jointly approximate the flow map $\Phi$ and the {\it \densitygain}, as will be shown in the next section.

\section{Density learning and online reachability density computation}
Let us take a closer look at  Eq.~\eqref{liouville-ode}. For an initial condition $[x_0, \rho_0(x_0)]$, the closed form of the solution $\rho$ is
$
\rho(\Phi(x_0, t), t) = \rho_0(x_0) \exp{\left(-\int_{0}^{t} \nabla \cdot f(\Phi(x_0, \tau)) d\tau\right)}.$
Interestingly, the solution of $\rho$ is linear in the initial condition $\rho_0$. The gained part denoted by $G(x_0, t) := \exp{\left(-\int_{0}^{t} \nabla \cdot f(\Phi(x_0, \tau)) d\tau\right)}$ is a function of the initial state $x_0$ and time $t$, but is independent of $\rho_0$. This mapping $G$ is completely determined by the underlying dynamics and we call it the {\it \densitygain}. With $G$ in hand, the density at an arbitrary time and state can be quickly computed from any $\rho_0$. However, $G$ is obviously hard to compute. Therefore, we use NN to approximate the {\it \densitygain}. In addition to $G$, we also use NN to learn the flow map $\Phi$, which will be a necessity for computing the reachable set distribution shown in Sec.~\ref{sec:rpm}.

\subsection{System dynamics and density learning framework}
\label{sec:nn}
Let the parameterized versions of the flow map and the {\it \densitygain} be $\Phi_{\omega}$ and $G_{\theta}$ respectively, where $\omega$ and $\theta$ are parameters. To train the neural network, we construct a dataset by randomly sampling $N$ trajectories of the system: $\mathcal{D}_{train}=\{\xi_i\}_{i=0}^{N-1}$ in $T$ time steps (with time interval $\Delta t$): $\xi_i=\{(x^i_{0},0),(x_1^i,\Delta t),...,(x_{T-1}^i,(T-1)\Delta t)\}$ where $x_{j}^i = \Phi(x_0^i, j\Delta t)$. Then, the goal of the learning is to find parameters $\omega$ and $\theta$ satisfying
\begin{equation}
\begin{cases}
    \Phi_{\omega}(x_0^i,k\Delta t) - x_k^i = 0 ,\,\forall i,k, \\
    \frac{\partial G_{\theta}(x_0^i, k\Delta t)}{\partial t} + G_{\theta}(x_0^i, k \Delta t) \cdot (\nabla \cdot f(x_k^i)) = 0 ,\,\forall i,k,
\end{cases}
\label{eq:optimization}
\end{equation}
where the first constraint is for the flow map estimation, and the second constraint enforces the Liouville equation for all the data points.

As for the implementation, we model $\Phi_{\omega}$ and  $G_{\theta}$ jointly as a fully-connected neural network $\text{NN}(\cdot,\cdot)$ with ReLU activations. To ensure numerical stability as $G$ is an exponential function, we add a nonlinear transform from the NN output to the {\it \densitygain}:
\begin{equation}
\begin{cases}
    G_{\theta}(x_0, t)=\exp({t \cdot \text{NN}_{[0:1]}(x_0, t)})=\exp(t\cdot z(x_0,t)),\\
    \Phi_{\omega}(x_0, t) = \text{NN}_{[1:n+1]}(x_0, t),
    \end{cases}
    \label{eq:transform}
\end{equation}
where $\text{NN}_{[i:j+1]}$ is to choose the $i,i+1,...,j$-th dimensions from the output of the NN, and $z$ is the intermediate density estimation from the NN. In this way, we guarantee that the \DRAFT{\it \densitygain} at $t=0$ is always 1. We optimize our NN via back propagation \DRAFT{with the loss function:}
\begin{equation}
\begin{aligned}
    &\mathcal{L}=\lambda \cdot \sum\limits_{i,k}\left[\Phi_{\omega}(x_0^i, k\Delta t)-x_k^i\right]^2 +\sum\limits_{i,k}\left[\dot{G}_{\theta}(x_0^i,k\Delta t) + G_{\theta}(x_0^i,k\Delta t) \left(\nabla \cdot f(x_k^i)\right) \right]^2,
\end{aligned}
\end{equation}
where the first term denotes the state estimator square error~\cite{chai2014root}, the second term indicates how far (in the sense of L2-norm) the solution deviates from the Liouville Equation, and $\lambda$ balances these two loss terms. We approximate the time derivative of the {\it \densitygain} by $\dot{G}_{\theta}(x_0^i, k\Delta t)=\left[G_{\theta}(x_0^i,(k+1)\Delta t)-G_{\theta}(x_0^i,k\Delta t)\right]/\Delta t$. Our method can also work for black-box systems if we approximate $\nabla \cdot f(\cdot)$ numerically. With some tools from statistical learning theory, we can show that with a large enough number $N$ of samples, the learned flow map and the {\it \densitygain} can be arbitrarily accurate. A formal proof is shown in \DRAFT{\suppref{A}}.

\subsection{System reachable set distribution computation via NN Reachability Analaysis}
\label{sec:rpm}

In the last section, we have learned an NN that can estimate the density at a single point given the initial density. In this section, we will boost this single-point estimation to set-based estimation by analyzing the reachability of the learned NN. To compute the set of all reachable states and the corresponding density from a set of initial conditions, we use the Reachable Polyhedron Marching (RPM) method~\citep{vincent2020reachable} to further process the learned NN for $G$ and $\Phi$.
RPM is a polyhedron-based approach for exact reachability analysis for ReLU NNs.
It partitions the input space into polyhedral cells so that in each cell the ReLU activation map does not change and the NN becomes a fixed affine mapping. With the cells and the corresponding affine mapping on each cell, the exact reachable set for a set of input can be quickly evaluated.
Backward reachability analysis can also be performed by computing the intersection of the pre-image of a query output set with the input polyhedron cells.

\noindent{\textbf{System forward reachable set with density.}} Recall that the input of the NN in Sec.~\ref{sec:nn} consists of the initial state $x_0$ and time $t$. For the simplicity of comparison with other methods, we only estimate the reachable set and the density at given fixed time instances $t$ and fix the last element of the input of the NN. Thus for each time step t, the input polyhedral cells generated from the RPM will be a set of linear inequality constraints on $x$, and those input cells together with the set of the affine mappings and output polyhedral cells can be represented as: $\{(A_k, b_k, C_k, d_k, E_k, f_k)\}_{k=1}^N$, where in each input cell $H_k := \{v \in \reals^d | A_k v \leq b_k\}$, the NN becomes an affine mapping $y = C_k v + d_k$, and thus the image of the input cell is also a polyhedron $M_k=\{y \in \reals^{d+1} | E_k y\leq f_k\}$.

Recall that the first dimension of our NN output estimates the {\it \densitygain} $z$, and the rest dimensions estimate the state $x$, thus the output cell can be written as $M_k=\{(z, x) | E_k [z,\,x]^T \leq f_k\}$. By projecting it to the state space, we get the reachable set of the system, i.e., $R_k^o := \{x \in \X | (z, x) \in M_k\}$. Then, in each cell $M_k$, we evaluate the lower and upper bounds of $z$, and denote them by $z_{k,min}$ and $z_{k,max}$.
The density bound for cell $M_k$ is then computed as
\begin{equation}
\begin{cases}
    \rho_{k,min} = \rho_0(\bar{x}_k) \cdot \exp({t \cdot z_{k,min}}),\\
    \rho_{k,max} = \rho_0(\bar{x}_k) \cdot \exp({t \cdot z_{k,max}}),
\end{cases}
    \label{eq:fwd-dens-2}
\end{equation}
where $\bar{x}_k$ is the center of $H_k$.
Finally the system forward reachable set is a union of projected output polyhedral cells: $\mathop{\bigcup}\limits_{k=1}^{N} R^{o}_k$ where each cell $R^{o}_k$ is associated with a density bound $[\rho_{k, min},\rho_{k,max}]$.

\noindent{\textbf{System reachable set probability computation.}}
%
Given an initial state distribution, we want to figure out the probability distribution of the system forward reachable sets, as well as the probability for the states land into a query set (e.g., the query set could be the unsafe region).


For an arbitrary initial probability density function (whose support is bounded), we can apply RPM to partition its support into cells\footnote{We can always further divide those input cells to make the bound tighter/ more precise, while still guaranteeing the Neural Network on each cell can be seen as an affine transformation.} as in the last section. Finally, we obtain the reachable sets with bounded state densities $\{(H_k, \rho_{k}^{min}, \rho_k^{max})\}_{k=1}^{N}$ \DRAFT{and the probability bound in each cell is:}
\begin{equation}
    \begin{cases}
      P_{k}^{min}= \text{Vol}(H_{k}) \rho_k^{min}, \\
      P_{k}^{max}= \text{Vol}(H_{k}) \rho_k^{max},
    \end{cases}
\end{equation}
where $\text{Vol}(\cdot)$ computes the volume for a polyhedron. By computing for all input cells $\{H_k\}_{k=1}^N$, we can derive the system forward reachable set and the corresponding probability bound as $\left\{(H_{k}, P_{k}^{min}, P_{k}^{max})\right\}_{k=1:N}$.
%
%
The backward reachable set probability can be computed in a similar fashion, by checking the intersection between the query output region and the output cells derived by RPM, computing each intersection's probability range by its volume and density bound and finally aggregating the probability of all intersections. Detailed computation is shown in \DRAFT{\suppref{B}}. \footnote{\DRAFT{The RPM method cannot handle systems with  higher than 4 dimensional state space in our experiments. But the technique discussed in Sec.~\ref{sec:rpm} can work with any set-based NN reachability tools with slight modification based on the set presentation of the tool. Developing a better NN reachability tool that can scale to higher dimensional systems is another topic and is out of the scope of our paper.}}

\vspace{-0.3em}
\section{Experimental evaluation in simulation}
\vspace{-0.5em}
Here we show the benefits of learning {\it\densitygain} from Liouville PDE in state density and reachable set distribution estimation. More experiments are provided in \DRAFT{\suppref{C$\sim$H}}.

\begin{table}[!tbp]
\RawFloats
\begin{minipage}{\textwidth}
\begin{minipage}[b]{0.38\textwidth}
    \centering
    \resizebox{0.99\textwidth}{!}{%
    \renewcommand{\arraystretch}{1.2}
   \begin{tabular}{|c |c c |} 
 \hline
  System& Dim. & Control\\ 
 \hline
Van der Pol Oscillator (vdp) & 2 & - \\
Double integrator (dint)~\citep{everett2021efficient}  & 2 & NN  \\
Kraichnan-Orszag system (kop)~\citep{orszag1967dynamical} & 3 & - \\
Inverted pendulum (pend)~\citep{chang2020neural} & 4  & LQR  \\ 
Ground robot navigation (rpbot)& 4 & NN \\
FACTEST car tracking system (car)~\citep{fan2020fast} & 5 & Tracking \\
Quadrotor control system (quad)~\citep{everett2021efficient} & 6  & NN \\
Adaptive cruise control system (acc)~\citep{tran2020nnv}& 7 & NN \\
F-16  Ground collision avoidance (gcas)~\citep{heidlauf2018verification} & 13 & Hybrid \\
8-Car platoon system (toon)~\citep{zhu2019inductive} & 16 & NN\\
\hline
\end{tabular}}
      \captionof{table}{\small\TBDRAFT{Benchmarks from low dimension academic models to complex systems with handcrafted or NN controllers.}}
      \label{table:density}
    \end{minipage}\hfill
  \begin{minipage}[b]{0.58\textwidth}
    \centering
    \includegraphics[width=0.99\textwidth]{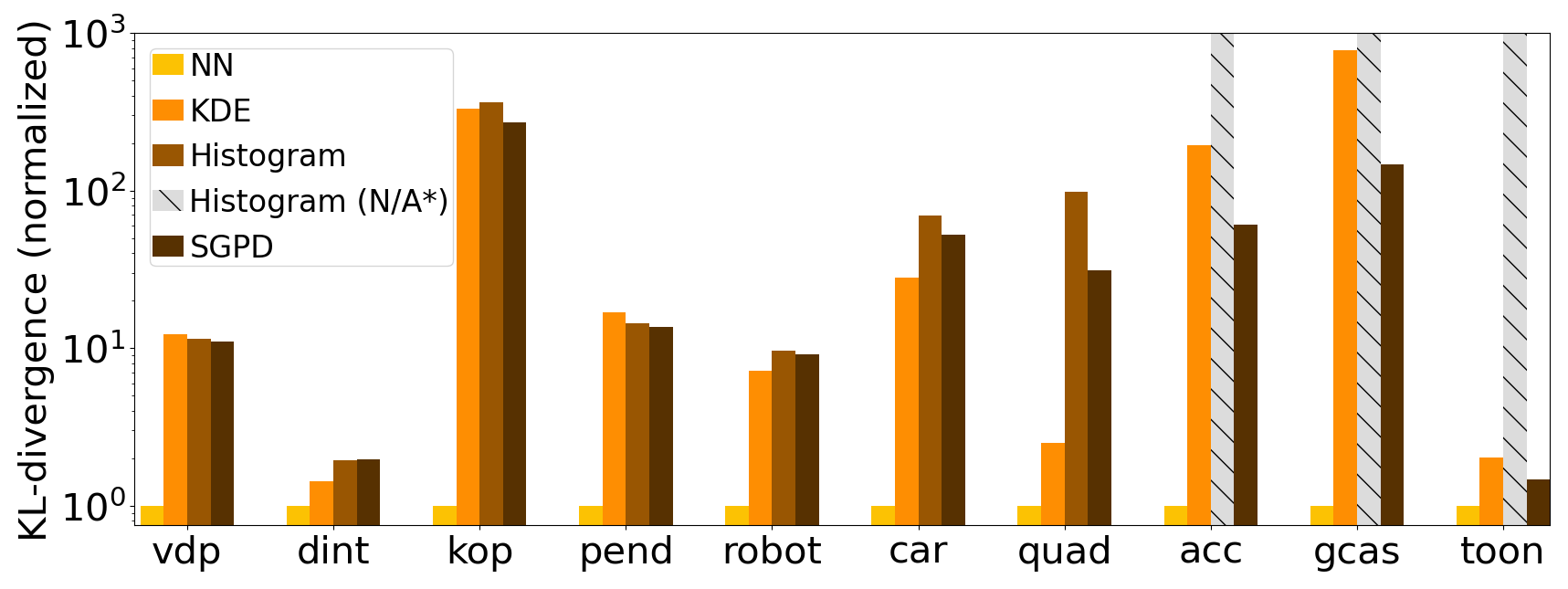}
    \captionof{figure}{\small\TBDRAFT{KL divergence with ODE-simulated density. Our method consistently outperforms other baselines. The histogram method cannot estimate the density for ``acc", ``gcas" and ``toon".}}
    \label{fig:density}
  \end{minipage}
  \hfill
  \end{minipage}
  \vspace{-1em}
\end{table}

\vspace{-0.3em}
\subsection{Implementation details}
\vspace{-0.5em}
\noindent{\textbf{Datasets:}} We investigate 10 benchmark dynamical systems as reported in Table~\ref{table:density}. These benchmark systems range from low dimension academic models (2$\sim$3 dimensions) to complex and even black-box systems (13$\sim$16 dimensions) controlled by handcrafted or NN controllers. All controllers (except for the ground robot navigation example) are from the original references. 
More details (the model description and initial distributions) will be provided in \DRAFT{\suppref{C}}.

\noindent{\textbf{Training:}} For each system, we generate 10k trajectories through simulation with varied trajectory lengths from 10 to 100 time-steps, depending on different configurations of the simulation environment. We use 80\% of the samples to train the NN as described in Sec.~\ref{sec:nn} and use the rest for validation. For the trajectory data, we collect the system states and compute for the system divergence term. For black-box systems,  we use the gradient perturbation method to approximate the derivatives. We use feed-forward NN which has 3 hidden layers with 64 hidden units in each layer. We use PyTorch~\citep{paszke2019pytorch} to train the NN and the training takes 1$\sim$2 hours on an RTX2080 Ti GPU.

\vspace{-0.3em}
\subsection{Density estimation verification}
\vspace{-0.5em}
We first test the density estimation accuracy of our learned NN. We compare our approach with other baselines including kernel density estimation (KDE)\DRAFT{, Sigmoidal Gaussian Process Density (SGPD)~\cite{donner2018efficient}} and the histogram approach. For each simulation scenario, we first solve Eq.~\eqref{liouville-ode} to generate 20k $\sim$ 100k trajectories of (state, density) pairs, and treat this density value as the ground truth. For the KDE method, we choose an Epanechnikov kernel.
We then measure the KL divergence between the density estimate of each method and the ground truth. As shown in \DRAFT{Fig.~\ref{fig:density}}, our approach has consistently outperformed KDE\DRAFT{, SGPD} and histogram approaches, with the largest reduction of 99.69\% in KL divergence when compared with the histogram approach for the Kraichnan-Orszag system, while our method doesn't use any ODE generated density values during training. Also in high-dimension systems (dimension $\geq 7$), the histogram approach fails to predict the density due to the curse of the dimensionality, whereas our approach can always predict the density, with a 30.13\% to 99.87\% decrease in KL divergence comparing to KDE. More plots will be given in \DRAFT{\suppref{D}}.

\vspace{-0.3em}
\subsection{Reachable set distribution analysis}
\vspace{-0.5em}
\label{sec:reachable_set_analysis}
\begin{figure}[!thbp]
\begin{subfigure}[b]{1.0\textwidth}
\includegraphics[width=0.21\textwidth]{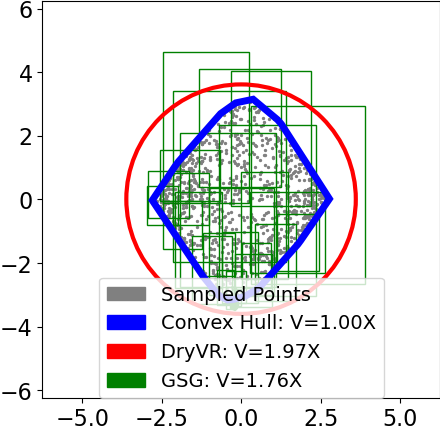} \hfill
\includegraphics[width=0.21\textwidth]{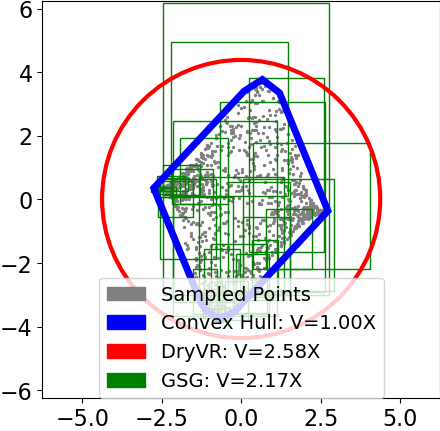} \hfill
\includegraphics[width=0.21\textwidth]{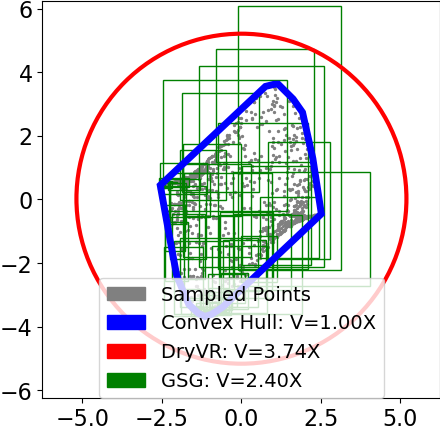} \hfill
\includegraphics[width=0.21\textwidth]{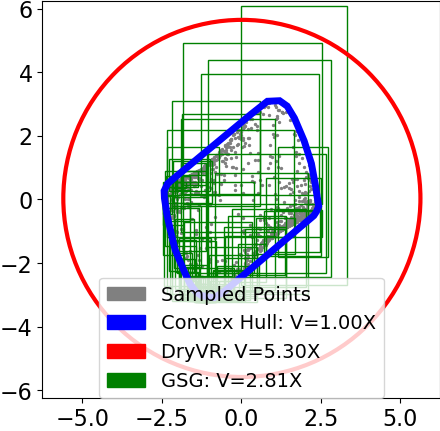} \hfill
\caption{\small The system forward reach set computed by the ConvexHull method(blue)~\citep{lew2020sampling}, DryVR(red)~\citep{fan2020fast} and GSC(green)~\citep{everett2020robustness}. We further show the relative volume of each kind of reachable set, comparing to the volume of the convex hull of the sampled points (gray). As time evolves, the conventional reachability methods often result in a larger over-approximation of the reachable sets with an increasing estimation error.}
\end{subfigure}



\begin{subfigure}[b]{1.0\textwidth}
\includegraphics[width=0.21\textwidth]{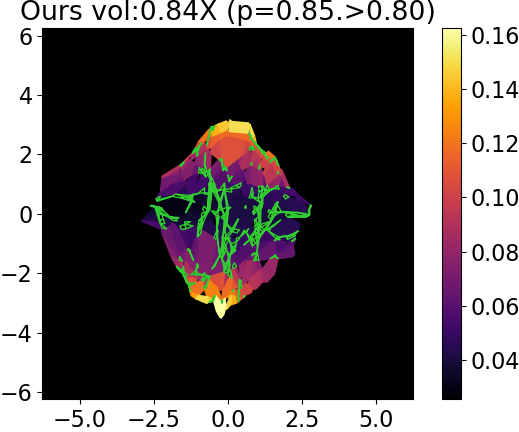} \hfill
\includegraphics[width=0.21\textwidth]{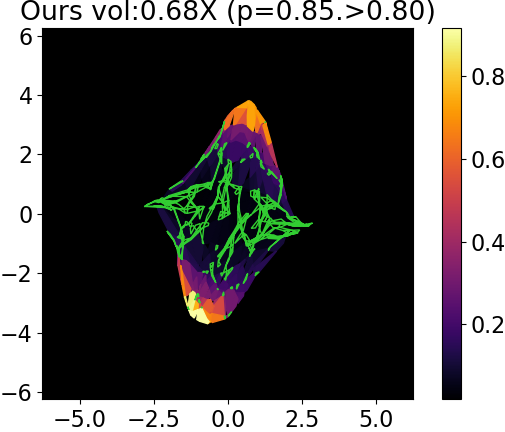} \hfill
\includegraphics[width=0.21\textwidth]{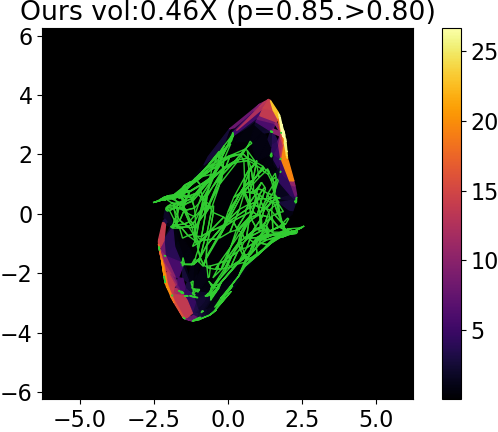} \hfill
\includegraphics[width=0.21\textwidth]{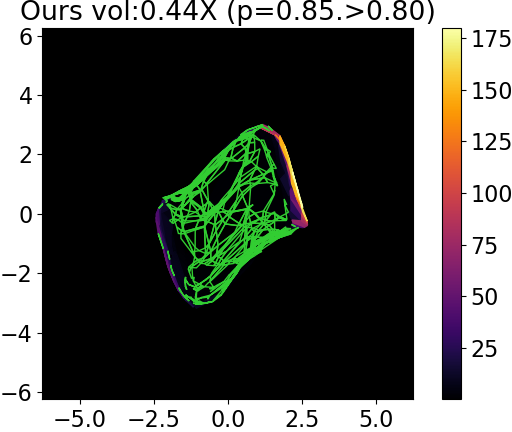} \hfill
\caption{\small The system forward reach set derived by RPM. The reachable sets are represented in polyhedral cells. The color ranged from dark purple to light yellow indicates the density inside the polyhedral cells. The edges colored in green indicate the boundaries of the RPM polyhedral cells with density below a threshold.}
\end{subfigure}

\centering
\caption{\small The Van der Pol oscillator forward reachable set comparison between (a) the worst-case reachability methods~\citep{fan2020fast,everett2020robustness} and (b) our probabilistic approach. Our approach clearly identifies reachable sets in high and low densities, and shows that as time evolves, the states will concentrate on a limit cycle, which is as expected.\vspace{-0.5em}}
\label{fig:reachability-vdp}
\end{figure}
\begin{figure}[!thbp]
\begin{subfigure}[b]{0.24\textwidth}
\includegraphics[width=1.0\textwidth]{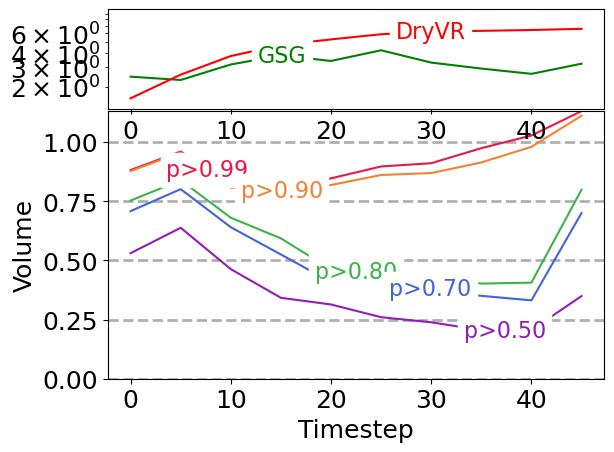}
\caption{\small Van der Pol system}
\end{subfigure}
\begin{subfigure}[b]{0.24\textwidth}
\includegraphics[width=1.0\textwidth]{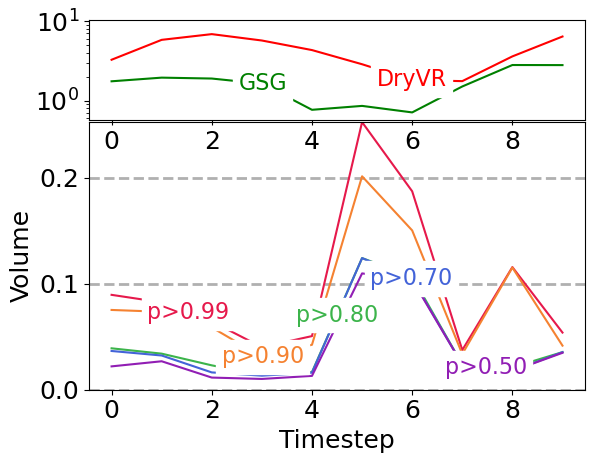}
\caption{\small Double integrator}
\end{subfigure}
\begin{subfigure}[b]{0.24\textwidth}
\includegraphics[width=1.0\textwidth]{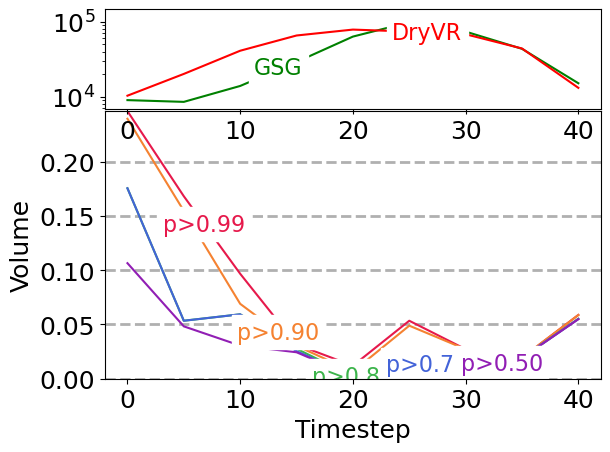}
\caption{\small \DRAFT{Ground robot model}}
\end{subfigure}
\begin{subfigure}[b]{0.24\textwidth}
\includegraphics[width=1.0\textwidth]{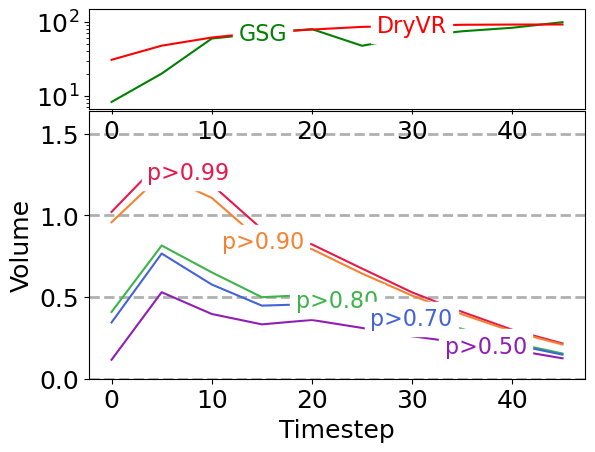}
\caption{\small FACTEST car model}
\end{subfigure}
\centering
\caption{\small The (relative) volume of reachable set  with different probability level using our method, and comparison to other methods. The top part of each figure is in logarithm scale.\vspace{-1em}}
\label{fig:eps}
\end{figure}
\paragraph{Forward reachable set distribution analysis.}
Being confident that our approach is able to provide an accurate state density estimation, we extend our NN to do distribution analysis, which is a valuable technique in safety-related applications like autonomous driving. Here we use an existing reachability tool RPM~\citep{vincent2020reachable} to compute the forward reachable sets with probability bounds. Details about how to derive the density and probability bound for the reachable sets are presented in Sec.~\ref{sec:rpm} and in \DRAFT{\suppref{B}}. Also, RPM was only able to parse the NN for Van der Pol, Double integrator, ground robot navigation, and the FACTEST car model. It fails in handling other high-dimension complex systems due to numerical issues when partitioning for the input set. Thus, we only report the results on those 4 models in this section. The main purpose is to show that for some systems the density tends to concentrate on certain states, where a small portion of the reachable sets contains the majority of states that are more likely to be reached. Therefore, our method can better quantify risks than worst-case reachability, by providing a flexible threshold for the probability of reachability.

We start with the Van der Pol oscillator. The initial states are uniformly sampled from a square region: $[-2.5, 2.5] \times [-2.5, 2.5]$. As illustrated in Fig.~\ref{fig:van-der-pol}, the system states will gradually converge to a limit cycle. Using worst-case reachability analysis for this system will result in a very conservative over-approximation, and this over-approximation will propagate over time and lead to increasing \DRAFT{conservativeness} of the reachable set estimation. As shown in Fig.~\ref{fig:reachability-vdp}(a), for the worst-case methods like DryVR(red)~\citep{fan2020fast, du2020online} and GSG(Green)~\citep{everett2020robustness}, the volume of their estimated reachable set relative to the volume of the convex hull of the system states keeps increasing over time, from 1.9670X to 5.3027X for the DryVR~\citep{fan2020fast} approach, and from 1.7636X to 2.8086X for GSG~\citep{everett2020robustness}. However, our method can give the probability bound for every reachable set in the state space as shown in the heatmap in Fig.~\ref{fig:reachability-vdp}(b), clearly identifying the region around the limit cycle in high density (bright color), and the rest space in low density (dark color).

We can also compute the relative volume of the reachable sets (comparing to the convex hull) preserving different levels of reachable probability, whose evolution over time reflects the system's tendency for concentration. As shown in Fig.~\ref{fig:eps}, we use the above 4 systems and study the volume of the reachable set with probability threshold 0.50, 0.70, 0.80, 0.90, and 0.99.
As expected, the relative volume will increase as the probability threshold increases.
In all cases, there exist some time instances where a small volume of the reachable set actually preserves high probability,
which shows the state concentration exists in many existing systems. While as shown in Fig.~\ref{fig:eps}, the worst-case reachability tools can only generate one curve which presents the (relative) volume of the reachable set that covers all possible states. Not surprisingly, the worst-case reachability tools give very conservative results. We believe using our proposed method to do reachable set distribution analysis will benefit future study for systems with uncertainty, and for systems where the failure case is inevitable but happens with a low probability. \DRAFT{More comparisons with state-of-the-art reachability methods (Verisig~\citep{ivanov2019verisig}, Sherlock~\citep{dutta2017output} and ReachNN~\citep{fan2020reachnn}) are shown in \suppref{H}}.
\paragraph{Online safety verification under different initial state distributions.}
\begin{wrapfigure}{R}{0.5\textwidth}
\begingroup
\vspace{-2.0em}
\setlength{\columnsep}{0pt}
\setlength{\intextsep}{-0pt}
\begin{subfigure}[b]{0.36\textwidth}
\includegraphics[width=1.0\textwidth]{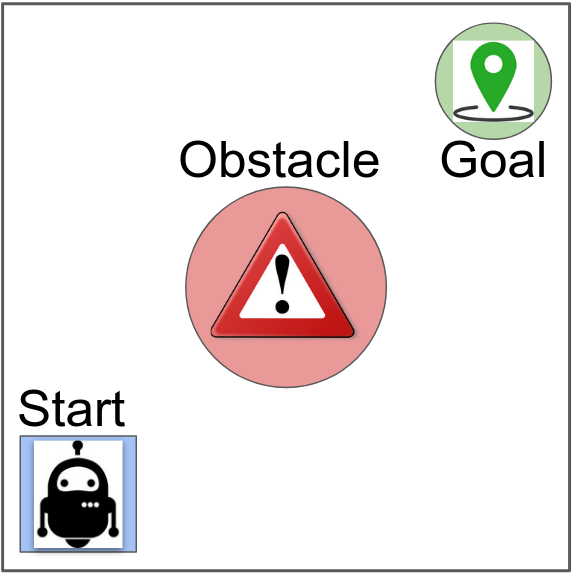}
\caption{\small\vspace{-0.5em}}
\end{subfigure}
\hfill
\begin{subfigure}[b]{0.6\textwidth}
\includegraphics[width=1.0\textwidth]{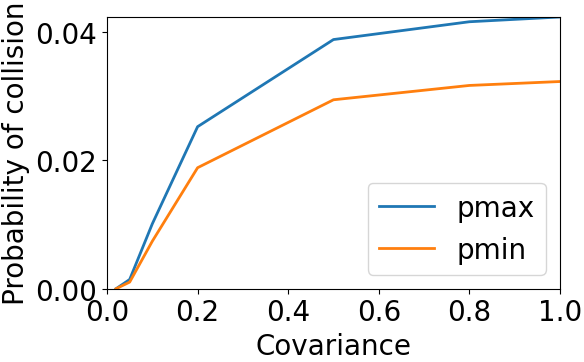}
\caption{\small\vspace{-0.5em}}
\end{subfigure}
\caption{\small \TBDRAFT{(a) A robot tries to reach the goal while avoiding the obstacle.} (b) Probabilistic safety verification under different initial distributions. As the \TBDRAFT{robot} has less uncertainty in its initial position and velocity, it will have lower probability to collide.\vspace{-1em}
}
\label{fig:safe}
\endgroup
\end{wrapfigure}
Since our approach learns the {\it \densitygain} instead of absolute density value, it has the flexibility to estimate online reach sets distribution with any possible (bounded support) initial distributions. \DRAFT{Consider the safety verification for the ground robot navigation problem (shown in Fig.~\ref{fig:safe}(a)): The probability of colliding with a obstacle in the center of the map} is determined by the initial state distribution \footnote{How to compute the probability of a reachable set is discussed in Sec.\ref{sec:rpm}}, which is parametrized as a truncated Gaussian distribution $N(\mu, \sigma^2 I)$ where $\mu$ and $\sigma$ measure the expectation and uncertainty of the initial robot state. Our method can estimate the upper and lower bounds for the probability of colliding with the obstacle, and this safety evaluation process can run in faster than 50Hz with parallel computation and heuristic searching used (see details in \DRAFT{\suppref{E}}). As shown in Fig.~\ref{fig:safe}(b), when the initial state uncertainty $\sigma$ decreases from $1.0$ to $0.02$, the upper and lower bounds for the probability of collision decrease to close to zero (from $0.04236$ to $3.5344\times 10^{-06}$, where other worst-case reachability analysis methods can only report a collision is inevitable, without quantifying the corresponding risks. \DRAFT{This also shows the advantage of our approach in adapting to different density conditions in computation without retraining or fine-tuning.}

\paragraph{\DRAFT{Limitations and trade-offs of performing reachable set distribution analysis.}} \DRAFT{
Experimental results show that our method can compute much less conservative probabilistic reachable sets than most worst-case reachability methods. This less conservative result benefits from RPM which can provide exact NN reachability analysis by sacrificing scalability. Therefore RPM also constrains us from performing online reachability analysis for high-dimensional systems or systems with large initial sets.
Technically, we are solving a harder problem than worst-case reachability approximation, as we need not only the reachable set, but also the density over those reachable states. Like worst-case analysis, this is the reason why it can only scale to lower-dimensional systems and smaller initial sets when we want to perform accurate reachable computation.
We believe that our approach can better quantify risks under different conditions, especially when unsafe is inevitable (similar to Fig.~\ref{fig:safe}(a)).
Our method can also give worst-case reachability by taking all output reachable cells produced by RPM regardless of their density. This worst-case reachability using RPM is less conservative than other NN reachability tools, at the cost of not scaling to high-dimensional systems.}


%

\section{Conclusion and discussion}\vspace{-1em}

In this paper, we propose a Neural Network (NN)-based probabilistic safety verification framework that \DRAFT{can estimate state density, compute reachable sets and corresponding probability.} \DRAFT{Our Liouville-based NN can accurately estimate the state density even for high-dimension systems.} \DRAFT{Our probabilistic reachable set framework} can handle nonlinear (and potentially black-box) systems with varying initial state distributions and can be used for fast online safety verification. \DRAFT{We recognize that the task of computing probabilistic reachable sets is very useful, and our method is more helpful than worst-case reachability} particularly when the system states are more likely to concentrate. One limitation of our approach is that the NN reachability tool we used (RPM) cannot handle high-dimension systems or cases where the initial set is very large, due to the numerical issues when partitioning for polyhedral cells. \DRAFT{
This limitation is due to the scalability and accuracy trade-off of NN reachability, which is an independent problem from our paper.} \DRAFT{We plan to explore other NN reachability methods and more complicated hybrid systems in real-world applications.}
\acknowledgments{The NASA University Leadership initiative (grant \#80NSSC20M0163) and Ford Motor Company provided funds to assist the authors with their research, but this article solely reflects the opinions and conclusions of its authors and not any NASA or Ford entity.}
\bibliography{z7_references}
\clearpage
\renewcommand\thesection{\Alph{section}}
\setcounter{page}{1}
\setcounter{section}{0}
\setcounter{figure}{0}

\begin{center}
\begin{doublespacing}
{\LARGE\textbf{Learning Density Distribution of Reachable States for Autonomous Systems (Supplementary Material)}}
\end{doublespacing}
\end{center}

\section{Generalization error bound for the learning framework}
\label{sec:a}
With sufficient amount of data and a large enough neural network, we can approximate the state and density estimation at arbitrary small errors \citep{leshno1993multilayer}. 
In the language of statistical learning theory, the neural network generating functions $(\Phi_{\omega}, G_{\theta})$ is called a {\it hypothesis} and denoted by $h$. The set containing all the possible hypotheses is called the hypothesis class $\mathcal{H}$. For a hypothesis $h$ generating $(\Phi_{\omega}, G_{\theta})$, we denote $l(h,\xi_i)=\sum\limits_{(x_i^k,k\Delta T)\in\xi_i}(\Phi_{\omega}(x_0^i,k\Delta t) - x_k^i)^2+(\frac{\partial G_{\theta}(x_0^i, k\Delta t)}{\partial t} + G_{\theta}(x_0^i, k \Delta t) \cdot (\nabla \cdot f(x_i^k)))^2-\gamma$ where $\gamma\geq 0$ is an error tolerance term which is further used to derive the probabilistic guarantee. Assume that the optimization problem in Eq.\eqref{eq:optimization} is feasible, and $\hat{\omega}$ and $\hat{\theta}$ solve Eq.~\eqref{eq:optimization}. Let $\hat{h}_N$ be the hypothesis that generates $(\Phi_{\hat{\omega}}, G_{\hat{\theta}})$. Furthermore, assume that $|l(\cdot,\cdot)|\leq B_l$, and denote the sample distribution $\mathcal{D}$ (where the training sample trajectories are sampled from). Then according to  Theorem 5 in \citep{srebro2010smoothness}, the following statement holds with probability at least $1-\delta$ over a training data set consisting of $N$ i.i.d. random trajectories:
\begin{equation}
\begin{aligned}
    &\mathop{\mathbb{P}}(\mathop{\mathbb{E}}\limits_{\xi\sim\mathcal{D}}l(\hat{h}_N,\xi)>0) \leq K \left(\frac{\log^3 N}{\gamma^2}\mathfrak{R}_N^2(\mathcal{H})+\frac{2\log(\log(4B_l/\gamma)/\delta)}{N}\right)
    \end{aligned}
\end{equation}
where $K$ is a universal constant, and $\mathfrak{R}_N(\mathcal{H})$ is the Rademacher complexity for $\mathcal{H}$ defined as:
\begin{equation}
    \mathfrak{R}_N(\mathcal{H})=\sup\limits_{\xi_1, \xi_2, \cdots, \xi_N}\left[ \mathop{\mathbb{E}}\limits_{\sigma}\left[ \sup\limits_{h\in\mathcal{H}}\frac{1}{N}\sum\limits_{i=1}^N\sigma_il(h,\xi_i) \right] \right]
\end{equation}
where $\sigma = [\sigma_1, \sigma_2, \cdots, \sigma_N]$ are i.i.d. random variables with $\mathbb{P}(\sigma_i=1)=\mathbb{P}(\sigma_i=-1)=0.5$. 

\begin{figure}[!htbp]
\includegraphics[width=0.5\textwidth]{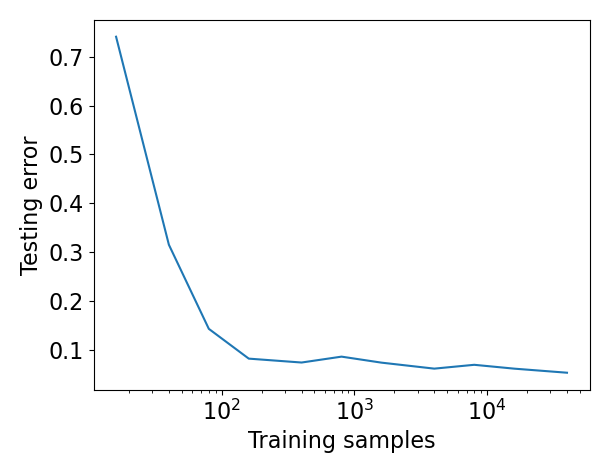}
\centering
\caption{The testing error decreases as more training samples are used.}
\label{fig:n_error}
\end{figure}

\noindent{\textbf{Remarks:}} Here we reduce bounding the generalization error to bounding the Rademacher complexity $\mathfrak{R}_N(\mathcal{H})$, where $\mathfrak{R}_N(\mathcal{H})$ can be further bounded as $\mathfrak{R}_N(\mathcal{H})\leq o (\frac{k}{N})$ for Lipschitz parametric function classes (including neural networks) where $k$ denotes the number of learnable parameters~\citep{boffi2020learning}[Theorem 4.2.]. In this way, we show that for a fixed error threshold $\gamma$, as the number of training samples $N$ increases, the probability that our learning framework fails to satisfy the Liouville equation or fails to estimate the system dynamics will gradually decrease to zero. We show an empirical result to support this in Figure \ref{fig:n_error}. 
For the Van der Pol Oscillator benchmark example, we train the neural network with different numbers of training samples (from $8\times10^0\sim8\times 10^4$) and report the testing error (mean square error for the state estimation and density concentration function comparing to the groundtruth) for a fixed testing set. As the number of training samples increases, the testing error gradually converges to zero. 

Assume the functions on the right hand side of Eq.~\eqref{liouville-ode} are uniformly Lipschitz continuous in $(x,\rho)$, then the function will have a unique solution according to Picard-Lindelöf theorem\citep{coddington1955theory}[Theorem I.3.1]. Then if our estimator satisfies the Liouville equation everywhere, we can recover the groundtruth density concentration function as well as the system dynamics.

\section{Implementation details for system reachable set probability computation using RPM}
\label{sec:b}
\subsection{Online query set probability bound computation under different initial state distributions}
\label{sec:b1}

\begin{figure}[!htbp]
\includegraphics[width=0.99\textwidth]{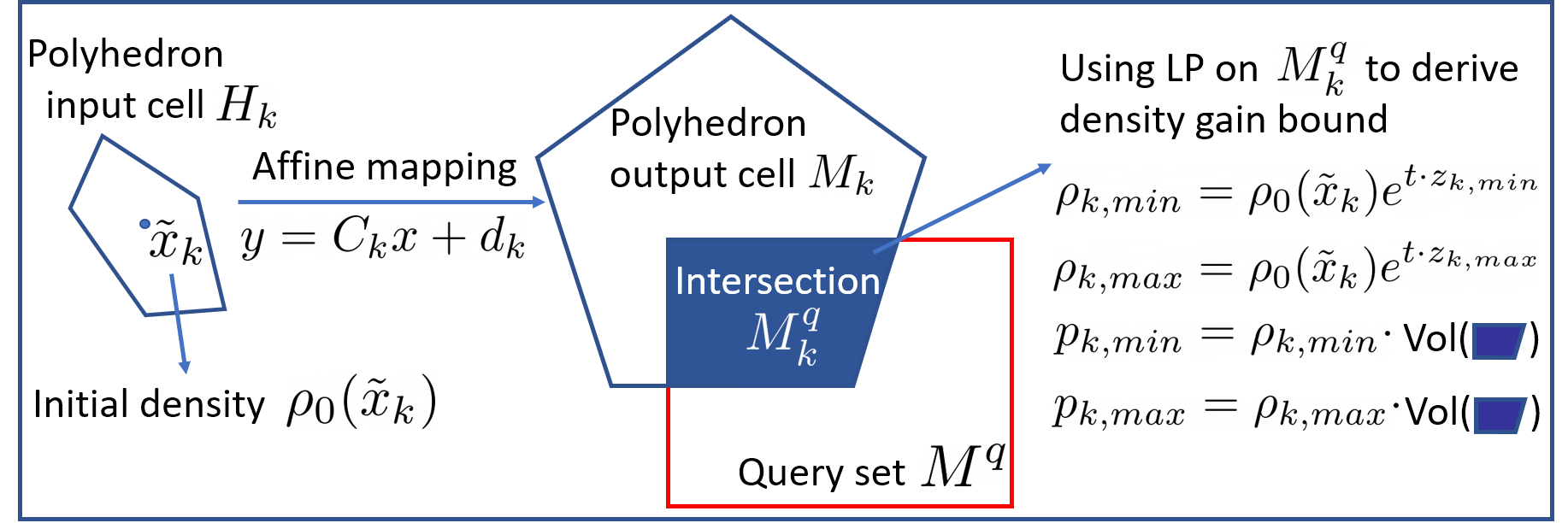}
\centering
\caption{Illustration for the online query set probability bound computation (for an output cell $M_k$).}
\label{fig:b1-fig}
\end{figure}

The problem formulation is: given a query set $R^q$ with \DRAFT{\it \densitygain} constraints $[z_{min},z_{max}]$ (the range that the \DRAFT{\it \densitygain} can change from the initial condition to the terminal condition; if this constraint is not specified, the default value is $-\infty \leq z \leq \infty$), compute the probability that the system will reach this query set (with optional density constraints). 

In our case, when using RPM to compute the reachable sets, we represent $R^q$ as a polyhedron, and since $[z_{min}\leq z\leq z_{max}]$ is a set of linear inequality constraints, the set $M^q=\{(z, v) | z_{min}\leq z\leq z_{max}, v\in R^q\}$ is also a polyhedron. At each time step $t$, from Sec.~\ref{sec:rpm} we can represent the NN input cells, affine mapping and output cells at this time step as $\{(A_k, b_k, C_k, d_k, E_k, f_k)\}_{k=1}^N$ (here we omit the subscript for $t$ for the brevity in the notation) where each input cell is a polyhedron $H_k=\{x\in\mathbb{R}^{d+1}|A_k x \leq b_k\}$, with an affine mapping $y=C_k x + d_k$ and the resulting output cell is also polyhedron $M_k=\{y\in \mathbb{R}^{d+1}|E_k y\leq f_k\}$. Then for each output cell $M_k$, we check for the intersection between the query cell and the output cell $M_k^q=\{y | y\in M^q,\,E_k y \leq f_k \}$. Next we can derive the intermediate \DRAFT{\it \densitygain} bound $[z_{k,min}, z_{k,max}]$ on $M_k^q$ by solving the following Linear Programming problem (here taking $z_{k,min}$ as an example; to solve $z_{k,max}$ we just need to change the ``min" to ``max" in the objective function in Eq.~\ref{eq:lp}; and here $[y]_0$ denotes the first coordinate of $y$, thus $[y]_0=z$ as we denote $y=(z,x)^T$):
\begin{equation}
\begin{cases}
    \min~[y]_0\\
    \text{s.t. } y\in M^q_k\\
\end{cases}
\label{eq:lp}
\end{equation}
After we derive the bound for $z$ on $M_k^q$, the density bound for $M_k^q$ is computed as (similar to Eq.(6)):
\begin{equation}
    \begin{cases}
        \rho_{k,min}=\rho_0(\tilde{x}_k) e^{t\cdot z_{k,min}}\\
        \rho_{k,max}=\rho_0(\tilde{x}_k)e^{t\cdot z_{k,max}}
    \end{cases}
    \label{eq:online-safe}
\end{equation}
where $\tilde{x}_k$ is the center of $H_k$. And the probability bound can be computed by $p_{k,min}=\rho_{k,min}\cdot \text{Vol}(M_k^q)$ and $p_{k,max}=\rho_{k,max}\cdot \text{Vol}(M_k^q)$ where the $\text{Vol}(M_k^q)$ is the volume for the intersection. Finally the probability of the system reach this query set at time $t$ is bounded by $ \left[P_{min}=\sum\limits_{k=1}^N p_{k,min}, P_{max}=\sum\limits_{k=1}^N p_{k,max}\right]$. An illustrative figure is shown in Fig.~\ref{fig:b1-fig}

\noindent\textbf{Remarks:} This algorithm can be used for online safety verification under different initial state distributions by just representing the dangerous set in $R^q$, and changing the $\rho_0(\cdot)$ function in \eqref{eq:online-safe} on the fly. Here we approximate the density distribution in $H_k$ using the density evaluated at $\tilde{x}_k$ which is the center of $H_k$ - the accuracy of this approximation will converge to 1 as the partition on $\rho_0$ gets finer.

\subsection{Backward reachable set probability computation}
\label{sec:b2}
The problem formulation is: given a query set $R^q$ with \DRAFT{\it \densitygain} constraints $[z_{min},z_{max}]$ (the range that the \DRAFT{\it \densitygain} can change from the initial condition to the terminal condition; if this constraint is not specified, the default value is $-\infty \leq z \leq \infty$), compute for all possible initial conditions as well as probabilities that lead the system to reach the query set (with optional density constraints).

Similar to Sec.~\ref{sec:b1}, we can denote this query set as $M^q=\{(z,v)|z_{min}\leq z\leq z_{max}, v\in R^q\}$. At each time step $t$, the NN input cells, affine mapping and output cells are $\{(A_k, b_k, C_k, d_k, E_k, f_k)\}_{k=1}^N$ where each input cell is a polyhedron $H_k=\{x\in\mathbb{R}^{d+1}|A_k x \leq b_k\}$, with an affine mapping $y=C_k x + d_k$ and the resulting output cell is also polyhedron $M_k=\{y\in \mathbb{R}^{d+1}|E_k y\leq f_k\}$. Then for each output cell $M_k$, we check for the intersection between the query cell and the output cell $M_k^q=\{y | y\in M^q,\,E_k y \leq f_k \}$. Using the affine mapping with invertible $C_k$\footnote{In practice, $C_k$ is in high probability to be invertible. This is because the set of all non-invertible random matrices forms a hyper-surface with Lebesque measure zero. When $C_k$ is singular, we can use elimination method like Fourier-Motzkin elimination as in~\citep{vincent2020reachable} to derive the set representation in the input side.}, we can derive the pre-image of this intersection to be $H_k^q=\{x| x = C_k^{-1}y-C_k^{-1}d_k,\, y\in M_k^q\}$. Thus the reachable set can be computed using projection: $R_k^{i,q}=\{x\in\mathcal{X}|(x,t)\in H_k^q\}$ and the corresponding probability is $p_{k}^{i,q}=\text{Vol}(R_k^{i,q}) \rho_0(\tilde{x}^{i,q}_k)$ where $\rho_0(\cdot)$ is the initial state distribution function and $\tilde{x^{i,q}_k}$ is the center of $R_k^{i,q}$. By performing this for all output cells and for all time steps $t$, we derive the backward reachable set $\{\{(R_{t,k}^{i,q}, p_{t,k}^{i,q})\}_{k=1}^{N}\}_{t=0}^{T-1}$.

\subsection{Speed up the probability computation by using hyper-rectangle heuristic}
\label{sec:b3}
The computation in both Sec.~\ref{sec:b1} and Sec.~\ref{sec:b2} requires checking the intersection between polyhedral $H_i$ and $H_j$, where one approach is to check whether a feasible solution exists for the linear programming problem : $\min 0^T x,\, \text{s.t. } x\in H_i \cap H_j$. Solving this for $x\in\mathbb{R}^{n}$ requires $O(n^{2.5})$ time when the interior method is used. To speed up the intersection checking process, we introduce a hyper-rectangle heuristic: at the pre-processing stage, we over-approximate each polyhedron $H_i$ by its outer hyper-rectangle $\tilde{H}_i$ (derived by computing the range for the vertices of $H_i$ in each dimension). When checking for the polyhedron intersection between $H_i$ and $H_j$, we first check whether their corresponding hyper-rectangles $\tilde{H}_i$ and $\tilde{H}_j$ will intersect. If $\tilde{H}_i$ and $\tilde{H}_j$ do not intersect, then it is guaranteed that the polyhedra $H_i$ and $H_j$ won't intersect. Otherwise, we further check the intersection of $H_i$ and $H_j$ by using the interior method. Checking hyper-rectangles' intersection can be implemented in $O(n)$, hence greatly accelerates the computation process. A detailed computation time comparison will be presented in Sec.~\ref{sec:e}.

\section{Simulation environments}
\label{sec:c}
In this section, we present the implementation details for all 10 simulation environments used in our main paper, sorted in the same order as shown in Table.~\ref{table:1}.

\subsection{Van der Pol Oscillator}
\label{sec:c1}
Consider the Van der Pol Oscillator problem: $\frac{d^2x}{dt^2}-\mu(1-x^2)\frac{dx}{dt}+x=0$ where the position variable $x$ is a function of $t$ and the scalar parameter $\mu$ indicates the strength of the system damping effect. By doing a transformation: $y=\dot{x}$, the original problem can be shaped to the following 2d system dynamics:
\begin{equation}
    \begin{cases}
        \dot{x} = y\\
        \dot{y} = \mu (1-x^2)y-x
    \end{cases}
\end{equation}
where the divergence term $\nabla \cdot f$ used in \eqref{liouville-ode} can be computed as: $\nabla \cdot f=\mu(1-x^2)$. In the simulation, we set $\mu=1.0$, the initial state distribution as an uniform distribution $\mathcal{U}_{[-2.5, 2.5]\times[-2.5, 2.5]}$ and the time step duration $\Delta t=0.05s$. We run each simulation for 50 time steps to collect the trajectories.

\subsection{Double Integrator with an NN controller}
\label{sec:c2}
We consider a discrete double integrator system introduced in~\citep{hu2020reach}:
\begin{equation}
    \begin{pmatrix}x_{t+1}\\y_{t+1}\end{pmatrix}=\begin{bmatrix}1 & 1 \\ 0 & 1 \end{bmatrix} \begin{pmatrix}x_{t}\\y_{t}\end{pmatrix} + \begin{bmatrix}
    0.5\\ 1
    \end{bmatrix} u_t
\end{equation}
where \DRAFT{$(x_t,y_t)^T$} denotes the 2d state variable, and $u_t$ is the output of a neural network controller which is trained to mimic the behavior of an MPC controller~\citep{hu2020reach,everett2021efficient}. We convert the system to the continuous system with state $(x, y)^T$ and time step duration $\Delta t=1.0s$ as :
\begin{equation}
    \dot{\begin{pmatrix}x \\ y\end{pmatrix}}=\begin{bmatrix} 0 & 1 \\ 0 & 0\end{bmatrix} {\begin{pmatrix}x \\ y\end{pmatrix}} + \begin{bmatrix}0.5 \\ 1\end{bmatrix} u
\end{equation}
and here the divergence term $\nabla \cdot f$ used in \eqref{liouville-ode} can be computed as\DRAFT{: $\nabla\cdot f=0.5\frac{\partial u}{\partial x} + \frac{\partial u}{\partial y}$, where the $\frac{\partial u}{\partial x}$ is the gradient of the neural network controller output $u$ with respect to the input $x$ (and similar for $\frac{\partial u}{\partial y}$and $y$)} and can be calculated using automatic differentiation engine in PyTorch~\citep{paszke2019pytorch}. We set the initial state distribution as an uniform distribution $\mathcal{U}_{[-0.5, 4.0]\times[-1.0, 1.0]}$. Similar to \citep{hu2020reach,everett2021efficient}, we run each simulation for 10 time steps to collect the trajectories.

\subsection{Kraichnan-Orszag system}
\label{sec:c3}
The system dynamics of the Kraichnan-Orszag problem~\citep{orszag1967dynamical,nakamura2019density} is defined as:
\begin{equation}
    \begin{cases}
    \dot{x}_1 = x_1 x_3 \\ 
    \dot{x}_2 = -x_2 x_3 \\
    \dot{x}_3 = -x_1^2 + x_2^2
    \end{cases} 
\end{equation}
and here an interesting fact is that the divergence term $\nabla \cdot f$ used in \eqref{liouville-ode} is just: $\nabla\cdot f=x_3-x_3+0=0$, which means the density along each trajectory won't change over time, and only depends on the initial state distribution. Similar to~\citep{nakamura2019density}, we set the initial state $x(0)=(x_1(0),x_2(0),x_3(0))^T$ distribution as an Gaussian distribution with:
\begin{equation}
\begin{cases}
    x_1(0)\sim \mathcal{N}(1, 1/4^2)\\
    x_2(0)\sim \mathcal{N}(0, 1/2^2)\\
    x_3(0)\sim \mathcal{N}(0, 1/2^2)
\end{cases}
\end{equation}
where we further truncate the initial state within the range $\{0\leq x_1(0)\leq 2, -2\leq x_2(0) \leq 2, -2 \leq x_3(0) \leq 2 \}$. We set the time step duration $\Delta t=0.125s$ and run each simulation for 80 time steps to collect the trajectories.

\subsection{Inverted pendulum}
\label{sec:c4}
The inverted pendulum problem~\citep{chang2020neural} is defined as $\ddot{\theta} + \frac{b}{mL^2}\dot{\theta}-\frac{g}{L}\sin\theta -\frac{1}{mL^2}u_{LQR}=0$, where $\theta$ denotes the pendulum's relative angle to the the up-right position, $m,L,g,b$ are pre-defined parameters and $u_{LQR}$ denotes the output of an LQR controller~\citep{chang2020neural} $u=K_1\theta + K_2 \dot{\theta}$ where $K_1$ and $K_2$ are scalar-valued coefficients. To test for the system performance under different coefficient settings for the LQR controller, we include $k_1$, $k_2$ into the system state variable and study the following system dynamics:
\begin{equation}
    \begin{cases}
       \dot{\theta}= \omega \\
       \dot{\omega} = \frac{1}{m\cdot L^2}(mgL\sin\theta -b \omega + u)\\
       \dot{k_1} = 0 \\
       \dot{k_2} = 0 \\
    \end{cases}
\end{equation}
where $u=\frac{K_1}{50} e^{k_1}\theta + \frac{K_2}{50} e^{k_2} \omega $. Now the divergence term $\nabla \cdot f$ used in \eqref{liouville-ode} can be computed as: $\nabla \cdot f=-\frac{b}{mL^2}+\frac{1}{mL^2}\frac{\partial u}{\partial w}=-\frac{b}{mL^2}+\frac{K_2 e^{k_2}}{50 mL^2}$. Based on ~\citep{chang2020neural}, we set $g=9.80, L=0.50, m=0.15, b=0.00, K_1=-23.59, K_2=-5.31$, the time step duration $\Delta t=0.02s$. We set the initial state distribution as a uniform distribution $\mathcal{U}_{[-2.1, 2.1]\times[-5.5, 5.5]\times[-2.0, 2.0]\times[-2.0, 2.0]}$ and run each simulation for 50 time steps to collect the trajectories.

\subsection{Ground robot navigation with an NN controller}
\label{sec:c5}
\begin{figure}[!htbp]
\includegraphics[width=0.4\textwidth]{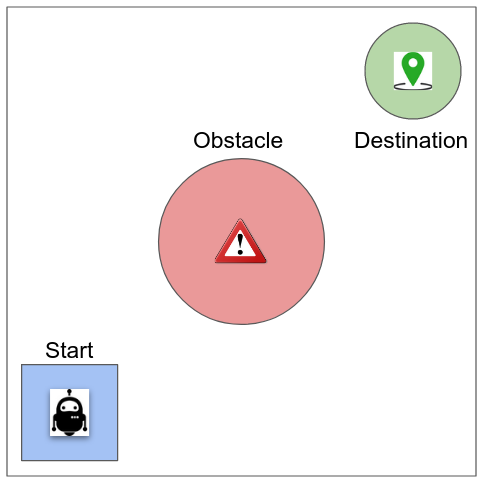}
\centering
\caption{The screenshot for robot navigation problem.}
\label{fig:robot_map}
\end{figure}
We design a ground robot navigation experiment (as shown in Fig.~\ref{fig:robot_map}), where the objective is to reach the green region $\{(x,y)|(x-x_{goal})^2+(y-y_{goal})^2\leq r_{goal}^2\}$ while avoiding to enter the red region $\{(x,y)|(x-x_{obs})^2+(y-y_{obs})^2\leq r_{obs}^2\}$. The robot is following an Dubins car model:
\begin{equation}
    \begin{cases}
    \dot{x}=v\cos\theta\\
    \dot{y}=v\sin\theta\\
    \dot{\theta}=u_w\\
    \dot{v}=u_a\\
    \end{cases}
\end{equation}
where $x,y,\theta,v$ represent robot's x and y position, heading angle and velocity respectively. We use an NN controller to output control signals $u_w,u_v$. The NN controller is a feedforward NN with 2 hidden layers and 32 hidden units in each layer. We use ReLU for the intermediate activation functions and use Tanh as the activation function for the last layer to make sure the control output is always bounded. During training, we use this NN controller to collect trajectory data and do back-propagation with the loss function: $\mathcal{L}=\sum\limits_{i=0}^{N-1}\sum\limits_{k=0}^{T-1} \alpha \left[(x^{i}_k-x_{goal})^2+(y^{i}_k-y_{goal})^2\right] + \mathbbm{1}\{d_{obs}<r_{obs}\} (d_{obs}-r_{obs})$ where $d_{obs}=\sqrt{(x^{i}_k-x_{obs})^2+(y^{i}_k-y_{obs})^2}$. Here the divergence term $\nabla\cdot f$ used in \eqref{liouville-ode} can be computed as: $\nabla \cdot f=\frac{\partial u_w}{\partial \theta} + \frac{\partial u_a}{\partial v}$. We set the initial state distribution as an uniform distribution $\mathcal{U}_{[-1.8, -1.2]\times[-1.8, -1.2]\times[0,\pi/2]\times[1.0, 1.5]}$. We run each simulation for 50 time steps with time duration $\Delta t=0.05s$ to collect the trajectories.

\subsection{FACTEST car tracking system}
\label{sec:c6}
Consider a rearwheel kinematic car in 2D scenarios where the dynamics is:
\begin{equation}
    \begin{bmatrix}\dot{x}\\\dot{y}\\\dot{\theta}
    \end{bmatrix}=\begin{bmatrix}
    \cos(\theta) & 0 \\ \sin(\theta) & 0 \\ 0 & 1 
    \end{bmatrix}\begin{bmatrix}
    v \\ \omega
    \end{bmatrix}
\end{equation}

and the corresponding errors are measured by:
\begin{equation}
\begin{bmatrix}
e_x\\e_y\\e_{\theta}
\end{bmatrix}
=\begin{bmatrix}
\cos(\theta) & \sin(\theta) & 0\\
-\sin(\theta) & \cos(\theta) & 0\\
0 & 0 & 1
\end{bmatrix}
\begin{bmatrix}
x_{ref}-x\\
y_{ref}-y\\
\theta_{ref}-\theta\\
\end{bmatrix}
\end{equation}
with $x_{ref}, y_{ref}, \theta_{ref}$ being some predefined tracking points (in this experiment, we assume the tracking points are not changing over time). With the following tracking controller defined in ($w_{ref}$ and $v_{ref}$ are referenced angular velocity and velocity respectively, $k_1,k_2,k_3$ are the parameters controlling how fast the system will converge to the reference point) \citep{fan2020fast}:
\begin{equation}
    \begin{cases}
    v=v_{ref}\cos(e_\theta)+k_1e_x\\
    \omega=\omega_{ref}+v_{ref}(k_2e_y+k_3\sin(e_\theta))
    \end{cases}
\end{equation}
and with an uncertainty error in the dynamics of $e_x$ and $e_y$ (denoted as $a$), the error dynamics become:
\begin{equation}
    \begin{bmatrix}
    \dot{e}_x\\
    \dot{e}_y\\
    \dot{e}_{\theta}\\
    \dot{a}
    \end{bmatrix}=\begin{bmatrix}
    (\omega_{ref}+v_{ref}(k_2 e_y + k_3\sin(e_\theta)))e_y-k_1 e_x\color{black}{+ae_x}\\
    -(\omega_{ref}+v_{ref}(k_2 e_y+k_3\sin(e_\theta)))e_x+v_{ref}\sin(e_{\theta})\color{black}{+ae_y}\\
    -v_{ref}(k_2e_y + k_3\sin(e_\theta))\\
    0
    \end{bmatrix}
\end{equation}
The uncertain parameter $a\in[0,1]$. We will show that althought now the reachable set will be much larger than the case when $a=0$, the probability that the system does not converge to the origin (zero-error) is very low.

Here the divergence term $\nabla\cdot f$ used in \eqref{liouville-ode} can be computed as: $\nabla \cdot f=2a-k_1-k_2 v_{ref} e_x - v_{ref} k_3 \cos e_\theta$. In our experiment, we set $x_{ref}=y_{ref}=\theta_{ref}=0,\,k_1=k_2=0.5,\,k_3=1.0,\,w_{ref}=0,\, v_{ref}=1$. We set the initial state distribution as an uniform distribution $\mathcal{U}_{[-2.1, 2.1]\times[-2.1,2.1]\times[0,0.1]\times[0.0, 1.0]}$. We run each simulation for 50 time steps with time duration $\Delta t=0.10s$ to collect the trajectories.

\subsection{6D Quadrotor with an NN controller}
\label{sec:c7}
Consider a 6D quadrotor ~\citep{everett2021efficient}:
\begin{equation}
    \dot{x}=\begin{bmatrix} 0_{3\times 3} & I_3\\ 0_{3\times 3} & 0_{3\times 3} \end{bmatrix} x + \begin{bmatrix} & g & 0 & 0 \\ 0_{3\times 3} & 0 & -g & 0\\ & 0 & 0 & 1 \end{bmatrix}^T u + \begin{bmatrix}0_{5\times 1}\\ -g \end{bmatrix}
\end{equation}
where the state vector $x$ contains 3D positions and velocities $[p_x,p_y,p_z, v_x,v_y,v_z]$, $g$ is the gravity (set to $9.8m/s^2$), and the control $u=(u_1,u_2,u_3)^T$ is from the output of an NN controller taking the state vector as the input~\citep{everett2021efficient}. Here the divergence term $\nabla\cdot f$ used in \eqref{liouville-ode} can be computed as: $\nabla \cdot f=g\cdot\frac{\partial u_1}{\partial v_x}-g\cdot\frac{\partial u_2}{\partial v_y}+\frac{\partial u_3}{\partial v_z}$. Similar to~\citep{everett2021efficient}, we set the initial state distribution as an uniform distribution $\mathcal{U}_{[4.65,4.75]\times[4.65,4.75]\times[2.95,3.05]\times[0.94, 0.96]\times[-0.05,0.05]\times[-0.5,0.5]}$. We run each simulation for 12 time steps with time duration $\Delta t=0.10s$ to collect the trajectories. 

\subsection{Adaptive cruise control system}
\label{sec:c8}
Consider a learning-based adaptive cruise control (ACC) problem with plant dynamics~\citep{tran2020nnv}:
\begin{equation}
    \begin{cases}
    \dot{x}_{rel} = v_{lead}-v_{ego}\\
    \dot{v}_{lead} = \gamma_{lead}\\
    \dot{\gamma}_{lead} = a_{lead}\\
    \dot{v}_{ego}=\gamma_{ego}\\
    \dot{\gamma}_{ego}=-2\gamma_{ego}+2u(x_{rel},v_{lead}-v_{ego}-\gamma_{ego}\tau,v_{ego}+\gamma_{ego}\tau)\\
    \dot{a}_{lead}=-2\gamma_{lead}\\
    \dot{\tau}=0
    \end{cases}
\end{equation}
here $x_{rel}$ denotes the relative distance from the leading vehicle to the ego vehicle, $v_{lead}$ and $v_{ego}$ denote the velocity of leading and ego vehicles and $\gamma_{lead}$ and $\gamma_{ego}$ denote the corresponding acceleration rates of the two vehicles ($a_{lead}$ models the change in the leading vehicle's acceleration rate, similar to the MATLAB implementation in~\citep{tran2020nnv}). And the controller $u$ is taking the relative distance, velocity, and ego vehicle's velocity as input and outputs the change in the ego vehicle's acceleration rate. We model the velocity perception uncertainty as $\tau$ and pass it through the neural network. Here the divergence term $\nabla\cdot f$ used in \eqref{liouville-ode} can be computed as: $\nabla \cdot f=-2-\frac{\partial u}{\partial (v_{lead}-v_{ego}-\gamma_{ego}\tau)} \tau + \frac{\partial u}{\partial (v_{ego}+\gamma_{ego}\tau)} \tau$. We set the initial state distribution as an uniform distribution $\mathcal{U}_{[59.0, 62.0]\times[26.0, 30.0]\times[-0.01,0.01]\times[30.0, 30.5]\times[-0.01,0.01]\times[-10.1,-9.9]\times[-2.0,2.0]}$ and run each simulation for 50 time steps with time duration $\Delta t=0.10s$ to collect the trajectories.

\subsection{F-16 ground-collision avoidance system}
\label{sec:c9}

This F-16 Ground-Collision Avoidance System (GCAS) performs a recovery maneuver for the F-16 aircraft when a ground collision is detected. The F-16 aircraft is modelled with 6 degrees of freedom (DoF) associated with 13 nonlinear equations (three equations each for forces, kinematics, moments and position of the aircraft, and one extra to capture the F-16 turbojet engine). The hierarchical control system has an outer-loop autopilot controller and an inner loop tracking and stabilizing controller (ILC). More details can be found in~\citep{heidlauf2018verification}. 
Specifically in this experiment, the GCAS drives the roll angle and its rate to 0 and then accelerates upwards to avoid ground collision. The safety specification is to make sure the altitude is always non-negative (not hitting the ground). We collect the trajectories using the F-16 simulator provided in~\citep{heidlauf2018verification}. The trajectories has a time step duration as $0.0333s$ and has 106 time steps in total. The hierarchical controller made the closed-loop F-16 system a black-box system without a clean ODE expression. Therefore, there is no analytical way to compute for the system dynamics. As we discussed in the main paper, we could approximate the divergence of the system dynamics by using gradient perturbation. Recall that for system $\dot{x}=(f_1(x), f_2(x),...,f_{d}(x))^T$, the system divergence is $\nabla \cdot f = \sum\limits_{i=1}^{d}\frac{\partial f_i}{\partial x_i}$, so we approximate the gradient for $\frac{\partial f_i(x)}{x_i}$ by $(f_i(x_1,...,x_i+\epsilon,...,x_n)-f_i(x_1,...,x_i-\epsilon,...,x_n))/(2\epsilon)$ where $\epsilon$ is a very small number and we set $\epsilon=10^{-8}$ in our experiments.

\subsection{8-car platooning with model error}
\label{sec:c10}
In this experiment we consider a 8-car platoon model~\citep{schurmann2017optimal,zhu2019inductive}. The state variable is $x\in\mathbb{R}^{15}$, where $x_1$ represents the first vehicle's (which is also the leading vehicle in the platoon) velocity, $x_{2k-1}$ (k=2,3...,8) represents the relative velocity of the $k-1$-th vehicle comparing to the $k$-th vehicle, and $x_{2k-2}$ (k=2,3...,8) represents the relative longitudinal offset of the $k-1$-th vehicle comparing to the $k$-th vehicle. The dynamics of the system hence is given by:
\begin{equation}
\begin{aligned}
    &\dot{x}_{2k-1}=\begin{cases}
    u_{1},\quad k=1\\
    u_{k-1}-u_{k}, \quad k=2,3,...8
    \end{cases}\\
    &\dot{x}_{2k-2} = x_{2k-1}+w,\,k=2,3,...8
\end{aligned}
\end{equation}
where $u=(u_1,...,u_8)^T$ is the NN controller's output (for changing the vehicles' acceleration rates) and $w$ models the noise in the vehicles' velocity dynamics. Here the neural network controller is trained via RL \citep{lillicrap2015continuous}. Here the divergence term $\nabla\cdot f$ used in \eqref{liouville-ode} can be computed as: $\nabla \cdot f=\frac{\partial u_1}{\partial x_1}+\sum\limits_{k=2}^8(\frac{\partial u_{k-1}}{\partial x_{2k-1}}-\frac{\partial u_{k}}{\partial x_{2k-1}})$. We set the initial state distribution as an uniform distribution $\mathcal{U}$ for $19.9\leq x_1\leq 20.1$, $0.9\leq x_{2k-3}\leq 1.1, k=2,3...,8$ and $-0.1\leq x_{2k-2}\leq 0.1, k=2,3...,8$ and $-0.01\leq w\leq 0.01$. We run each simulation for 50 time steps with time duration $\Delta t=0.15s$ to collect the trajectories.

\section{Forward reachable set distribution under different probability thresholds}
\label{sec:d}
\begin{figure}[!htbp]
\begin{subfigure}[b]{0.32\textwidth}
\includegraphics[width=1.0\textwidth]{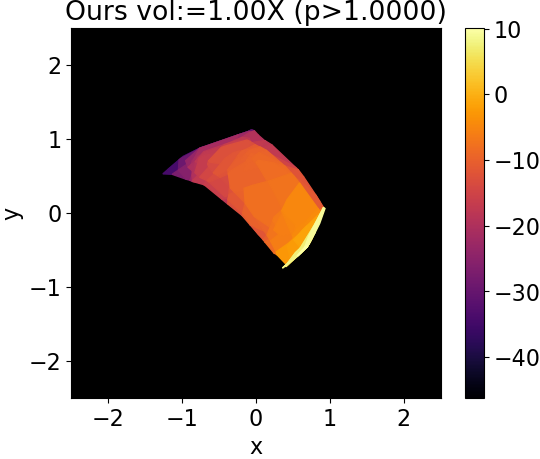} \hfill
\caption{vol=1.00X; p=1.0000}
\end{subfigure}
\begin{subfigure}[b]{0.32\textwidth}
\includegraphics[width=1.0\textwidth]{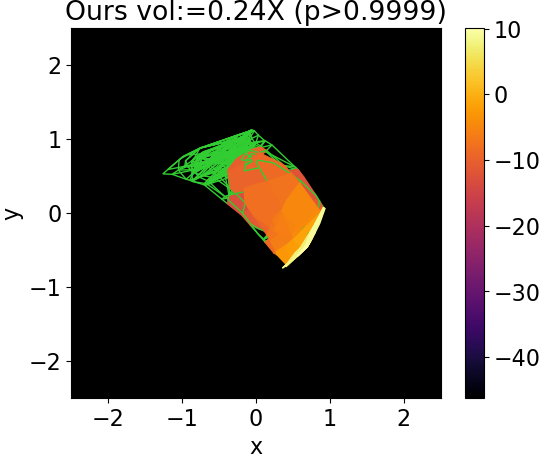} \hfill
\caption{vol=0.24X; p $>=$ 0.9999}
\end{subfigure}
\begin{subfigure}[b]{0.32\textwidth}
\includegraphics[width=1.0\textwidth]{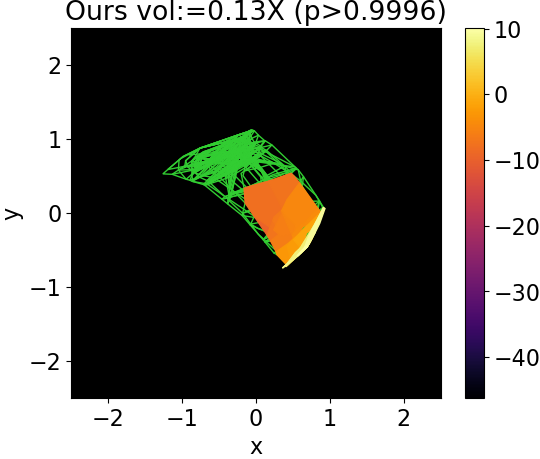} \hfill
\caption{vol=0.13X; p $>=$ 0.9995}
\end{subfigure}
\begin{subfigure}[b]{0.32\textwidth}
\includegraphics[width=1.0\textwidth]{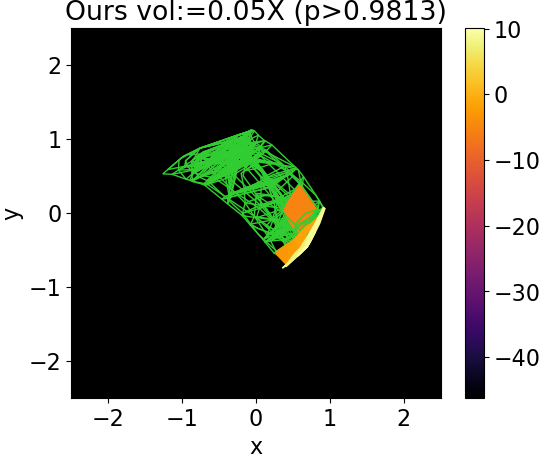} \hfill
\caption{vol=0.05X; p $>=$ 0.9813}
\end{subfigure}
\begin{subfigure}[b]{0.32\textwidth}
\includegraphics[width=1.0\textwidth]{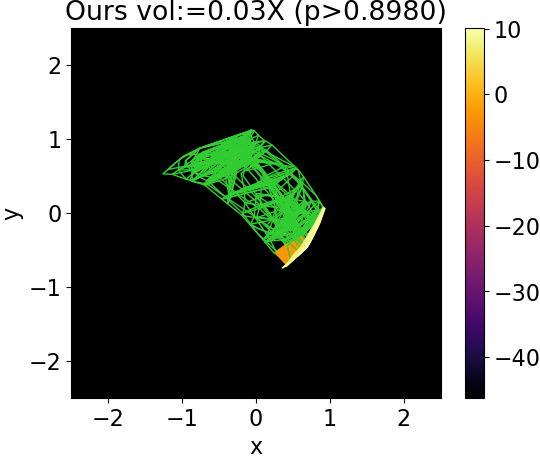} \hfill
\caption{vol=0.03X; p $>=$ 0.8980}
\end{subfigure}
\begin{subfigure}[b]{0.32\textwidth}
\includegraphics[width=1.0\textwidth]{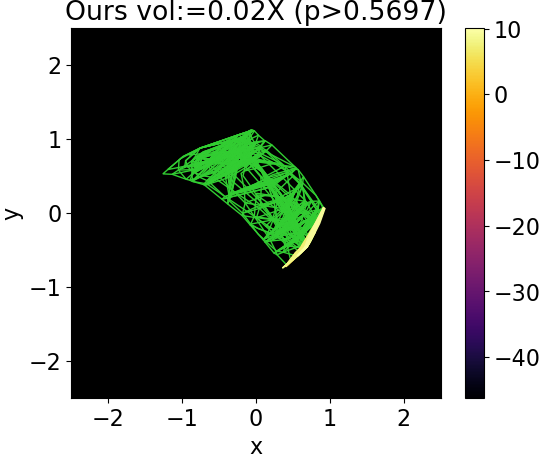} \hfill
\caption{vol=0.02X; p $>=$ 0.5697}
\end{subfigure}
\centering
\caption{The system forward reach set distribution under different probability thresholds at t=1.0s with initial condition $\mathcal{N}_1=N(\mu=(-1.7, -1.7, 0.2, 1.4)^T; \Sigma= 0.02 I$). The color  ranged from dark purple to light yellow indicates  the density inside the polyhedral cells. The density is shown in logarithm magnitude. The  edges colored in green indicate the boundaries of the RPM polyhedral cells with density below a threshold. As the probability thresholds (p) decreases, the relative volume (vol) of the reachable set decreases drastically. Our approach indicates under this distribution, the system state has large probability concentrating in the right bottom curve as shown in Fig.~\ref{fig:fwd-prob-1}(f).}
\label{fig:fwd-prob-1}
\end{figure}

\begin{figure}[!htbp]
\begin{subfigure}[b]{0.32\textwidth}
\includegraphics[width=1.0\textwidth]{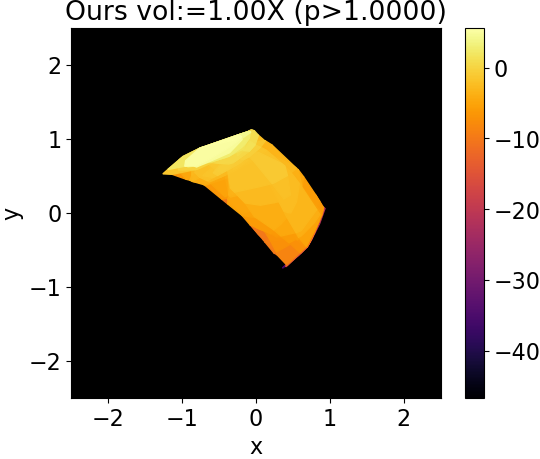} \hfill
\caption{vol=1.00X; p=1.0000}
\end{subfigure}
\begin{subfigure}[b]{0.32\textwidth}
\includegraphics[width=1.0\textwidth]{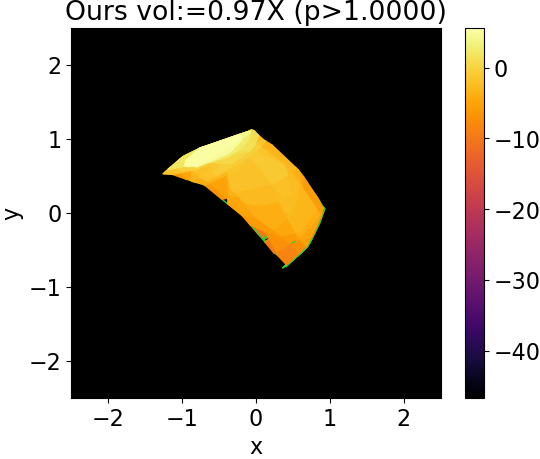} \hfill
\caption{vol=0.97X; p $>=$ 0.9999}
\end{subfigure}
\begin{subfigure}[b]{0.32\textwidth}
\includegraphics[width=1.0\textwidth]{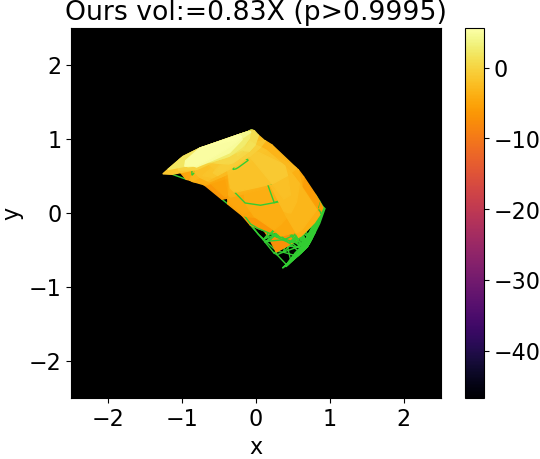} \hfill
\caption{vol=0.83X; p $>=$ 0.9995}
\end{subfigure}
\begin{subfigure}[b]{0.32\textwidth}
\includegraphics[width=1.0\textwidth]{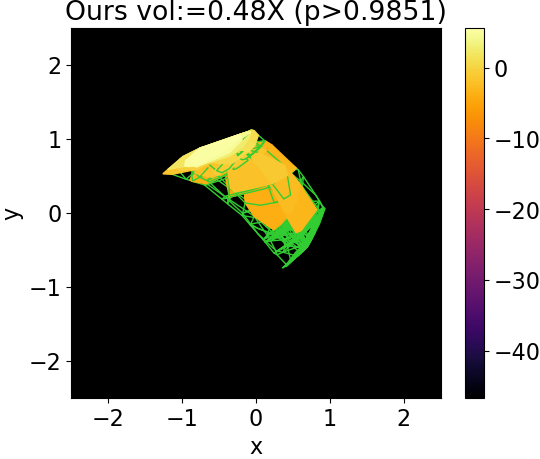} \hfill
\caption{vol=0.48X; p $>=$ 0.9851}
\end{subfigure}
\begin{subfigure}[b]{0.32\textwidth}
\includegraphics[width=1.0\textwidth]{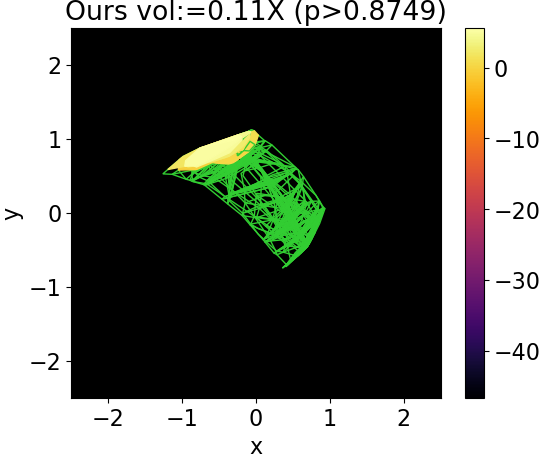} \hfill
\caption{vol=0.11X; p $>=$ 0.8749}
\end{subfigure}
\begin{subfigure}[b]{0.32\textwidth}
\includegraphics[width=1.0\textwidth]{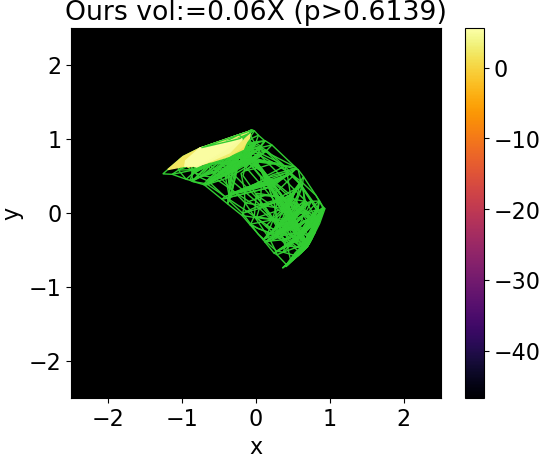} \hfill
\caption{vol=0.06X; p $>=$ 0.6139}
\end{subfigure}
\centering
\caption{The system forward reach set distribution under different probability thresholds at t=1.0s with initial condition $\mathcal{N}_2=N(\mu=(-1.7, -1.7, 1.3, 1.4)^T; \Sigma= 0.02 I)$. The color  ranged from dark purple to light yellow indicates  the density inside the polyhedral cells. The density is shown in logarithm magnitude. The  edges colored in green indicate the boundaries of the RPM polyhedral cells with density below a threshold. As the probability thresholds (p) decreases, the relative volume (vol) of the reachable set decreases drastically. Our approach indicates under this distribution, the system state has large probability concentrating in the top left curve as shown in Fig.~\ref{fig:fwd-prob-2}(f).}
\label{fig:fwd-prob-2}
\end{figure}

\begin{figure}[!htbp]
\begin{subfigure}[b]{0.32\textwidth}
\includegraphics[width=1.0\textwidth]{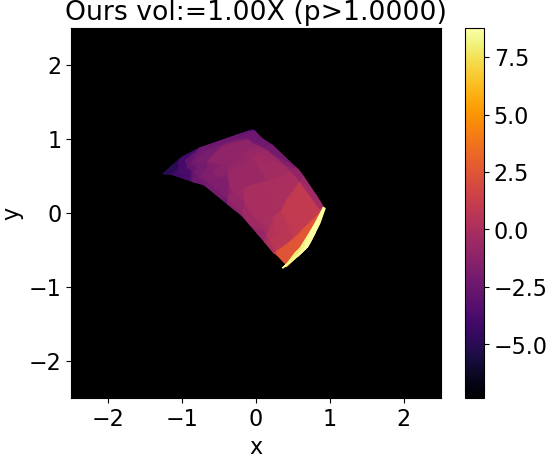} \hfill
\caption{vol=1.00X; p=1.0000}
\end{subfigure}
\begin{subfigure}[b]{0.32\textwidth}
\includegraphics[width=1.0\textwidth]{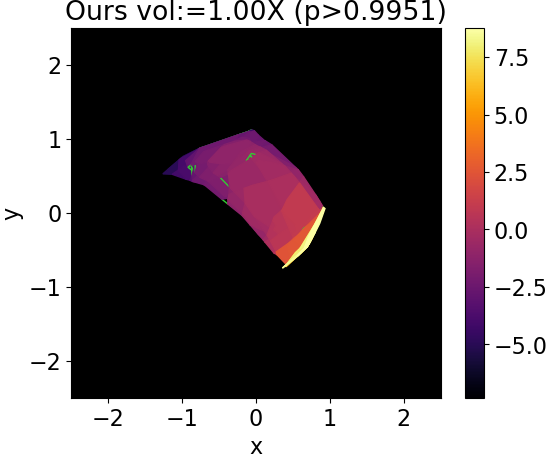} \hfill
\caption{vol=0.99X; p $>=$ 0.9951}
\end{subfigure}
\begin{subfigure}[b]{0.32\textwidth}
\includegraphics[width=1.0\textwidth]{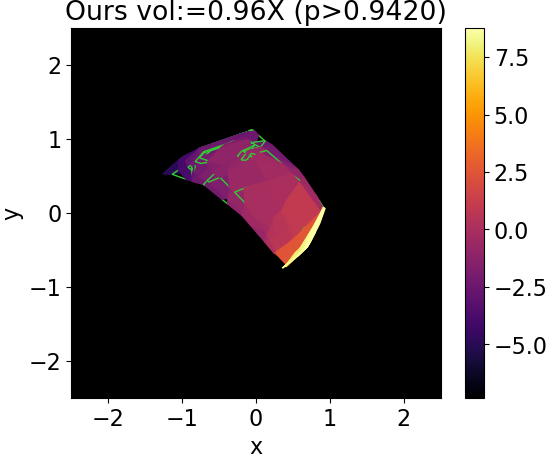} \hfill
\caption{vol=0.96X; p $>=$ 0.9420}
\end{subfigure}
\begin{subfigure}[b]{0.32\textwidth}
\includegraphics[width=1.0\textwidth]{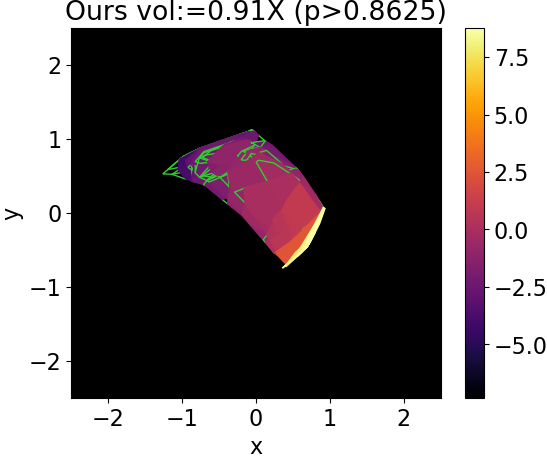} \hfill
\caption{vol=0.91X; p $>=$ 0.8625}
\end{subfigure}
\begin{subfigure}[b]{0.32\textwidth}
\includegraphics[width=1.0\textwidth]{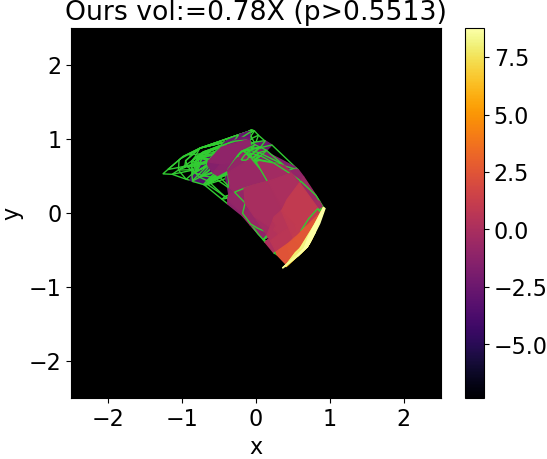} \hfill
\caption{vol=0.77X; p $>=$ 0.5513}
\end{subfigure}
\begin{subfigure}[b]{0.32\textwidth}
\includegraphics[width=1.0\textwidth]{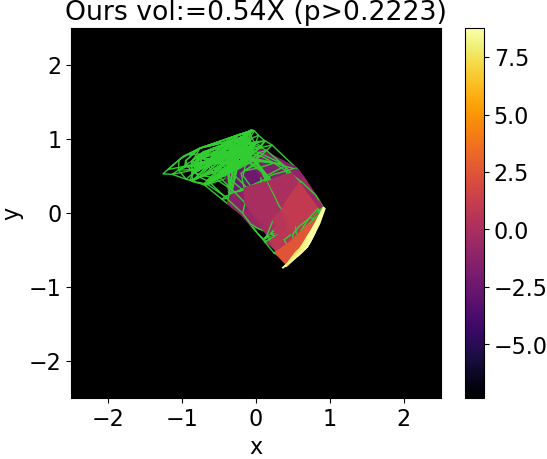} \hfill
\caption{vol=0.46X; p $>=$ 0.2223}
\end{subfigure}
\centering
\caption{The system forward reach set distribution under different probability thresholds at t=1.0s with initial condition $\mathcal{N}_3=N(\mu=(-1.7, -1.7, 0.2, 1.4)^T; \Sigma= 0.1 I)$. The color ranged from dark purple to light yellow indicates the density inside the polyhedral cells. The density is shown in logarithm magnitude. The  edges colored in green indicate the boundaries of the RPM polyhedral cells with density below a threshold. As the probability thresholds (p) decreases, the relative volume (vol) of the reachable set decreases drastically. Our approach indicates under this distribution, the system state has large probability concentrating in the right bottom curve as shown in Fig.~\ref{fig:fwd-prob-3}(f), but is not as concentrated as shown in Fig.~\ref{fig:fwd-prob-1}(f).}
\label{fig:fwd-prob-3}
\end{figure}

Instead of over-approximating the reachable sets like traditional methods, our approach can render varied sizes of reachable sets under different probability thresholds and under different initial state distributions. We compute for the varied reachable sets at t=1.0s for ground robot navigation experiment under three different (truncated) multivariate Gaussian distributions: $\mathcal{N}_1=\mathcal{N}(\mu=(-1.7, -1.7, 0.2, 1.4)^T; \Sigma= 0.02 I)$, $\mathcal{N}_2=\mathcal{N}(\mu=(-1.7, -1.7, 1.3, 1.4)^T; \Sigma= 0.02 I)$, and $\mathcal{N}_3=\mathcal{N}(\mu=(-1.7, -1.7, 0.2, 1.4)^T; \Sigma= 0.1 I)$. The difference between $\mathcal{N}_1$ and $\mathcal{N}_2$ is the change of the mean vector, and the difference between $\mathcal{N}_1$ and $\mathcal{N}_3$ is the change in the covariance matrix. As shown in Fig.~\ref{fig:fwd-prob-1}$\sim$Fig.~\ref{fig:fwd-prob-3}, as the probability threshold decreases, the relative volume of the reachable set (comparing to the volume in Fig.3(a)) decreases drastically. And our approach shows that under the initial distribution $\mathcal{N}_1$, a large portion of the states (p$>=$0.8980) actually only reside in a small region (vol=0.03X) in the state space (as shown in Fig.~\ref{fig:fwd-prob-1}(e)). Whereas under different initial state distributions, the concentration region might be different (comparing Fig.~\ref{fig:fwd-prob-1}(f) and Fig.~\ref{fig:fwd-prob-2}(f)) or the degree of concentration is different (comparing Fig.~\ref{fig:fwd-prob-1}(f) and Fig.~\ref{fig:fwd-prob-3}(e)).

\section{Runtime for fast safety checking}
\label{sec:e}
\begin{table*}[!htbp]
\begin{center}
\begin{tabular}{|c|c|c|c|c|c|c|}
\hline
&
\multicolumn{2}{|c|}{Low density} & \multicolumn{2}{|c|}{Medium density} & \multicolumn{2}{|c|}{High density} \\
&
\multicolumn{2}{|c|}{$e^{-10}\leq\rho\leq e^{2}$} & \multicolumn{2}{|c|}{$e^{2}\leq\rho\leq e^{3}$} & \multicolumn{2}{|c|}{$\rho\geq e^{3}$} \\
\hline
& Vanilla & Heuristic & Vanilla & Heuristic & Vanilla & Heuristic \\
\hline
Time (sec) & 3.1594 & 0.8425 & 3.0854 & 0.8122 & 3.0585 & 0.7644  \\
\#(Rect) &  -& 391 & -& 31 & - & 4\\
\#(Poly) & 303 & 303 & 2 & 2 & 0 &  0\\
Is safe? & No & No & No & No & Yes & Yes \\
\hline
\end{tabular}
\end{center}
\caption{Online safe verification comparison under different density conditions (Low / Medium / High). We measure the computation time (``Time""), number of rectangle intersections (``\#(Rect)"), number of polyhedral intersections (``\#(Poly)") and whether the initial condition will avoid to drive to unsafe region (``Is safe?") under each density condition with and without using the hyper-rectangle heuristics (``Heuristic"/``Vanilla"). As shown in Table.~\ref{table:1}, the trajectories sampled from the initial state will only reach the unsafe region under low and medium densities while won't reach the unsafe region in high density. Using heuristics can reduce the computation time in all conditions by ~70\%.}
\label{table:1}
\end{table*}

We also perform the system safety verification for the ground robot task. Specifically, we want to verify whether the trajectories starting from the initial condition $S_{init}$ will drive to the unsafe region $S_{unsafe}$ under different density conditions. We set $S_{init}=\{-1.8\leq x\leq-1.2,\,-1.8\leq y\leq -1.2, 0.0\leq \theta \leq \pi/2, 0.0\leq v \leq 1.0\}$, $S_{unsafe}=\{-0.5\leq x\leq 0.0, -0.5\leq y\leq 0.0\}$ and try three different density constraints: which are low density ($e^{-10}\leq \rho\leq e^2$), medium density ($e^2\leq \rho\leq e^3$) and high density ($\rho\geq e^3$). We measure whether the initial condition will avoid to lead the system to reach the unsafe region under each density condition (``Is safe?"). To illustrate how the heuristic method introduced in Sec.~\ref{sec:b3} accelerates the computation process, we also measure the computation time (``Time""), number of rectangle intersections (``\#(Rect)"), number of polyhedral intersections (``\#(Poly)") and , with and without using the hyper-rectangle heuristics(``Heuristic"/``Vanilla"). Our program is implemented in Python with parallel computation deployed on a 12-core CPU. 

As shown in Table.~\ref{table:1}, the trajectories sampled from the initial state will only reach the unsafe region under low and medium densities,  and won't reach the unsafe region in high density. This can be helpful when we are considering planning problems with density constraints. Besides, our approach with hyper-rectangle heuristic can finish the online safety verification for 50 time steps in only 0.8 seconds, which reduces ~70\% of the computation time comparing to the vanilla algorithm. Doing safety verification for each time step only needs $0.016s$, which is much smaller than the actual $\Delta t$ used for the ground robot navigation benchmark ($\Delta t=0.05s$). With code-level optimization (e.g. write the program in C++ or Julia) and more CPU cores being used in parallel, our approach can further benefit for real-time applications.
\section{Density (Ours, KDE, histogram, groundtruth) and reachability visualizations}
\label{sec:f}

Here we compare the density prediction results on all 10 benchmark examples mentioned in Table~\ref{table:density}, and compare our reachable set result with other worst-case reachability tools (Convex Hull~\citep{lew2020sampling}, GSG~\citep{everett2020robustness} and DryVR~\citep{fan2020fast}) on 4 of the benchmark examples. As shown in figures in \ref{sec:f1}, our approach can consistently achieve the closest state density distribution among other approaches (Kernel density, histogram), and doesn't have a restriction for high-dimension systems (whereas the histogram method cannot estimate the density for high-dimension systems like in Fig.~\ref{fig:dens_e07_acc_000} $\sim$ Fig.~\ref{fig:dens_e09_toon_049}). For the reachability comparison, different from the worst-case reachability analysis tools (Convex Hull~\citep{lew2020sampling}, GSG~\citep{everett2020robustness} and DryVR~\citep{fan2020fast}), our approach can compute the density and probability for each of the reachable set, hence is able to tell where do states concentrate (a high probability of states only reside in a small region in the state space, as shown in Fig.~\ref{fig:reach_e00_vdp_049}, Fig.~\ref{fig:reach_e01_dint_007}, Fig.~\ref{fig:reach_e04_robot_020},  Fig.~\ref{fig:reach_e05_car_049}, etc). Our method is more precise and informative than those worst-case reachability analysis approaches. More figures can be found out in the supplementary video.

\subsection{Comparison of density prediction accuracies}
\label{sec:f1}
\begin{figure}[!htbp]
\begin{subfigure}[b]{0.190\textwidth}
\includegraphics[width=1.0\textwidth]{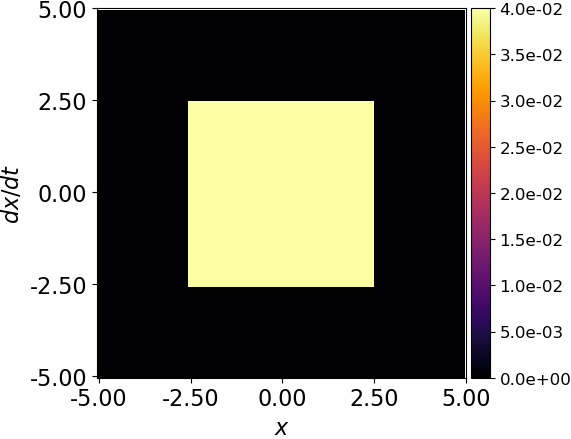} \hfill
\caption{Groundtruth}
\end{subfigure}
\begin{subfigure}[b]{0.190\textwidth}
\includegraphics[width=1.0\textwidth]{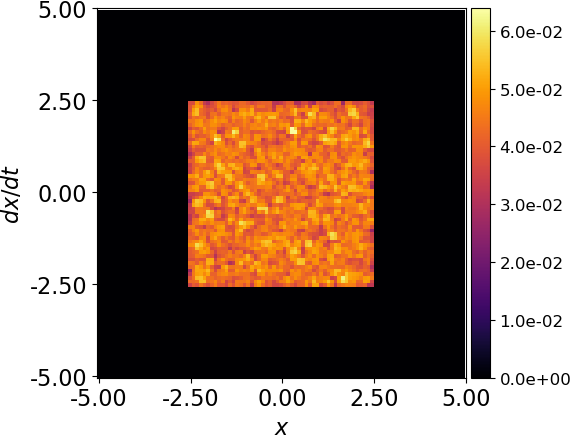} \hfill
\caption{Kernel density}
\end{subfigure}
\begin{subfigure}[b]{0.190\textwidth}
\includegraphics[width=1.0\textwidth]{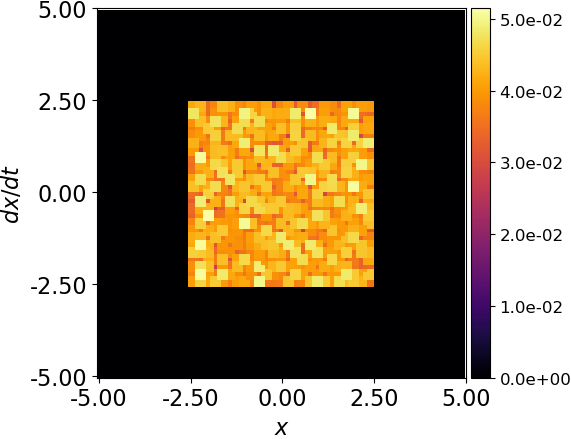} \hfill
\caption{Histogram}
\end{subfigure}
\begin{subfigure}[b]{0.190\textwidth}
\includegraphics[width=1.0\textwidth]{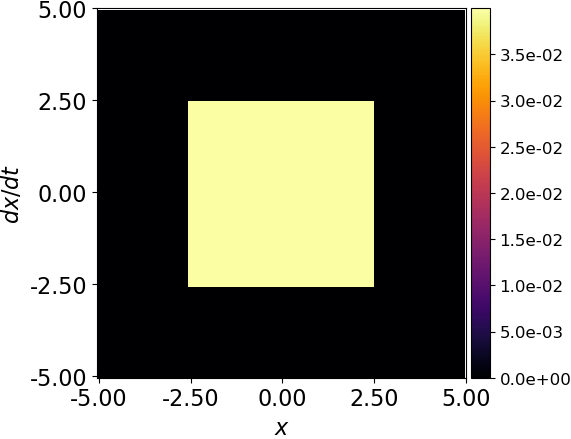} \hfill
\caption{Ours}
\end{subfigure}
\begin{subfigure}[b]{0.190\textwidth}
\includegraphics[width=1.0\textwidth]{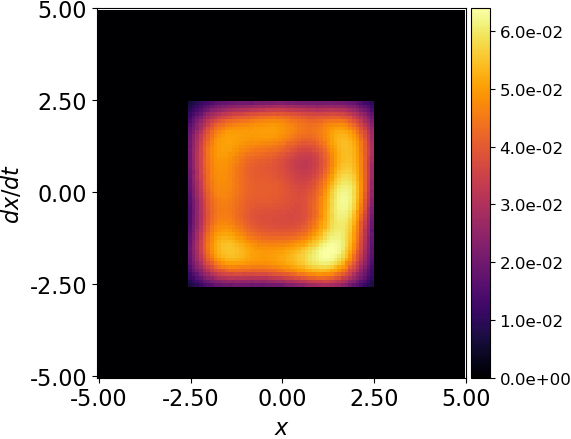} \hfill
\caption{SGPD}
\end{subfigure}
\centering
\caption{Comparison of density prediction accuracies (Van der Pol, t=0)}
\label{fig:dens_e00_vdp_000}
\end{figure}

\begin{figure}[!htbp]
\begin{subfigure}[b]{0.190\textwidth}
\includegraphics[width=1.0\textwidth]{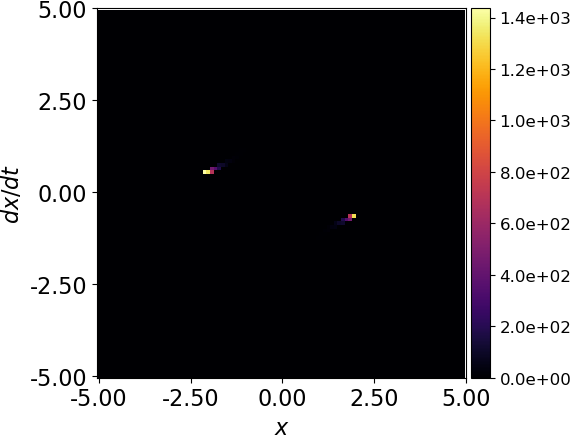} \hfill
\caption{Groundtruth}
\end{subfigure}
\begin{subfigure}[b]{0.190\textwidth}
\includegraphics[width=1.0\textwidth]{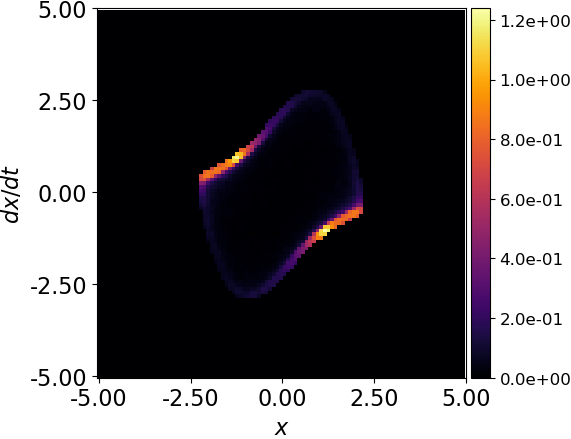} \hfill
\caption{Kernel density}
\end{subfigure}
\begin{subfigure}[b]{0.190\textwidth}
\includegraphics[width=1.0\textwidth]{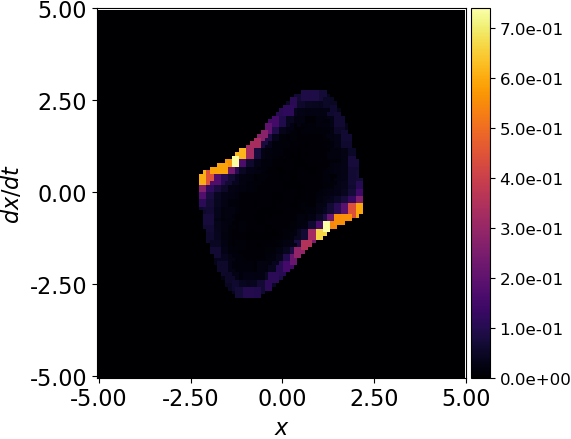} \hfill
\caption{Histogram}
\end{subfigure}
\begin{subfigure}[b]{0.190\textwidth}
\includegraphics[width=1.0\textwidth]{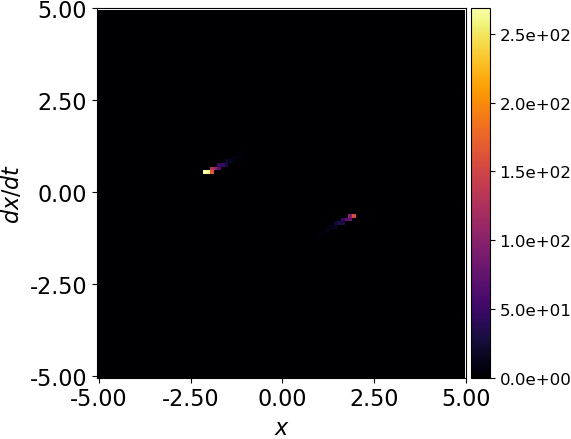} \hfill
\caption{Ours}
\end{subfigure}
\begin{subfigure}[b]{0.190\textwidth}
\includegraphics[width=1.0\textwidth]{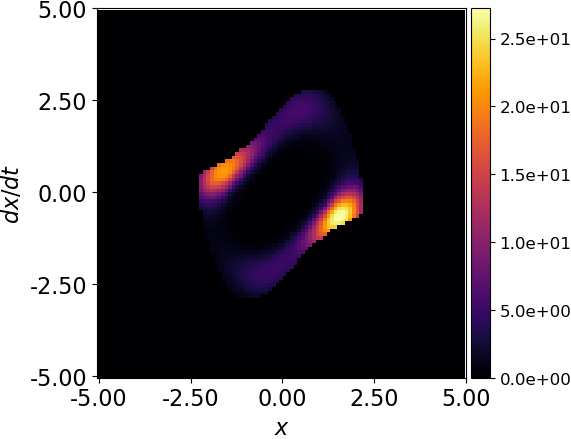} \hfill
\caption{SGPD}
\end{subfigure}
\centering
\caption{Comparison of density prediction accuracies (Van der Pol, t=20)}
\label{fig:dens_e00_vdp_020}
\end{figure}

\begin{figure}[!htbp]
\begin{subfigure}[b]{0.190\textwidth}
\includegraphics[width=1.0\textwidth]{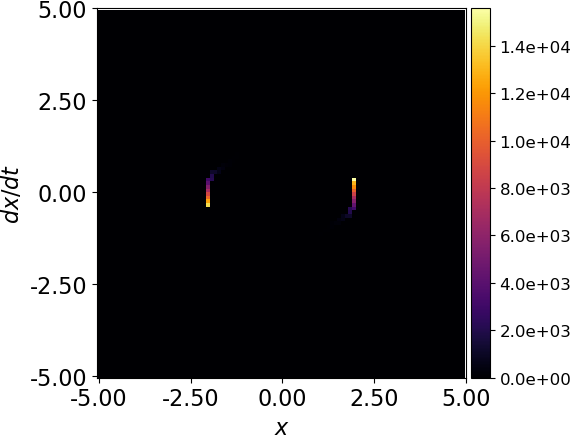} \hfill
\caption{Groundtruth}
\end{subfigure}
\begin{subfigure}[b]{0.190\textwidth}
\includegraphics[width=1.0\textwidth]{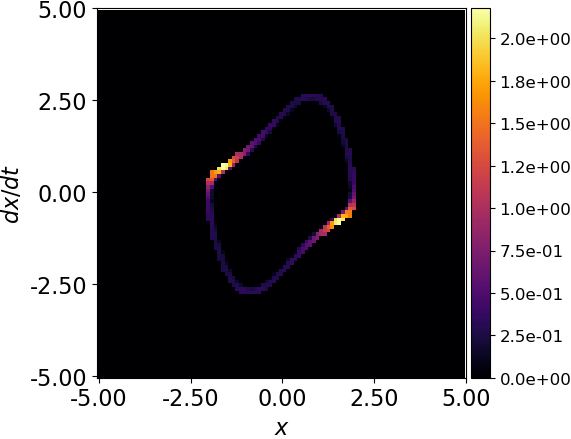} \hfill
\caption{Kernel density}
\end{subfigure}
\begin{subfigure}[b]{0.190\textwidth}
\includegraphics[width=1.0\textwidth]{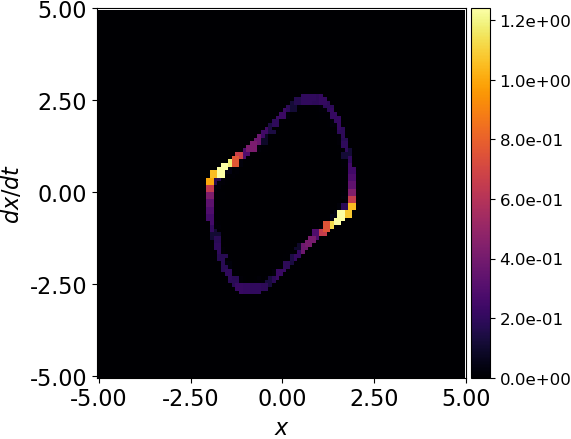} \hfill
\caption{Histogram}
\end{subfigure}
\begin{subfigure}[b]{0.190\textwidth}
\includegraphics[width=1.0\textwidth]{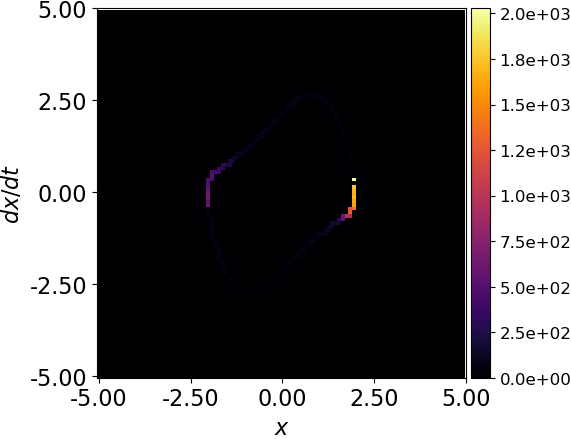} \hfill
\caption{Ours}
\end{subfigure}
\begin{subfigure}[b]{0.190\textwidth}
\includegraphics[width=1.0\textwidth]{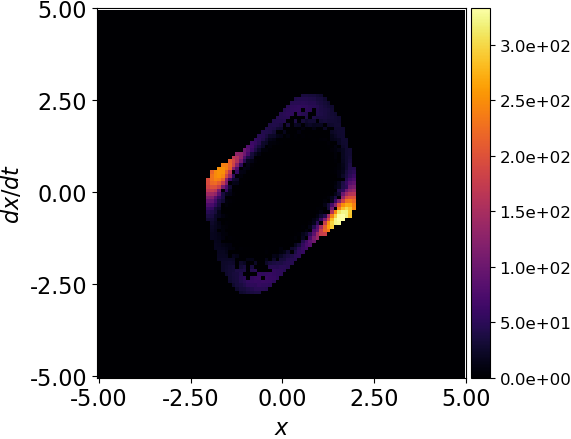} \hfill
\caption{SGPD}
\end{subfigure}
\centering
\caption{Comparison of density prediction accuracies (Van der Pol, t=49)}
\label{fig:dens_e00_vdp_049}
\end{figure}

\begin{figure}[!htbp]
\begin{subfigure}[b]{0.190\textwidth}
\includegraphics[width=1.0\textwidth]{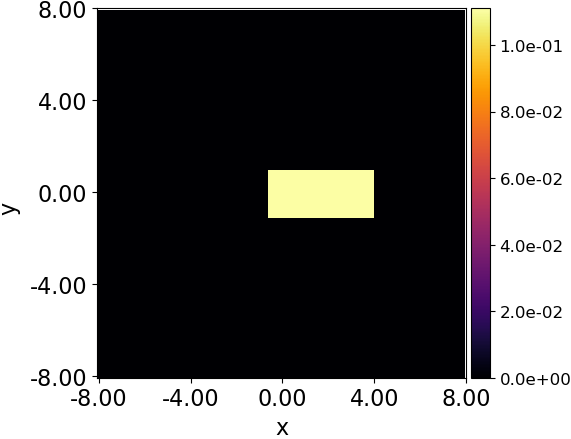} \hfill
\caption{Groundtruth}
\end{subfigure}
\begin{subfigure}[b]{0.190\textwidth}
\includegraphics[width=1.0\textwidth]{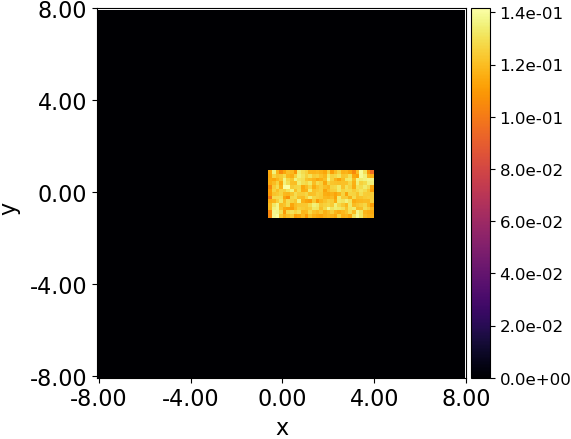} \hfill
\caption{Kernel density}
\end{subfigure}
\begin{subfigure}[b]{0.190\textwidth}
\includegraphics[width=1.0\textwidth]{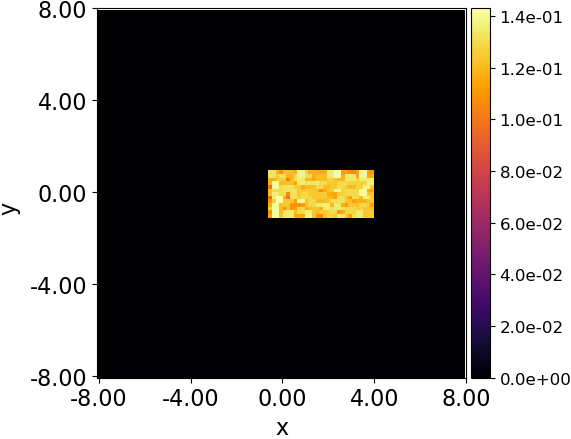} \hfill
\caption{Histogram}
\end{subfigure}
\begin{subfigure}[b]{0.190\textwidth}
\includegraphics[width=1.0\textwidth]{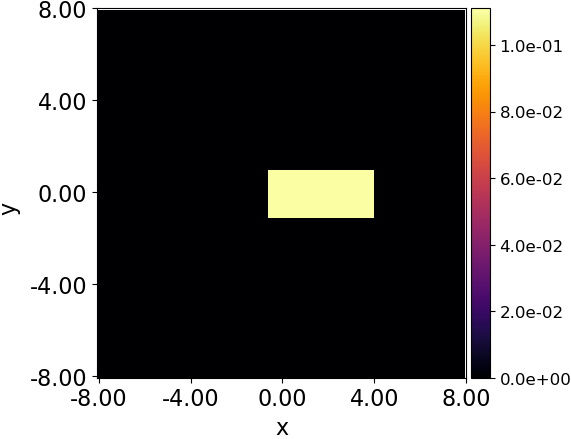} \hfill
\caption{Ours}
\end{subfigure}
\begin{subfigure}[b]{0.190\textwidth}
\includegraphics[width=1.0\textwidth]{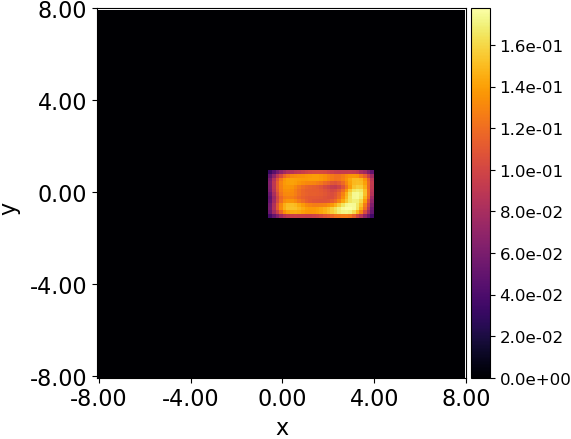} \hfill
\caption{SGPD}
\end{subfigure}
\centering
\caption{Comparison of density prediction accuracies (Double integrator, t=0)}
\label{fig:dens_e01_dint_000}
\end{figure}

\begin{figure}[!htbp]
\begin{subfigure}[b]{0.190\textwidth}
\includegraphics[width=1.0\textwidth]{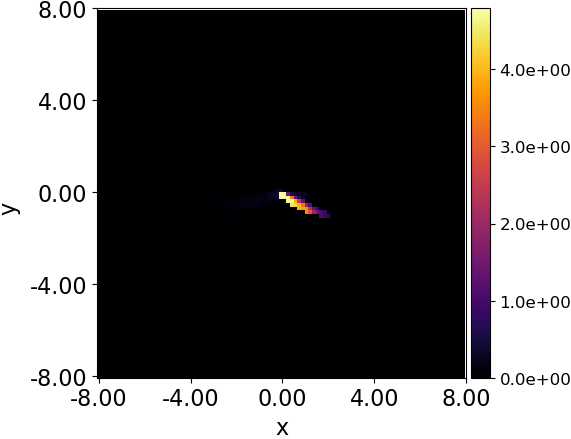} \hfill
\caption{Groundtruth}
\end{subfigure}
\begin{subfigure}[b]{0.190\textwidth}
\includegraphics[width=1.0\textwidth]{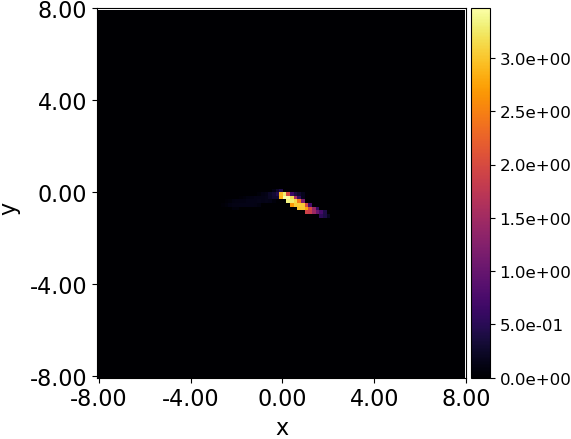} \hfill
\caption{Kernel density}
\end{subfigure}
\begin{subfigure}[b]{0.190\textwidth}
\includegraphics[width=1.0\textwidth]{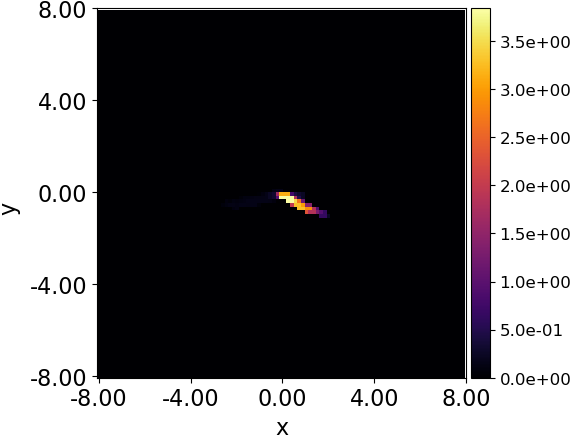} \hfill
\caption{Histogram}
\end{subfigure}
\begin{subfigure}[b]{0.190\textwidth}
\includegraphics[width=1.0\textwidth]{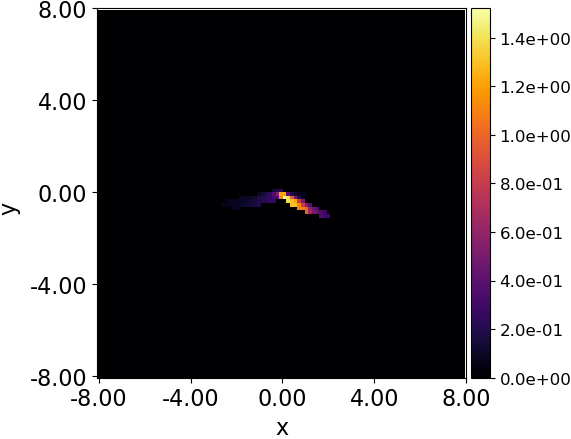} \hfill
\caption{Ours}
\end{subfigure}
\begin{subfigure}[b]{0.190\textwidth}
\includegraphics[width=1.0\textwidth]{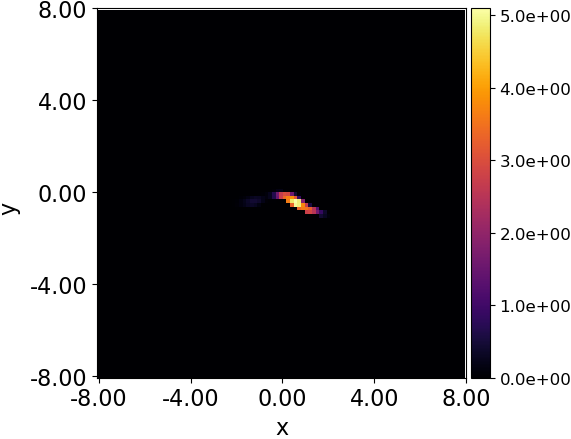} \hfill
\caption{SGPD}
\end{subfigure}
\centering
\caption{Comparison of density prediction accuracies (Double integrator, t=3)}
\label{fig:dens_e01_dint_003}
\end{figure}

\begin{figure}[!htbp]
\begin{subfigure}[b]{0.190\textwidth}
\includegraphics[width=1.0\textwidth]{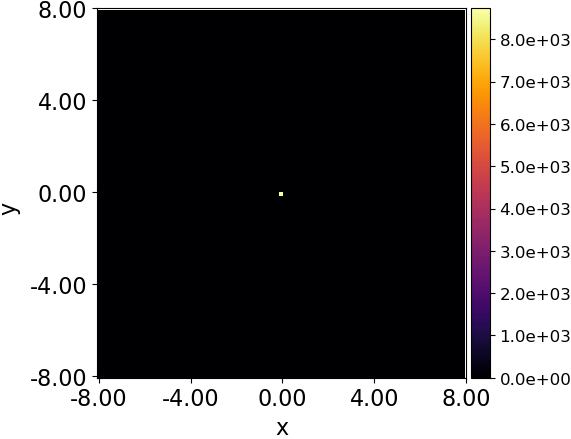} \hfill
\caption{Groundtruth}
\end{subfigure}
\begin{subfigure}[b]{0.190\textwidth}
\includegraphics[width=1.0\textwidth]{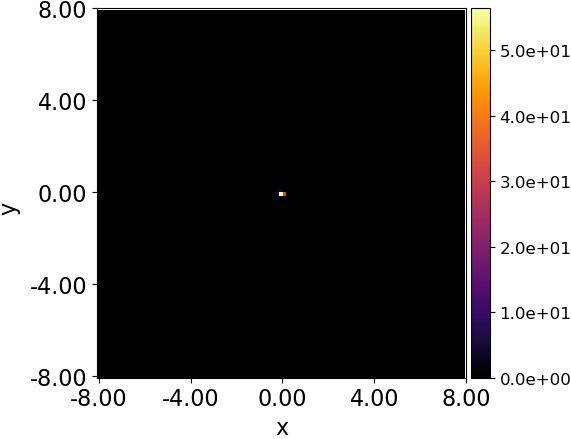} \hfill
\caption{Kernel density}
\end{subfigure}
\begin{subfigure}[b]{0.190\textwidth}
\includegraphics[width=1.0\textwidth]{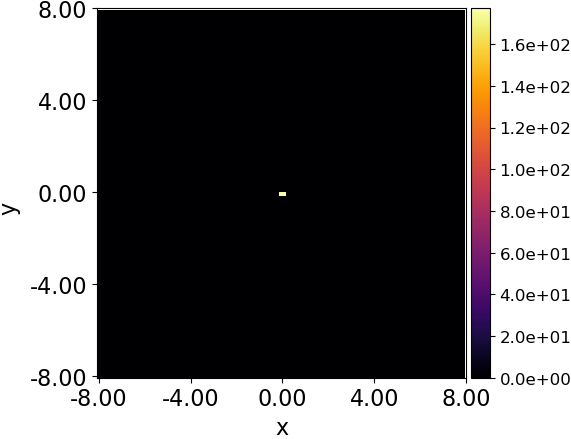} \hfill
\caption{Histogram}
\end{subfigure}
\begin{subfigure}[b]{0.190\textwidth}
\includegraphics[width=1.0\textwidth]{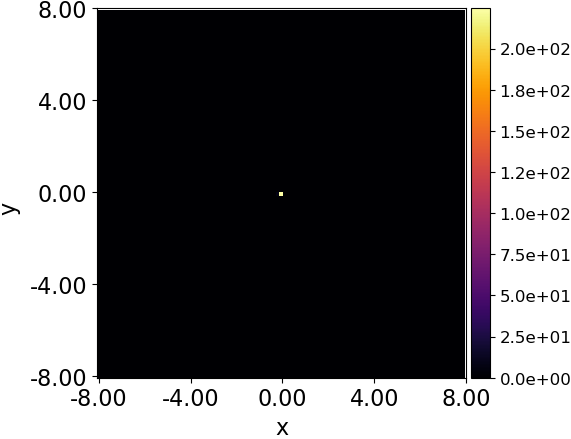} \hfill
\caption{Ours}
\end{subfigure}
\begin{subfigure}[b]{0.190\textwidth}
\includegraphics[width=1.0\textwidth]{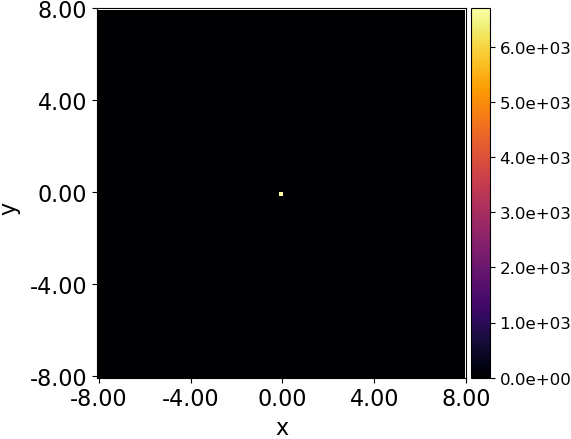} \hfill
\caption{SGPD}
\end{subfigure}
\centering
\caption{Comparison of density prediction accuracies (Double integrator, t=9)}
\label{fig:dens_e01_dint_009}
\end{figure}

\begin{figure}[!htbp]
\begin{subfigure}[b]{0.190\textwidth}
\includegraphics[width=1.0\textwidth]{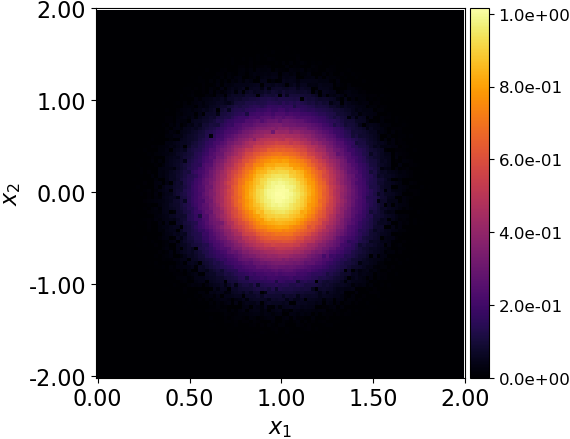} \hfill
\caption{Groundtruth}
\end{subfigure}
\begin{subfigure}[b]{0.190\textwidth}
\includegraphics[width=1.0\textwidth]{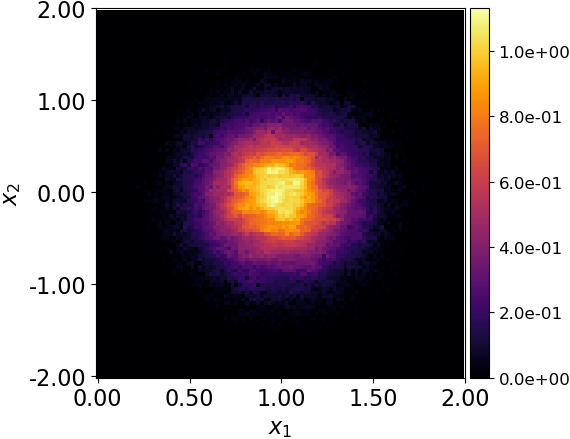} \hfill
\caption{Kernel density}
\end{subfigure}
\begin{subfigure}[b]{0.190\textwidth}
\includegraphics[width=1.0\textwidth]{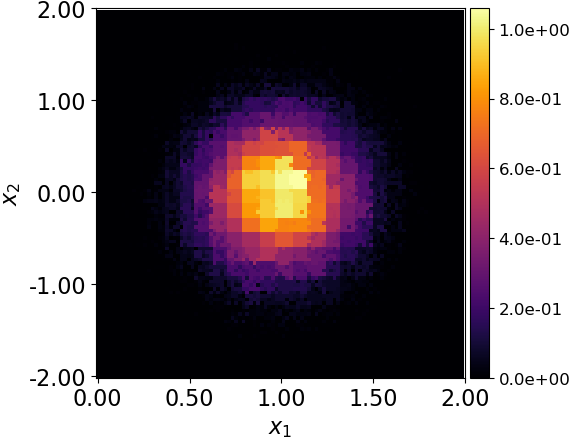} \hfill
\caption{Histogram}
\end{subfigure}
\begin{subfigure}[b]{0.190\textwidth}
\includegraphics[width=1.0\textwidth]{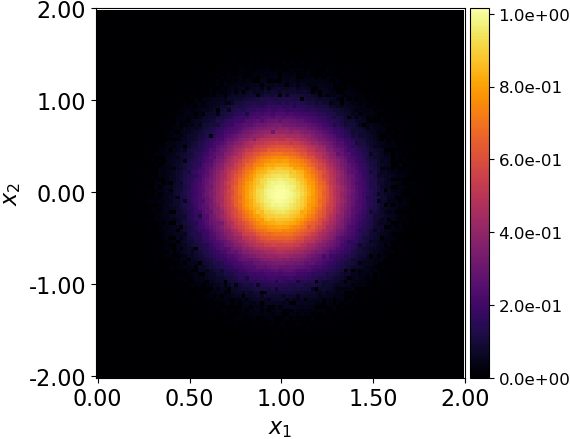} \hfill
\caption{Ours}
\end{subfigure}
\begin{subfigure}[b]{0.190\textwidth}
\includegraphics[width=1.0\textwidth]{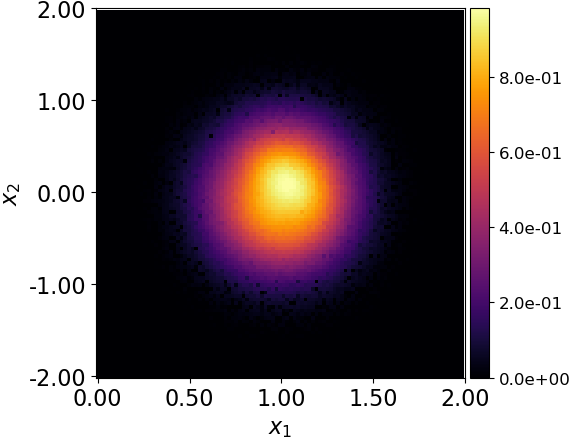} \hfill
\caption{SGPD}
\end{subfigure}
\centering
\caption{Comparison of density prediction accuracies (Kraichnan-Orszag system, t=0)}
\label{fig:dens_e02_kop_000}
\end{figure}

\begin{figure}[!htbp]
\begin{subfigure}[b]{0.190\textwidth}
\includegraphics[width=1.0\textwidth]{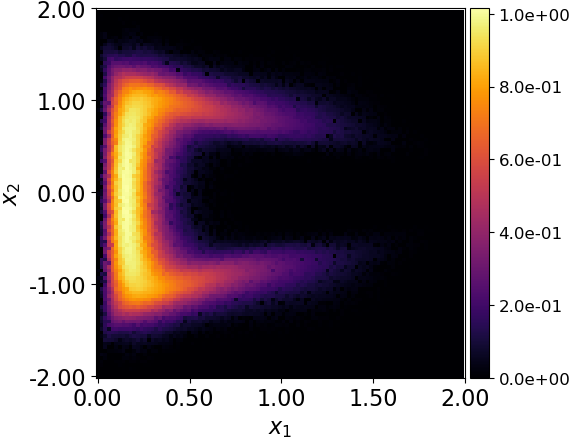} \hfill
\caption{Groundtruth}
\end{subfigure}
\begin{subfigure}[b]{0.190\textwidth}
\includegraphics[width=1.0\textwidth]{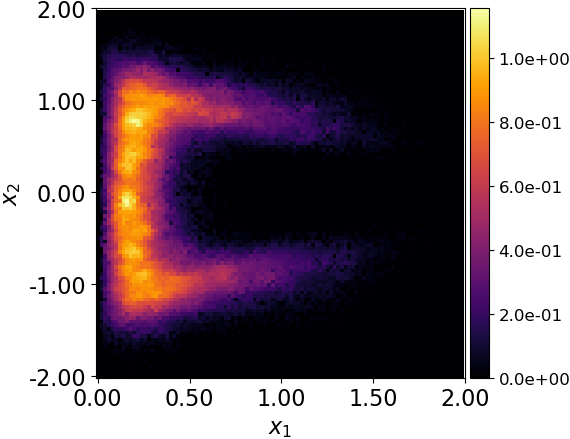} \hfill
\caption{Kernel density}
\end{subfigure}
\begin{subfigure}[b]{0.190\textwidth}
\includegraphics[width=1.0\textwidth]{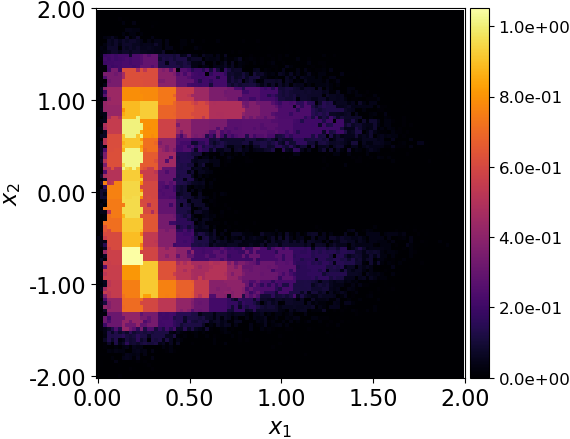} \hfill
\caption{Histogram}
\end{subfigure}
\begin{subfigure}[b]{0.190\textwidth}
\includegraphics[width=1.0\textwidth]{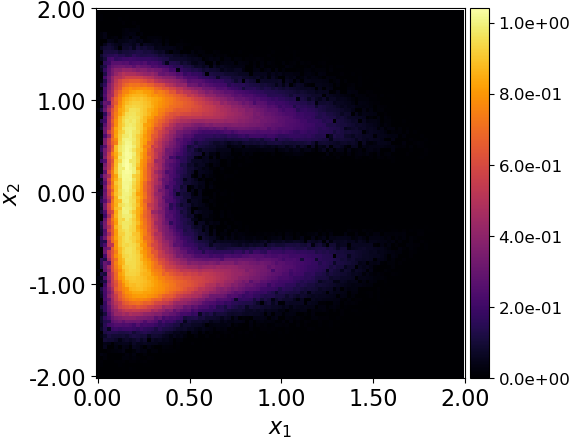} \hfill
\caption{Ours}
\end{subfigure}
\begin{subfigure}[b]{0.190\textwidth}
\includegraphics[width=1.0\textwidth]{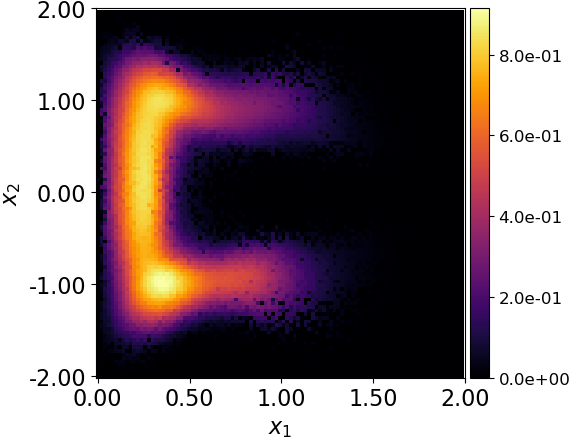} \hfill
\caption{SGPD}
\end{subfigure}
\centering
\caption{Comparison of density prediction accuracies (Kraichnan-Orszag system, t=20)}
\label{fig:dens_e02_kop_020}
\end{figure}

\begin{figure}[!htbp]
\begin{subfigure}[b]{0.190\textwidth}
\includegraphics[width=1.0\textwidth]{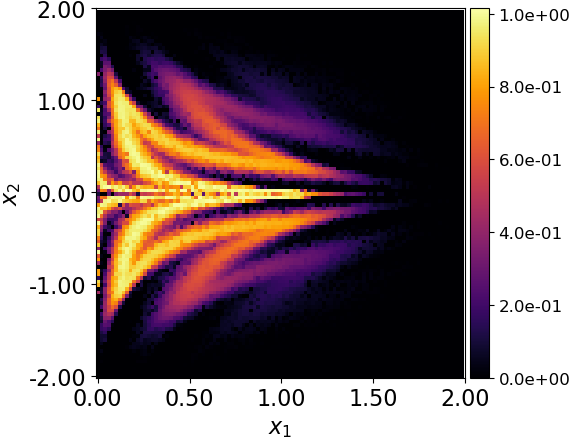} \hfill
\caption{Groundtruth}
\end{subfigure}
\begin{subfigure}[b]{0.190\textwidth}
\includegraphics[width=1.0\textwidth]{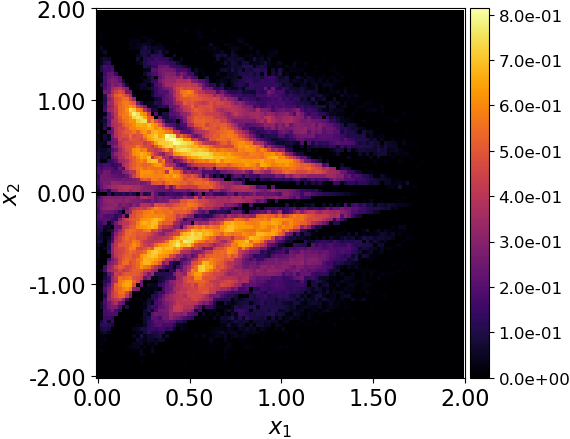} \hfill
\caption{Kernel density}
\end{subfigure}
\begin{subfigure}[b]{0.190\textwidth}
\includegraphics[width=1.0\textwidth]{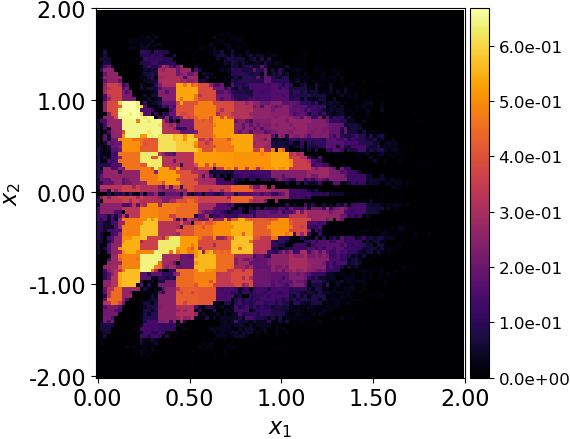} \hfill
\caption{Histogram}
\end{subfigure}
\begin{subfigure}[b]{0.190\textwidth}
\includegraphics[width=1.0\textwidth]{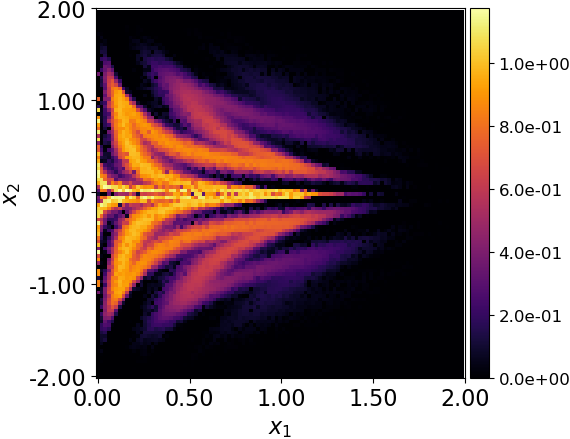} \hfill
\caption{Ours}
\end{subfigure}
\begin{subfigure}[b]{0.190\textwidth}
\includegraphics[width=1.0\textwidth]{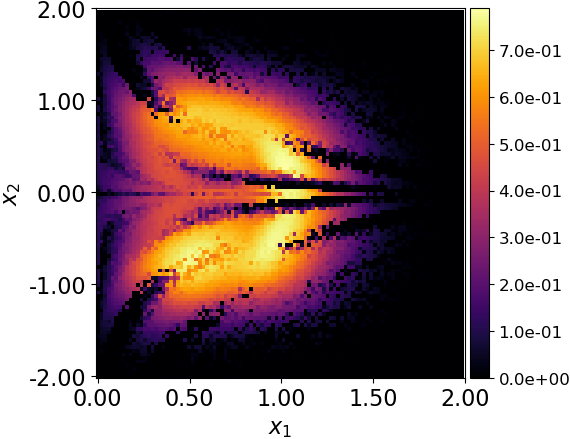} \hfill
\caption{SGPD}
\end{subfigure}
\centering
\caption{Comparison of density prediction accuracies (Kraichnan-Orszag system, t=79)}
\label{fig:dens_e02_kop_079}
\end{figure}

\begin{figure}[!htbp]
\begin{subfigure}[b]{0.190\textwidth}
\includegraphics[width=1.0\textwidth]{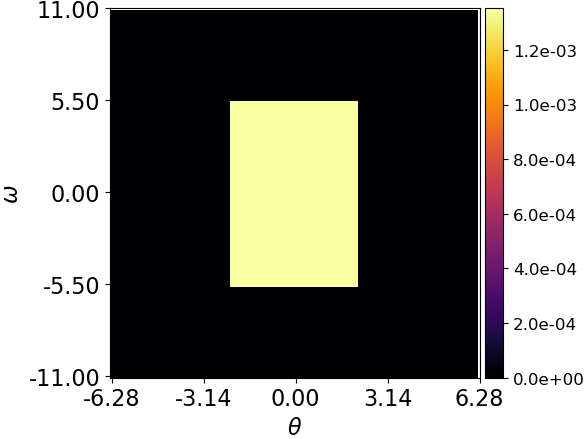} \hfill
\caption{Groundtruth}
\end{subfigure}
\begin{subfigure}[b]{0.190\textwidth}
\includegraphics[width=1.0\textwidth]{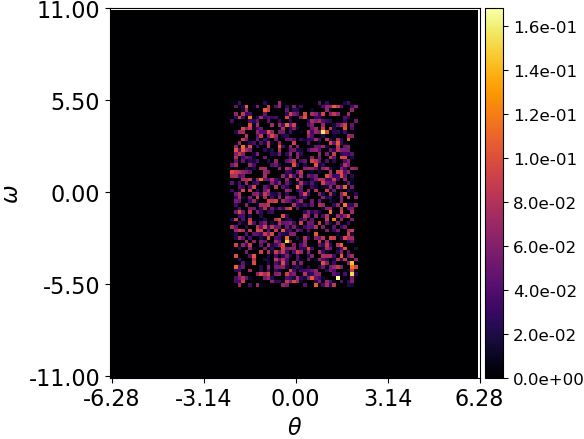} \hfill
\caption{Kernel density}
\end{subfigure}
\begin{subfigure}[b]{0.190\textwidth}
\includegraphics[width=1.0\textwidth]{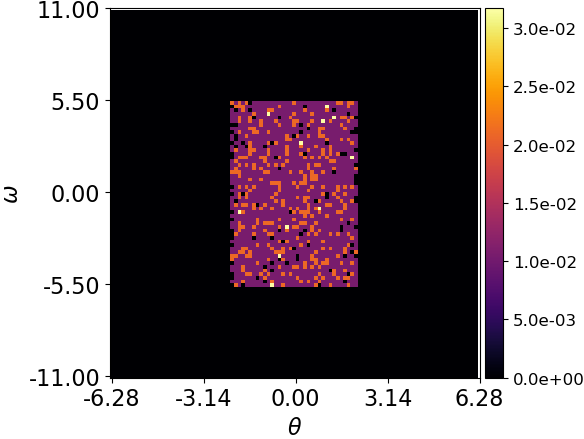} \hfill
\caption{Histogram}
\end{subfigure}
\begin{subfigure}[b]{0.190\textwidth}
\includegraphics[width=1.0\textwidth]{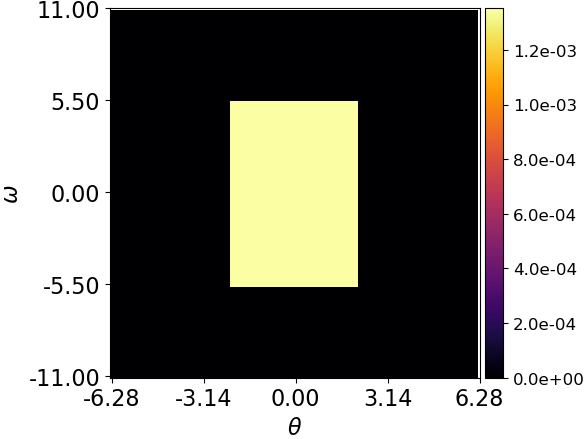} \hfill
\caption{Ours}
\end{subfigure}
\begin{subfigure}[b]{0.190\textwidth}
\includegraphics[width=1.0\textwidth]{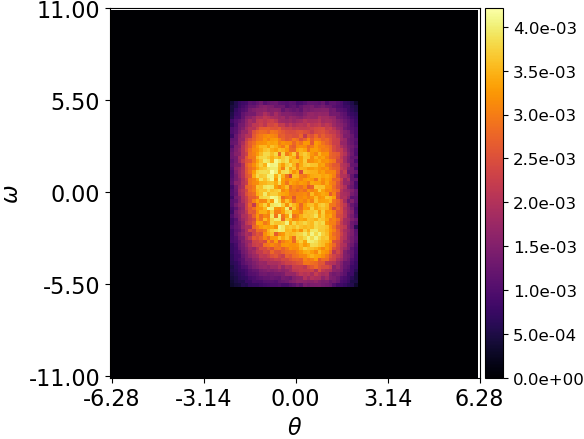} \hfill
\caption{SGPD}
\end{subfigure}
\centering
\caption{Comparison of density prediction accuracies (Inverted pendulum, t=0)}
\label{fig:dens_e03_pend_000}
\end{figure}

\begin{figure}[!htbp]
\begin{subfigure}[b]{0.190\textwidth}
\includegraphics[width=1.0\textwidth]{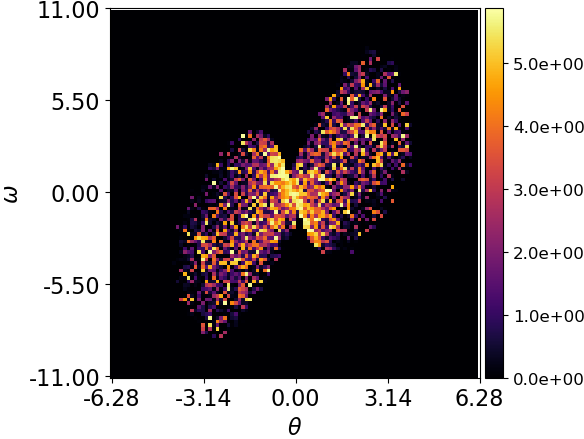} \hfill
\caption{Groundtruth}
\end{subfigure}
\begin{subfigure}[b]{0.190\textwidth}
\includegraphics[width=1.0\textwidth]{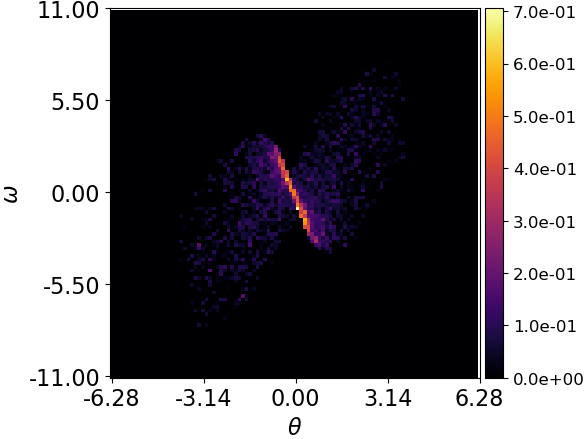} \hfill
\caption{Kernel density}
\end{subfigure}
\begin{subfigure}[b]{0.190\textwidth}
\includegraphics[width=1.0\textwidth]{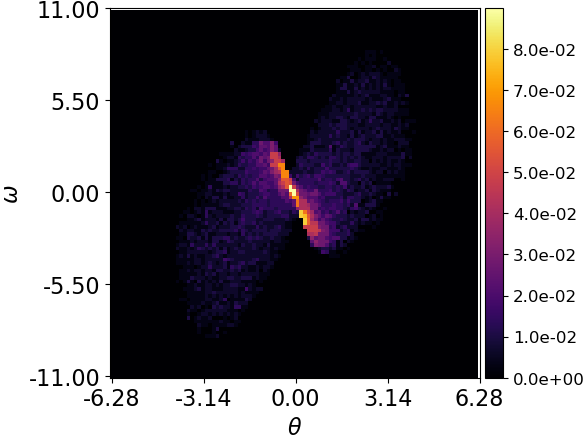} \hfill
\caption{Histogram}
\end{subfigure}
\begin{subfigure}[b]{0.190\textwidth}
\includegraphics[width=1.0\textwidth]{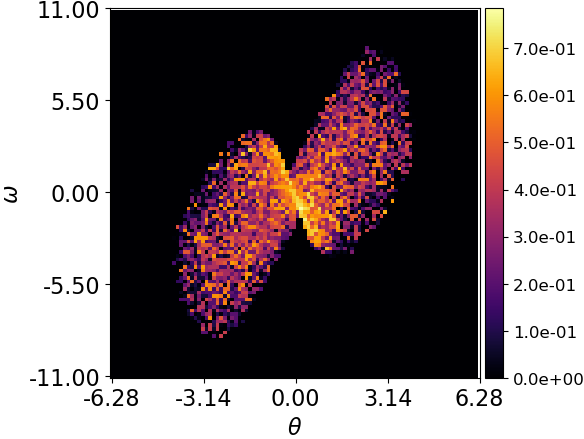} \hfill
\caption{Ours}
\end{subfigure}
\begin{subfigure}[b]{0.190\textwidth}
\includegraphics[width=1.0\textwidth]{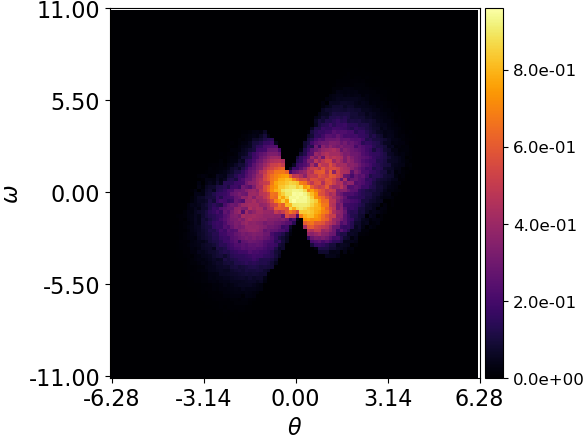} \hfill
\caption{SGPD}
\end{subfigure}
\centering
\caption{Comparison of density prediction accuracies (Inverted pendulum, t=20)}
\label{fig:dens_e03_pend_020}
\end{figure}

\begin{figure}[!htbp]
\begin{subfigure}[b]{0.190\textwidth}
\includegraphics[width=1.0\textwidth]{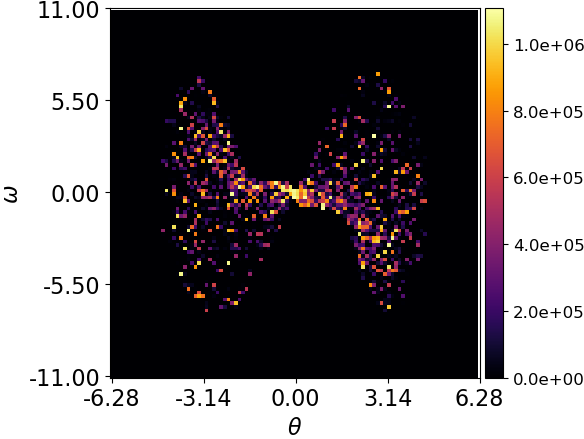} \hfill
\caption{Groundtruth}
\end{subfigure}
\begin{subfigure}[b]{0.190\textwidth}
\includegraphics[width=1.0\textwidth]{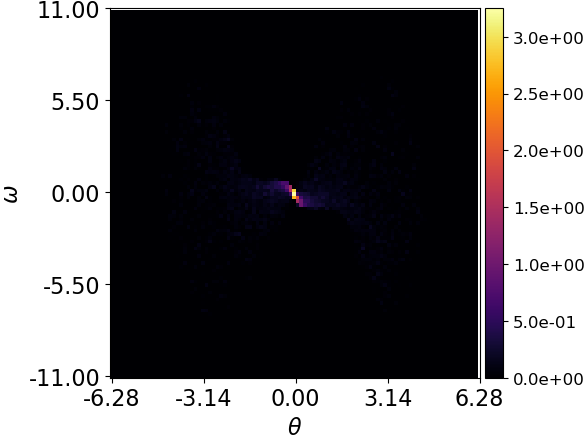} \hfill
\caption{Kernel density}
\end{subfigure}
\begin{subfigure}[b]{0.190\textwidth}
\includegraphics[width=1.0\textwidth]{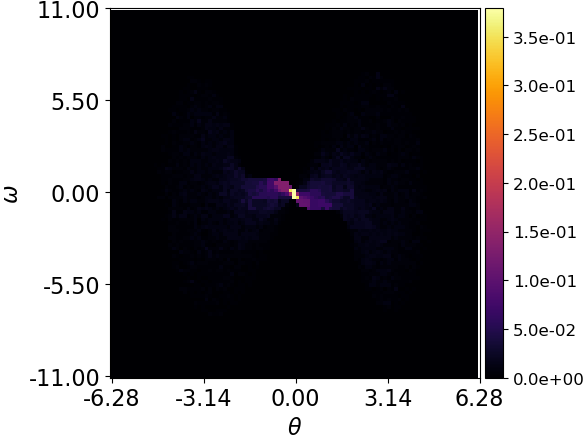} \hfill
\caption{Histogram}
\end{subfigure}
\begin{subfigure}[b]{0.190\textwidth}
\includegraphics[width=1.0\textwidth]{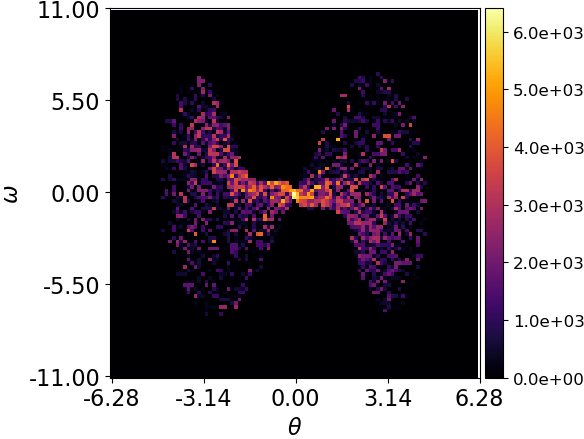} \hfill
\caption{Ours}
\end{subfigure}
\begin{subfigure}[b]{0.190\textwidth}
\includegraphics[width=1.0\textwidth]{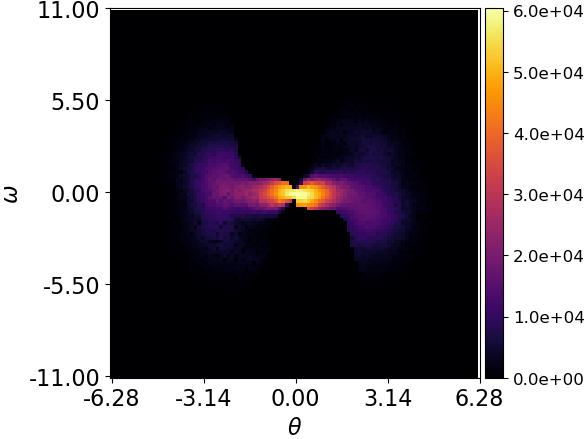} \hfill
\caption{SGPD}
\end{subfigure}
\centering
\caption{Comparison of density prediction accuracies (Inverted pendulum, t=49)}
\label{fig:dens_e03_pend_049}
\end{figure}

\begin{figure}[!htbp]
\begin{subfigure}[b]{0.190\textwidth}
\includegraphics[width=1.0\textwidth]{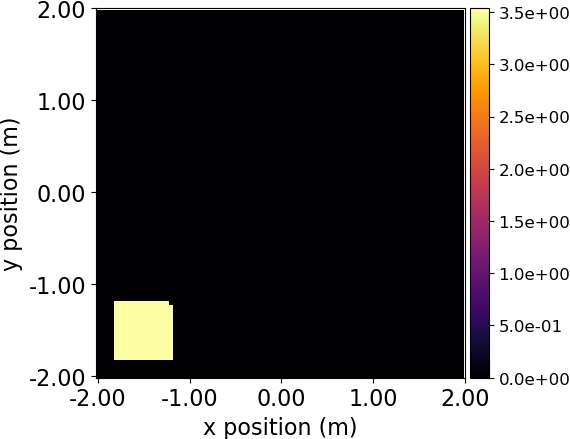} \hfill
\caption{Groundtruth}
\end{subfigure}
\begin{subfigure}[b]{0.190\textwidth}
\includegraphics[width=1.0\textwidth]{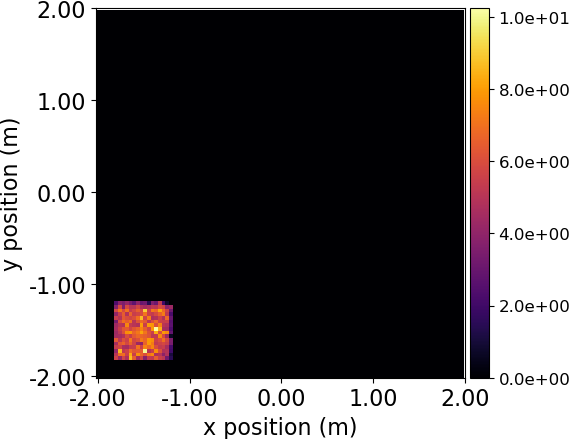} \hfill
\caption{Kernel density}
\end{subfigure}
\begin{subfigure}[b]{0.190\textwidth}
\includegraphics[width=1.0\textwidth]{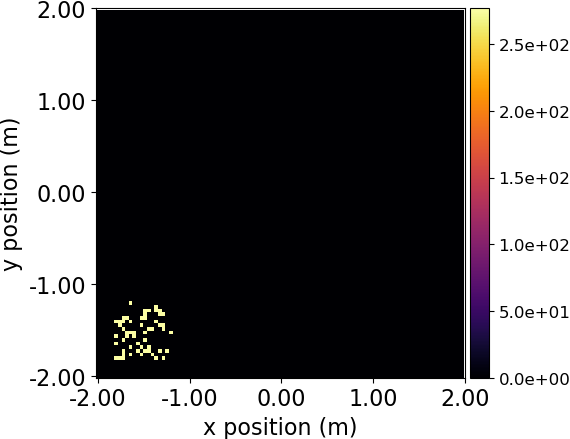} \hfill
\caption{Histogram}
\end{subfigure}
\begin{subfigure}[b]{0.190\textwidth}
\includegraphics[width=1.0\textwidth]{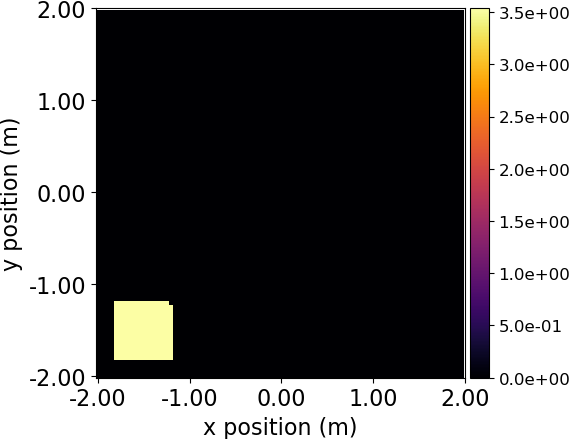} \hfill
\caption{Ours}
\end{subfigure}
\begin{subfigure}[b]{0.190\textwidth}
\includegraphics[width=1.0\textwidth]{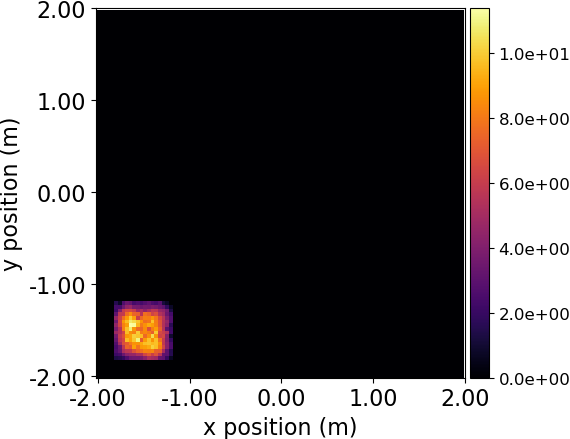} \hfill
\caption{SGPD}
\end{subfigure}
\centering
\caption{Comparison of density prediction accuracies (Ground robot navigation, t=0)}
\label{fig:dens_e04_robot_000}
\end{figure}

\begin{figure}[!htbp]
\begin{subfigure}[b]{0.190\textwidth}
\includegraphics[width=1.0\textwidth]{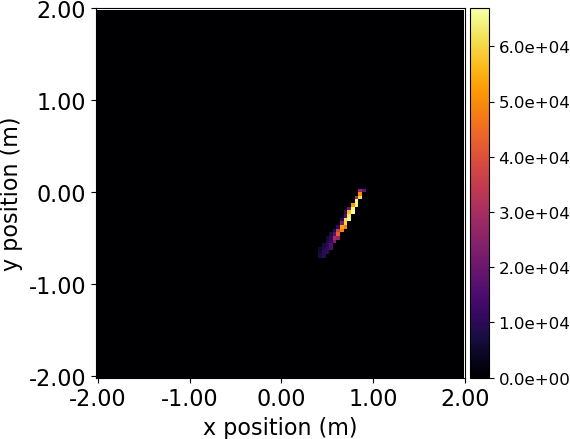} \hfill
\caption{Groundtruth}
\end{subfigure}
\begin{subfigure}[b]{0.190\textwidth}
\includegraphics[width=1.0\textwidth]{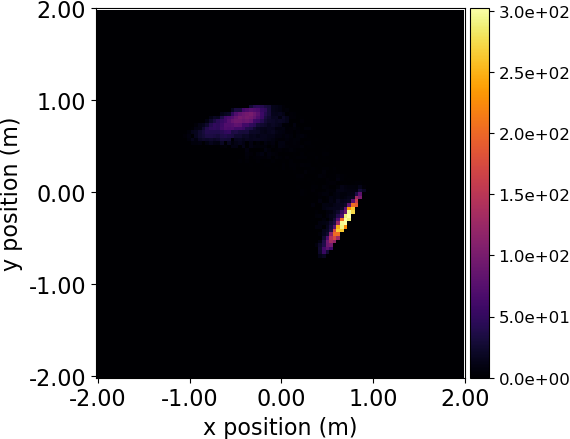} \hfill
\caption{Kernel density}
\end{subfigure}
\begin{subfigure}[b]{0.190\textwidth}
\includegraphics[width=1.0\textwidth]{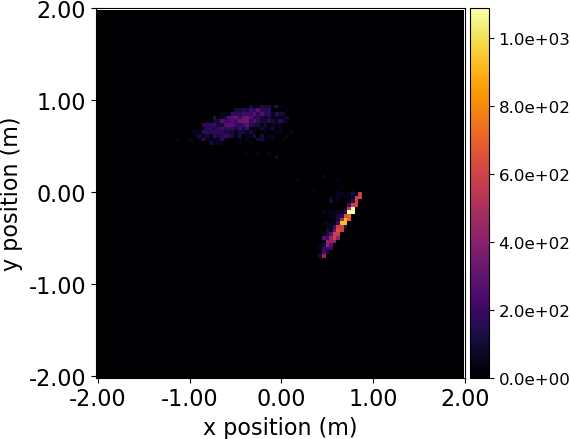} \hfill
\caption{Histogram}
\end{subfigure}
\begin{subfigure}[b]{0.190\textwidth}
\includegraphics[width=1.0\textwidth]{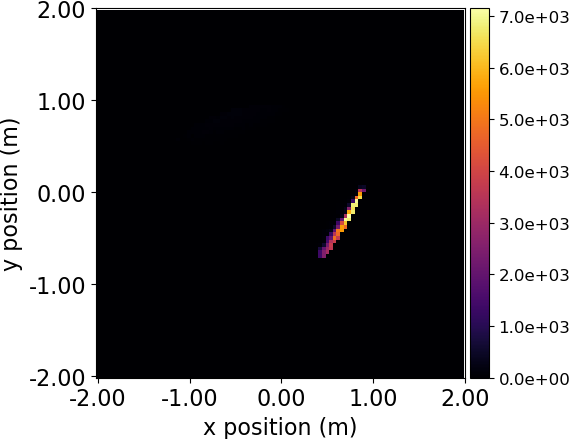} \hfill
\caption{Ours}
\end{subfigure}
\begin{subfigure}[b]{0.190\textwidth}
\includegraphics[width=1.0\textwidth]{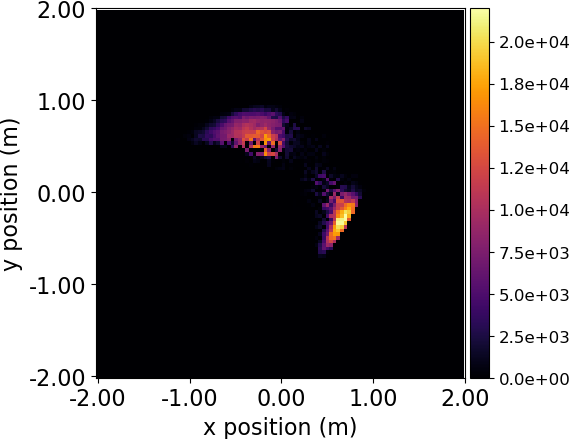} \hfill
\caption{SGPD}
\end{subfigure}
\centering
\caption{Comparison of density prediction accuracies (Ground robot navigation, t=20)}
\label{fig:dens_e04_robot_020}
\end{figure}

\begin{figure}[!htbp]
\begin{subfigure}[b]{0.190\textwidth}
\includegraphics[width=1.0\textwidth]{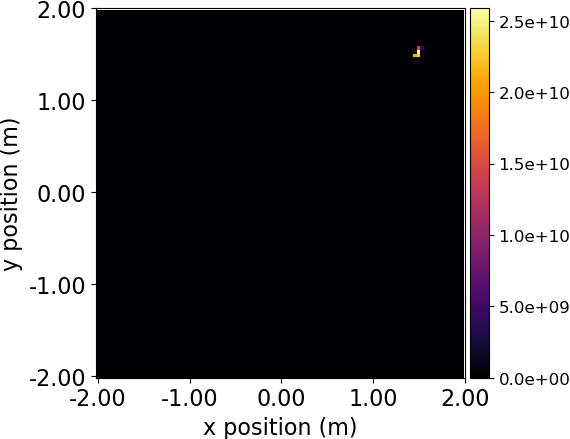} \hfill
\caption{Groundtruth}
\end{subfigure}
\begin{subfigure}[b]{0.190\textwidth}
\includegraphics[width=1.0\textwidth]{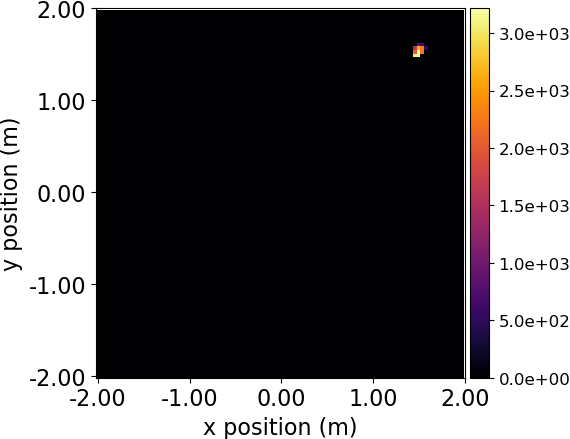} \hfill
\caption{Kernel density}
\end{subfigure}
\begin{subfigure}[b]{0.190\textwidth}
\includegraphics[width=1.0\textwidth]{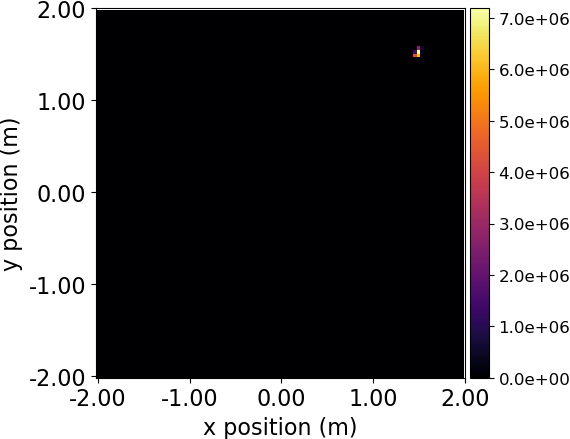} \hfill
\caption{Histogram}
\end{subfigure}
\begin{subfigure}[b]{0.190\textwidth}
\includegraphics[width=1.0\textwidth]{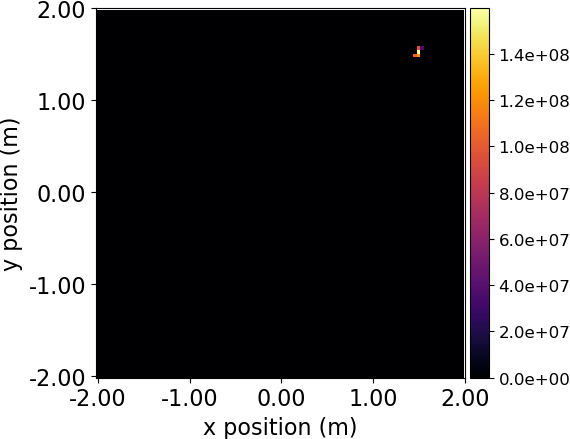} \hfill
\caption{Ours}
\end{subfigure}
\begin{subfigure}[b]{0.190\textwidth}
\includegraphics[width=1.0\textwidth]{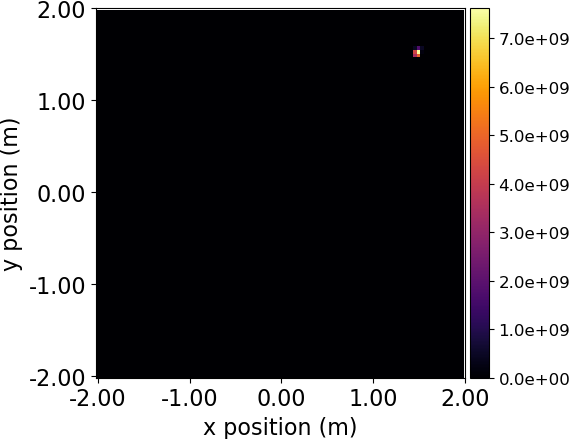} \hfill
\caption{SGPD}
\end{subfigure}
\centering
\caption{Comparison of density prediction accuracies (Ground robot navigation, t=49)}
\label{fig:dens_e04_robot_049}
\end{figure}

\begin{figure}[!htbp]
\begin{subfigure}[b]{0.190\textwidth}
\includegraphics[width=1.0\textwidth]{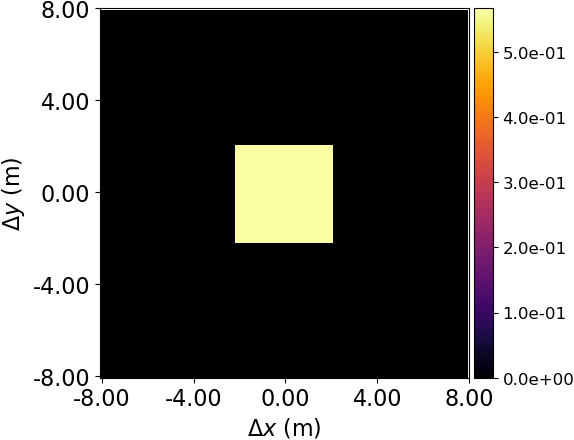} \hfill
\caption{Groundtruth}
\end{subfigure}
\begin{subfigure}[b]{0.190\textwidth}
\includegraphics[width=1.0\textwidth]{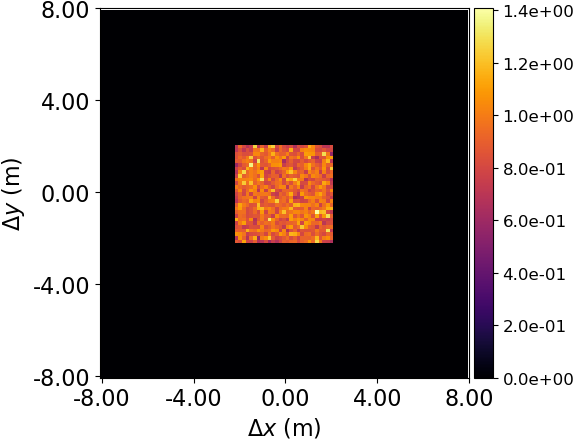} \hfill
\caption{Kernel density}
\end{subfigure}
\begin{subfigure}[b]{0.190\textwidth}
\includegraphics[width=1.0\textwidth]{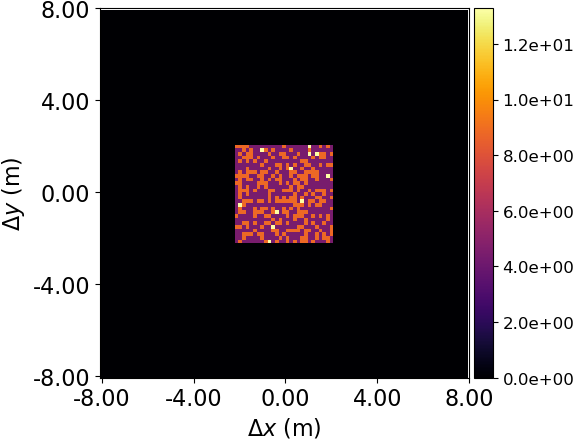} \hfill
\caption{Histogram}
\end{subfigure}
\begin{subfigure}[b]{0.190\textwidth}
\includegraphics[width=1.0\textwidth]{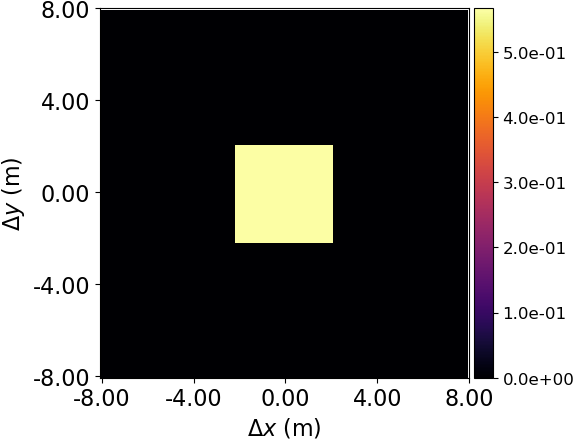} \hfill
\caption{Ours}
\end{subfigure}
\begin{subfigure}[b]{0.190\textwidth}
\includegraphics[width=1.0\textwidth]{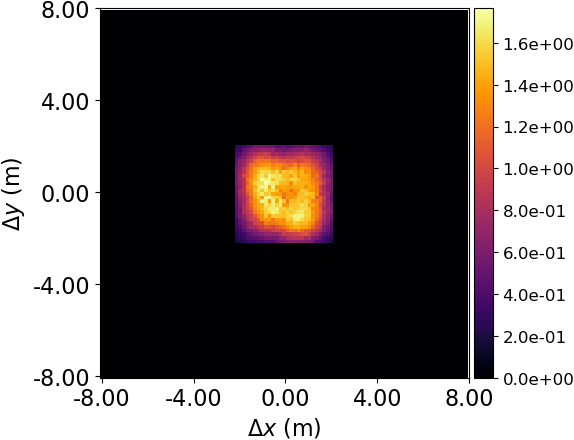} \hfill
\caption{SGPD}
\end{subfigure}
\centering
\caption{Comparison of density prediction accuracies (FACTEST car model, t=0)}
\label{fig:dens_e05_car_000}
\end{figure}

\begin{figure}[!htbp]
\begin{subfigure}[b]{0.190\textwidth}
\includegraphics[width=1.0\textwidth]{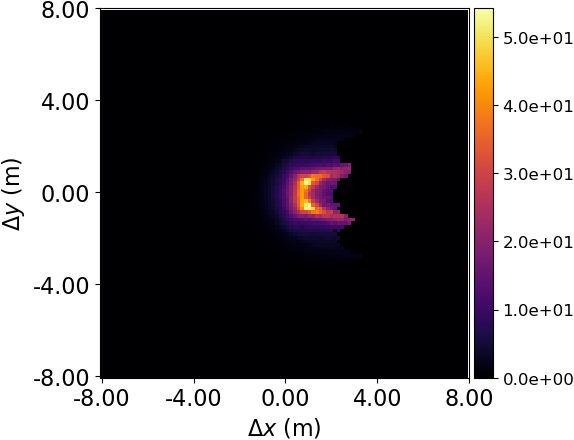} \hfill
\caption{Groundtruth}
\end{subfigure}
\begin{subfigure}[b]{0.190\textwidth}
\includegraphics[width=1.0\textwidth]{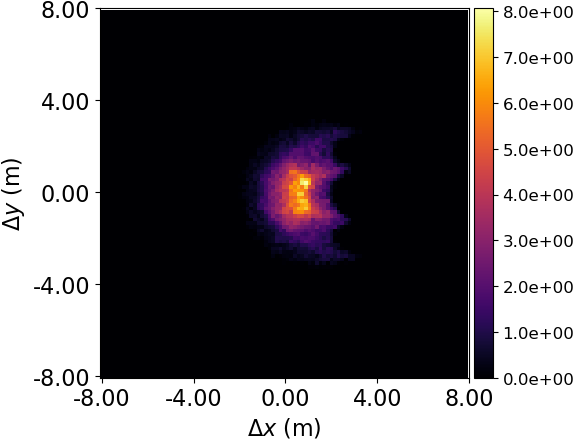} \hfill
\caption{Kernel density}
\end{subfigure}
\begin{subfigure}[b]{0.190\textwidth}
\includegraphics[width=1.0\textwidth]{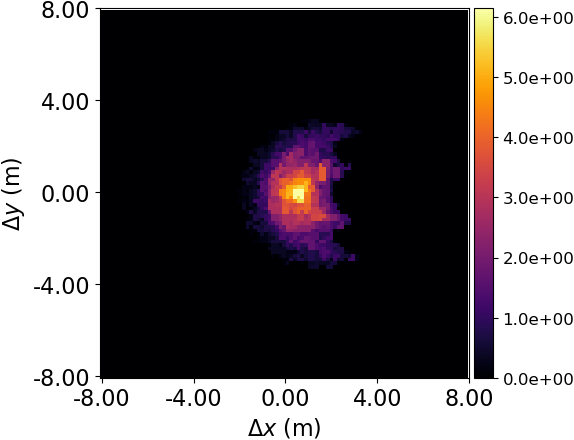} \hfill
\caption{Histogram}
\end{subfigure}
\begin{subfigure}[b]{0.190\textwidth}
\includegraphics[width=1.0\textwidth]{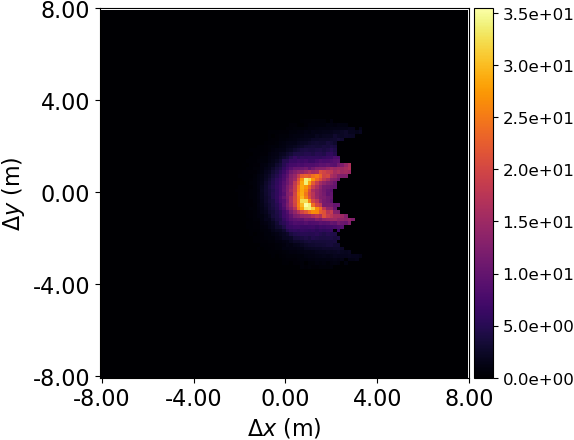} \hfill
\caption{Ours}
\end{subfigure}
\begin{subfigure}[b]{0.190\textwidth}
\includegraphics[width=1.0\textwidth]{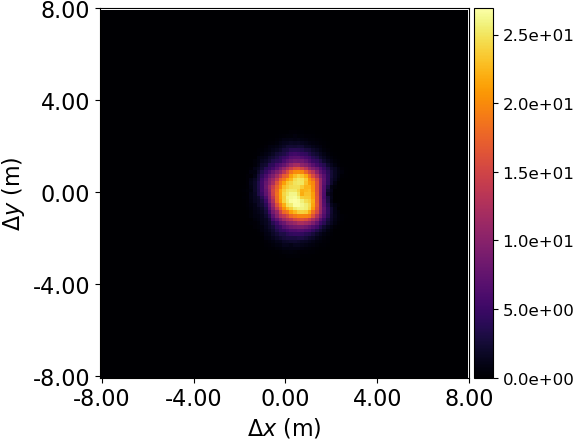} \hfill
\caption{SGPD}
\end{subfigure}
\centering
\caption{Comparison of density prediction accuracies (FACTEST car model, t=20)}
\label{fig:dens_e05_car_020}
\end{figure}

\begin{figure}[!htbp]
\begin{subfigure}[b]{0.190\textwidth}
\includegraphics[width=1.0\textwidth]{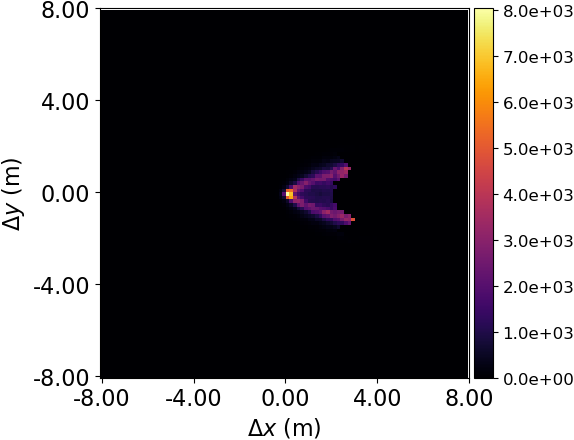} \hfill
\caption{Groundtruth}
\end{subfigure}
\begin{subfigure}[b]{0.190\textwidth}
\includegraphics[width=1.0\textwidth]{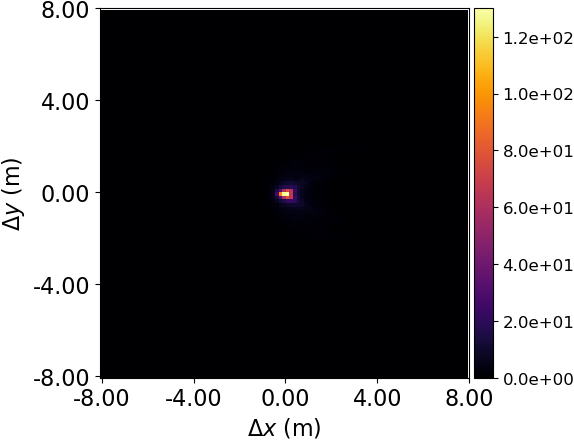} \hfill
\caption{Kernel density}
\end{subfigure}
\begin{subfigure}[b]{0.190\textwidth}
\includegraphics[width=1.0\textwidth]{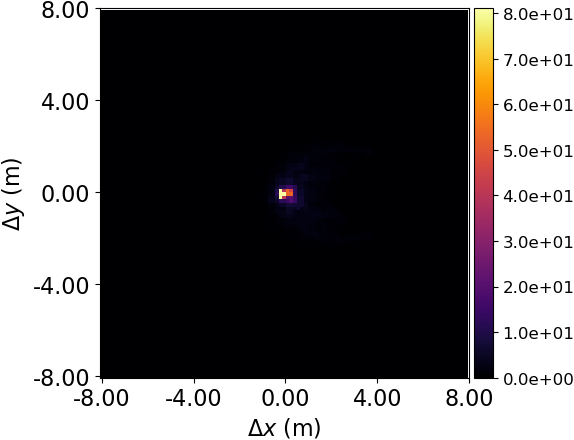} \hfill
\caption{Histogram}
\end{subfigure}
\begin{subfigure}[b]{0.190\textwidth}
\includegraphics[width=1.0\textwidth]{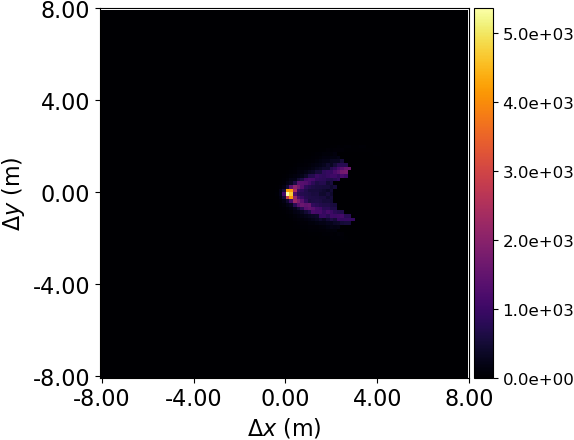} \hfill
\caption{Ours}
\end{subfigure}
\begin{subfigure}[b]{0.190\textwidth}
\includegraphics[width=1.0\textwidth]{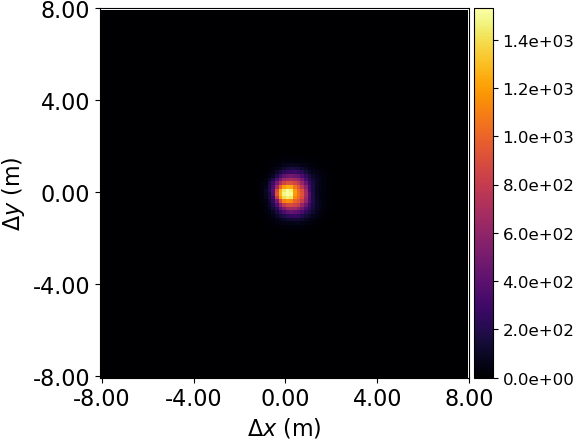} \hfill
\caption{SGPD}
\end{subfigure}
\centering
\caption{Comparison of density prediction accuracies (FACTEST car model, t=49)}
\label{fig:dens_e05_car_049}
\end{figure}

\begin{figure}[!htbp]
\begin{subfigure}[b]{0.190\textwidth}
\includegraphics[width=1.0\textwidth]{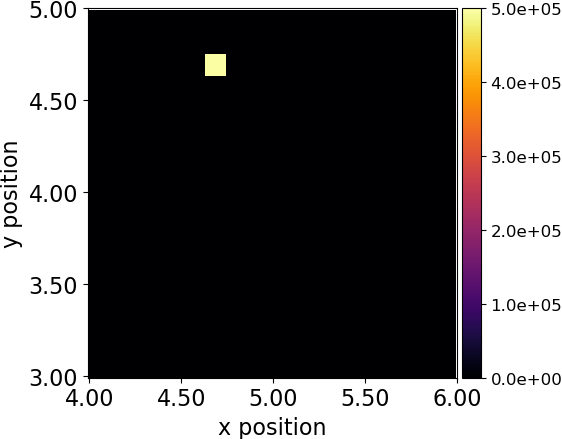} \hfill
\caption{Groundtruth}
\end{subfigure}
\begin{subfigure}[b]{0.190\textwidth}
\includegraphics[width=1.0\textwidth]{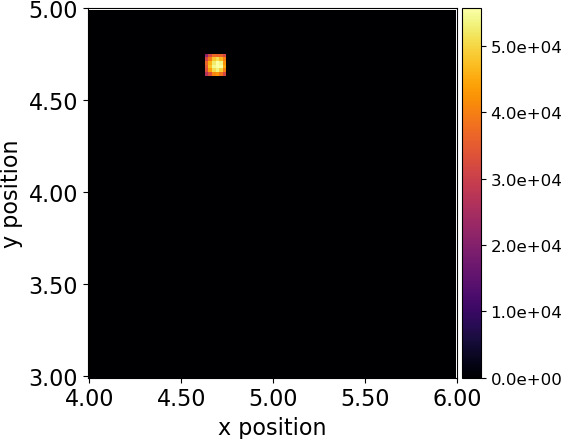} \hfill
\caption{Kernel density}
\end{subfigure}
\begin{subfigure}[b]{0.190\textwidth}
\includegraphics[width=1.0\textwidth]{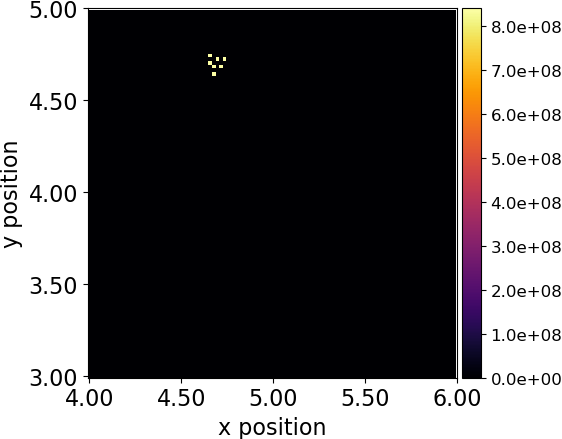} \hfill
\caption{Histogram}
\end{subfigure}
\begin{subfigure}[b]{0.190\textwidth}
\includegraphics[width=1.0\textwidth]{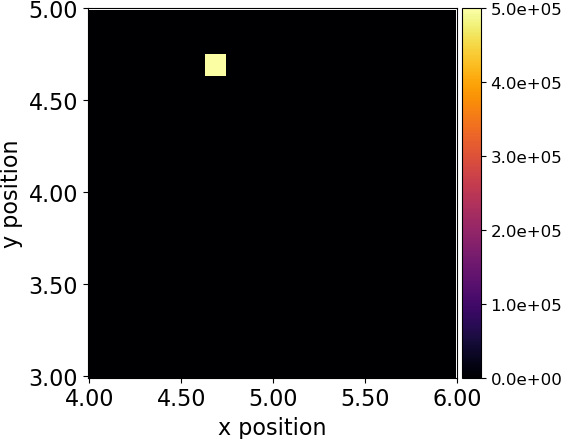} \hfill
\caption{Ours}
\end{subfigure}
\begin{subfigure}[b]{0.190\textwidth}
\includegraphics[width=1.0\textwidth]{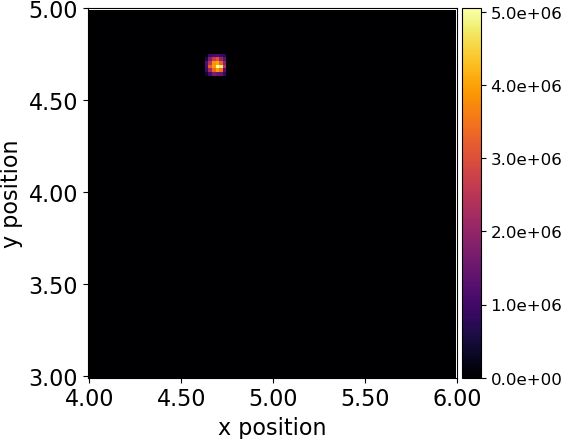} \hfill
\caption{SGPD}
\end{subfigure}
\centering
\caption{Comparison of density prediction accuracies (Quadrotor control system, t=0)}
\label{fig:dens_e06_quad_000}
\end{figure}

\begin{figure}[!htbp]
\begin{subfigure}[b]{0.190\textwidth}
\includegraphics[width=1.0\textwidth]{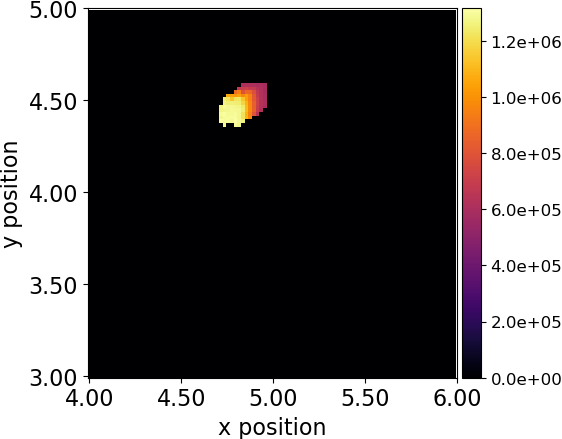} \hfill
\caption{Groundtruth}
\end{subfigure}
\begin{subfigure}[b]{0.190\textwidth}
\includegraphics[width=1.0\textwidth]{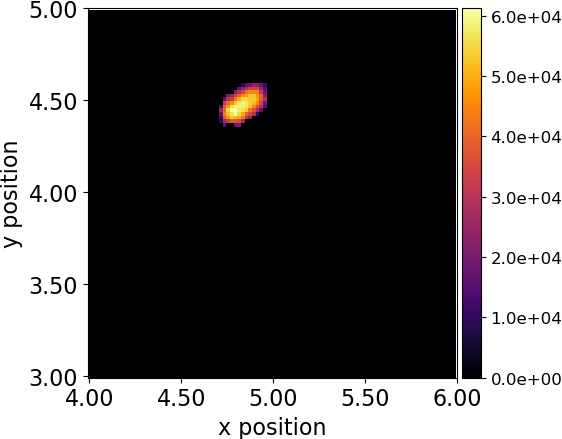} \hfill
\caption{Kernel density}
\end{subfigure}
\begin{subfigure}[b]{0.190\textwidth}
\includegraphics[width=1.0\textwidth]{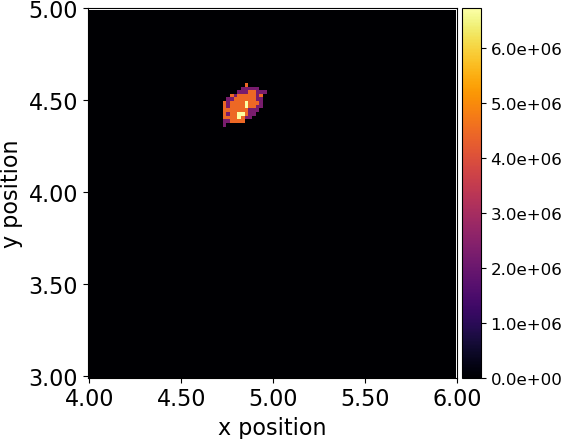} \hfill
\caption{Histogram}
\end{subfigure}
\begin{subfigure}[b]{0.190\textwidth}
\includegraphics[width=1.0\textwidth]{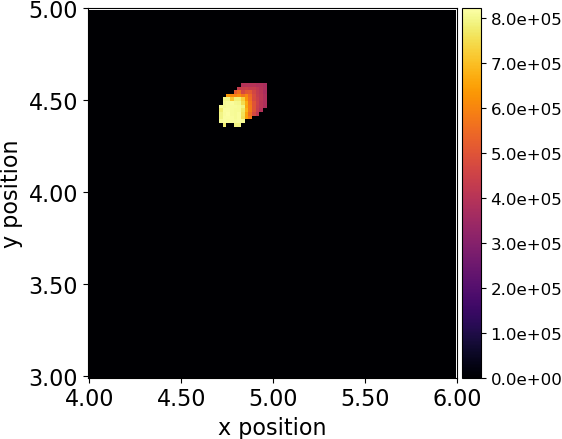} \hfill
\caption{Ours}
\end{subfigure}
\begin{subfigure}[b]{0.190\textwidth}
\includegraphics[width=1.0\textwidth]{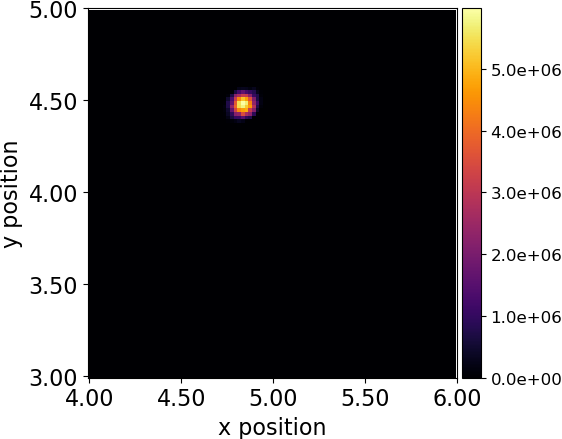} \hfill
\caption{SGPD}
\end{subfigure}
\centering
\caption{Comparison of density prediction accuracies (Quadrotor control system, t=4)}
\label{fig:dens_e06_quad_004}
\end{figure}

\begin{figure}[!htbp]
\begin{subfigure}[b]{0.190\textwidth}
\includegraphics[width=1.0\textwidth]{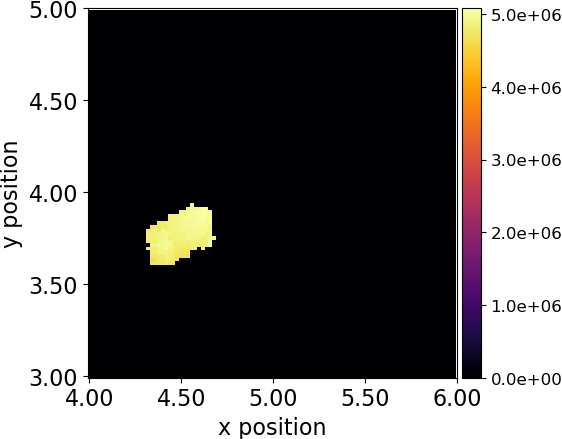} \hfill
\caption{Groundtruth}
\end{subfigure}
\begin{subfigure}[b]{0.190\textwidth}
\includegraphics[width=1.0\textwidth]{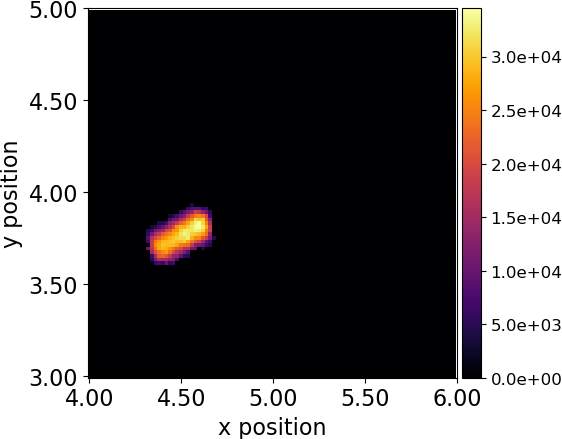} \hfill
\caption{Kernel density}
\end{subfigure}
\begin{subfigure}[b]{0.190\textwidth}
\includegraphics[width=1.0\textwidth]{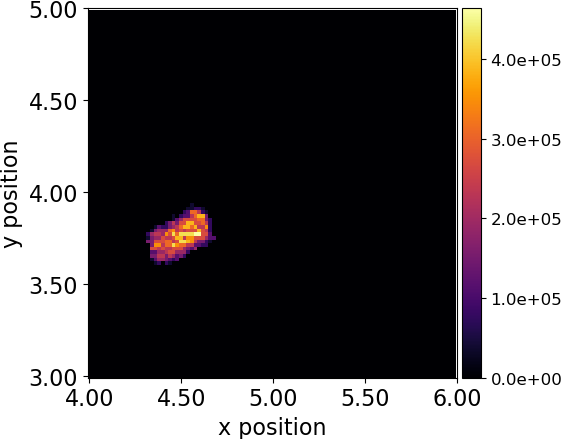} \hfill
\caption{Histogram}
\end{subfigure}
\begin{subfigure}[b]{0.190\textwidth}
\includegraphics[width=1.0\textwidth]{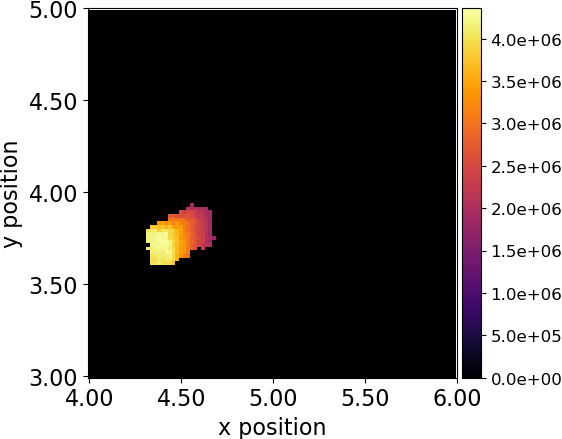} \hfill
\caption{Ours}
\end{subfigure}
\begin{subfigure}[b]{0.190\textwidth}
\includegraphics[width=1.0\textwidth]{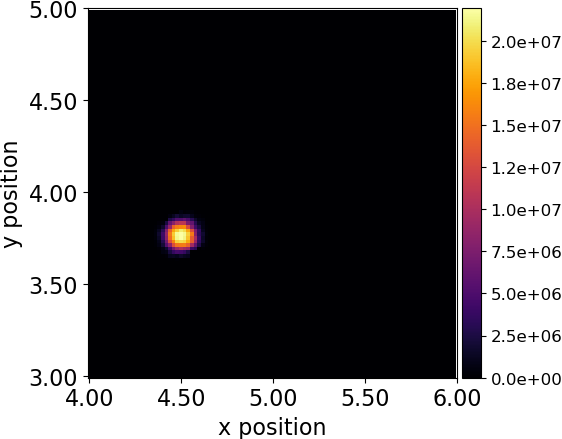} \hfill
\caption{SGPD}
\end{subfigure}
\centering
\caption{Comparison of density prediction accuracies (Quadrotor control system, t=11)}
\label{fig:dens_e06_quad_011}
\end{figure}

\begin{figure}[!htbp]
\begin{subfigure}[b]{0.190\textwidth}
\includegraphics[width=1.0\textwidth]{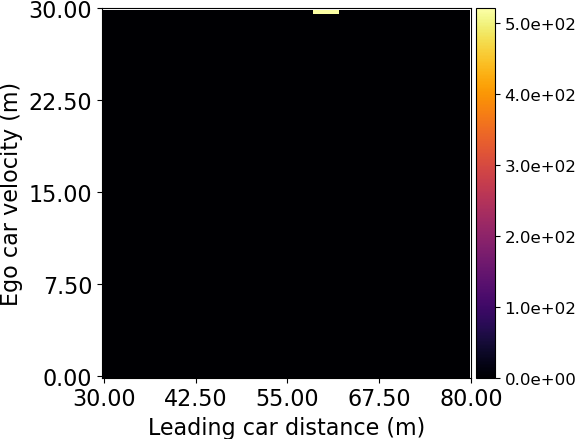} \hfill
\caption{Groundtruth}
\end{subfigure}
\begin{subfigure}[b]{0.190\textwidth}
\includegraphics[width=1.0\textwidth]{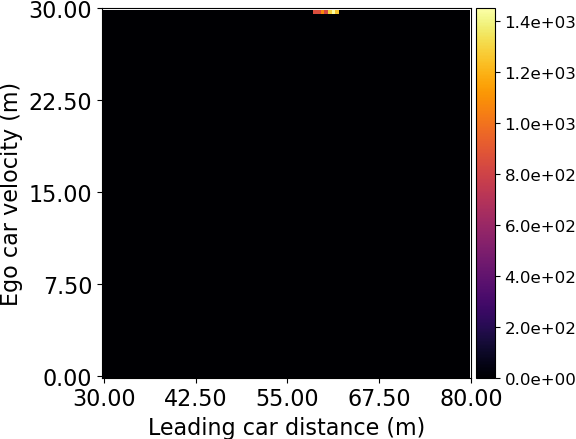} \hfill
\caption{Kernel density}
\end{subfigure}
\begin{subfigure}[b]{0.190\textwidth}
\includegraphics[width=1.0\textwidth]{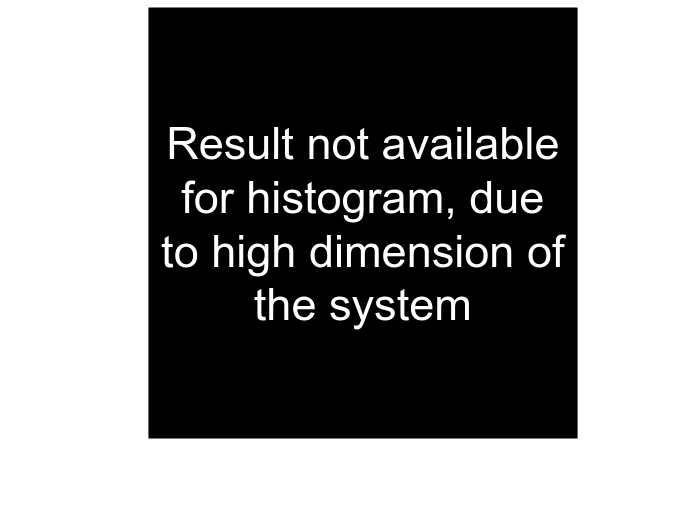} \hfill
\caption{Histogram}
\end{subfigure}
\begin{subfigure}[b]{0.190\textwidth}
\includegraphics[width=1.0\textwidth]{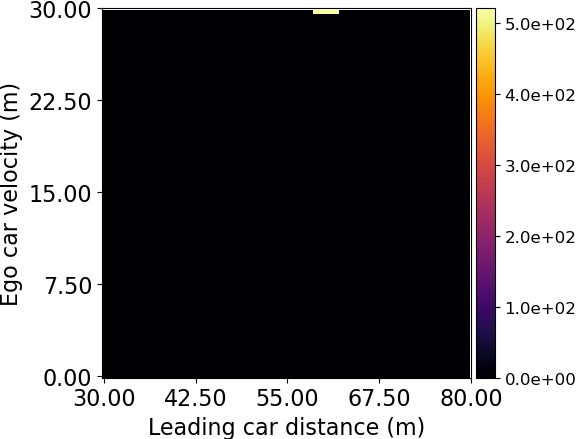} \hfill
\caption{Ours}
\end{subfigure}
\begin{subfigure}[b]{0.190\textwidth}
\includegraphics[width=1.0\textwidth]{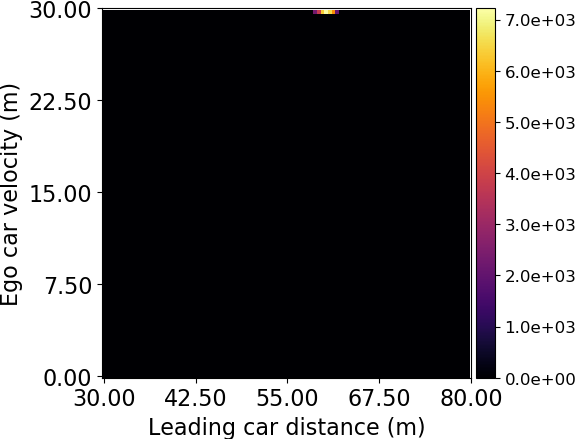} \hfill
\caption{SGPD}
\end{subfigure}
\centering
\caption{Comparison of density prediction accuracies (Adaptive cruise control system, t=0)}
\label{fig:dens_e07_acc_000}
\end{figure}

\begin{figure}[!htbp]
\begin{subfigure}[b]{0.190\textwidth}
\includegraphics[width=1.0\textwidth]{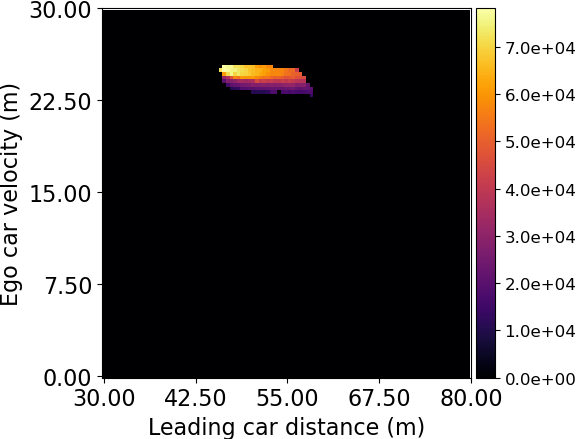} \hfill
\caption{Groundtruth}
\end{subfigure}
\begin{subfigure}[b]{0.190\textwidth}
\includegraphics[width=1.0\textwidth]{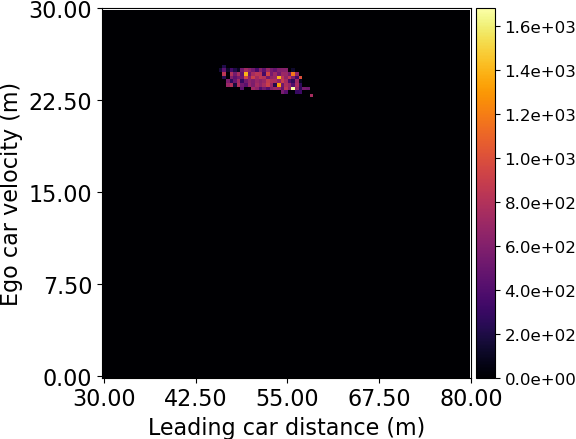} \hfill
\caption{Kernel density}
\end{subfigure}
\begin{subfigure}[b]{0.190\textwidth}
\includegraphics[width=1.0\textwidth]{supple/figs/g0625-095217_images/void_hist.png} \hfill
\caption{Histogram}
\end{subfigure}
\begin{subfigure}[b]{0.190\textwidth}
\includegraphics[width=1.0\textwidth]{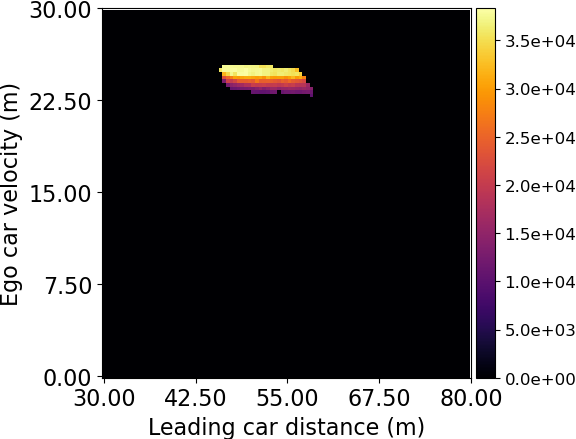} \hfill
\caption{Ours}
\end{subfigure}
\begin{subfigure}[b]{0.190\textwidth}
\includegraphics[width=1.0\textwidth]{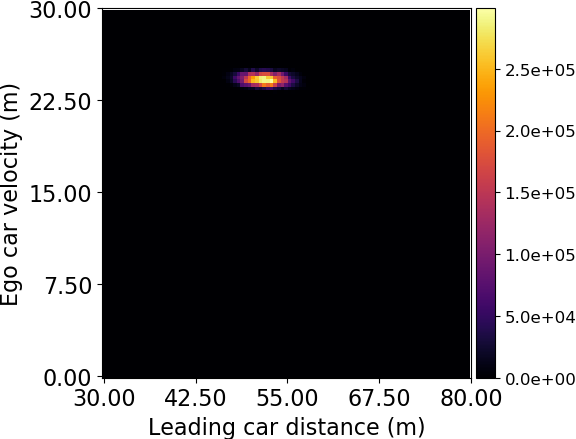} \hfill
\caption{SGPD}
\end{subfigure}
\centering
\caption{Comparison of density prediction accuracies (Adaptive cruise control system, t=20)}
\label{fig:dens_e07_acc_020}
\end{figure}

\begin{figure}[!htbp]
\begin{subfigure}[b]{0.190\textwidth}
\includegraphics[width=1.0\textwidth]{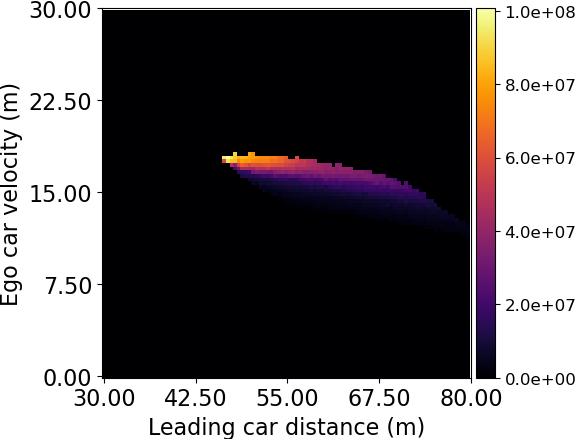} \hfill
\caption{Groundtruth}
\end{subfigure}
\begin{subfigure}[b]{0.190\textwidth}
\includegraphics[width=1.0\textwidth]{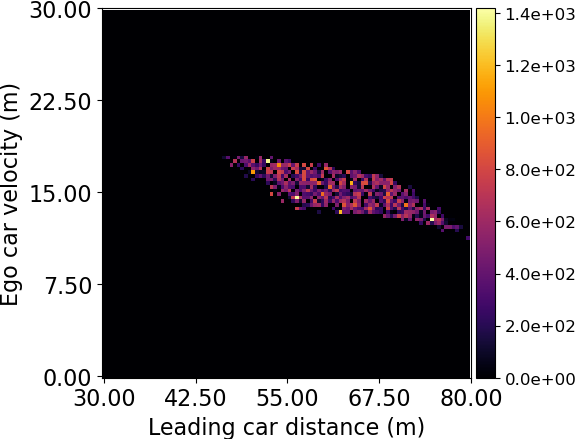} \hfill
\caption{Kernel density}
\end{subfigure}
\begin{subfigure}[b]{0.190\textwidth}
\includegraphics[width=1.0\textwidth]{supple/figs/g0625-095217_images/void_hist.png} \hfill
\caption{Histogram}
\end{subfigure}
\begin{subfigure}[b]{0.190\textwidth}
\includegraphics[width=1.0\textwidth]{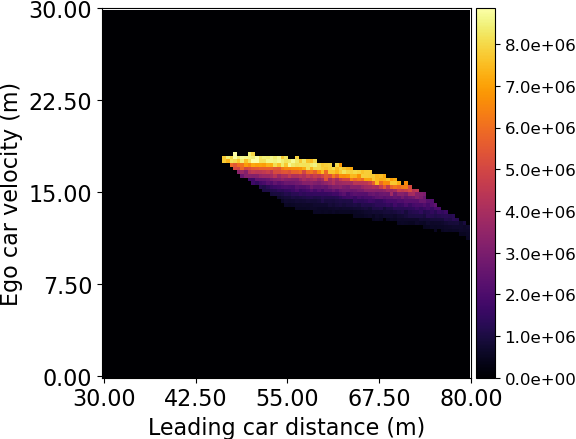} \hfill
\caption{Ours}
\end{subfigure}
\begin{subfigure}[b]{0.190\textwidth}
\includegraphics[width=1.0\textwidth]{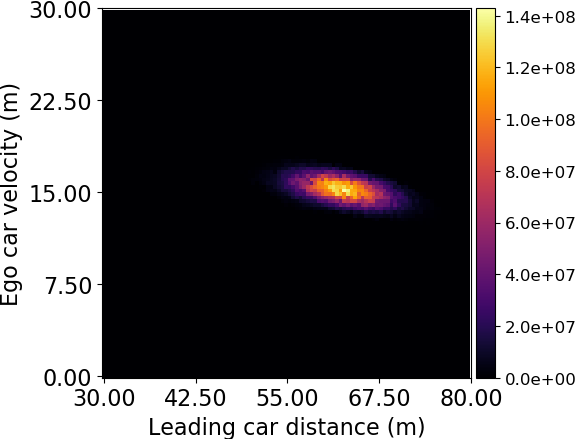} \hfill
\caption{SGPD}
\end{subfigure}
\centering
\caption{Comparison of density prediction accuracies (Adaptive cruise control system, t=49)}
\label{fig:dens_e07_acc_049}
\end{figure}

\begin{figure}[!htbp]
\begin{subfigure}[b]{0.190\textwidth}
\includegraphics[width=1.0\textwidth]{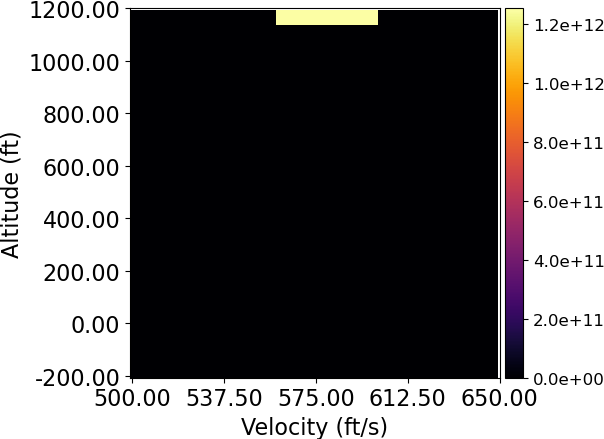} \hfill
\caption{Groundtruth}
\end{subfigure}
\begin{subfigure}[b]{0.190\textwidth}
\includegraphics[width=1.0\textwidth]{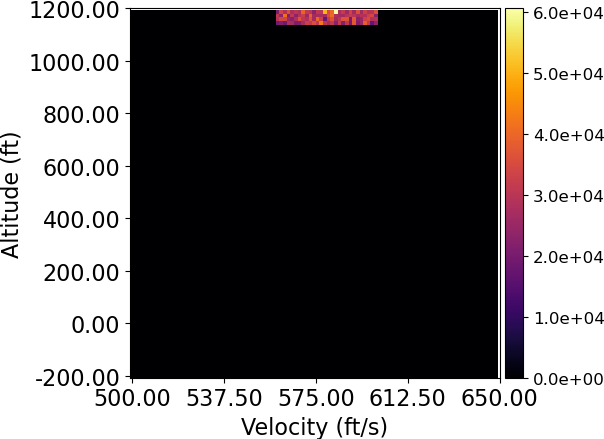} \hfill
\caption{Kernel density}
\end{subfigure}
\begin{subfigure}[b]{0.190\textwidth}
\includegraphics[width=1.0\textwidth]{supple/figs/g0625-095217_images/void_hist.png} \hfill
\caption{Histogram}
\end{subfigure}
\begin{subfigure}[b]{0.190\textwidth}
\includegraphics[width=1.0\textwidth]{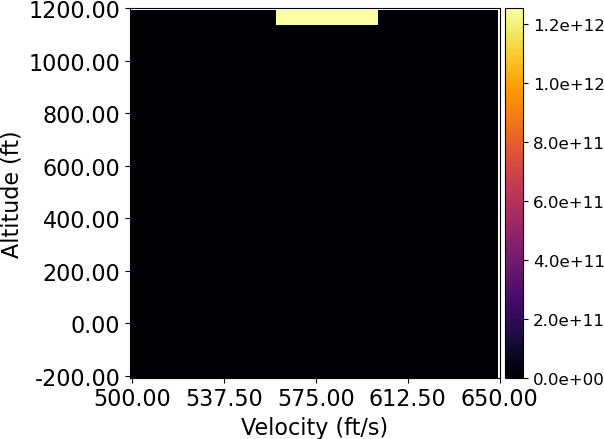} \hfill
\caption{Ours}
\end{subfigure}
\begin{subfigure}[b]{0.190\textwidth}
\includegraphics[width=1.0\textwidth]{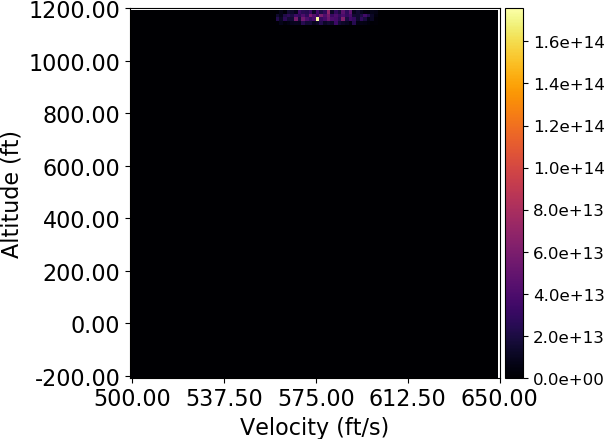} \hfill
\caption{SGPD}
\end{subfigure}
\centering
\caption{Comparison of density prediction accuracies (Ground collision avoidance system, t=0)}
\label{fig:dens_e08_gcas_000}
\end{figure}

\begin{figure}[!htbp]
\begin{subfigure}[b]{0.190\textwidth}
\includegraphics[width=1.0\textwidth]{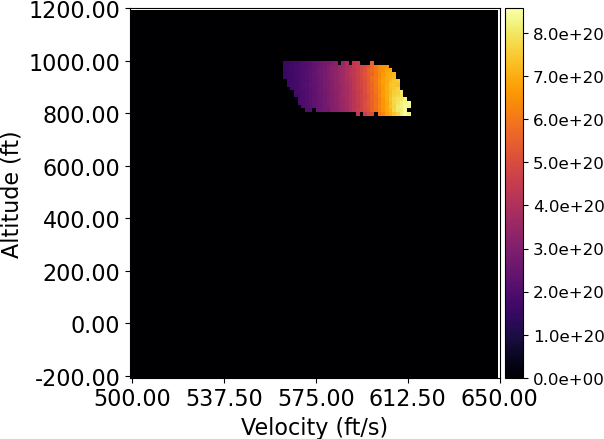} \hfill
\caption{Groundtruth}
\end{subfigure}
\begin{subfigure}[b]{0.190\textwidth}
\includegraphics[width=1.0\textwidth]{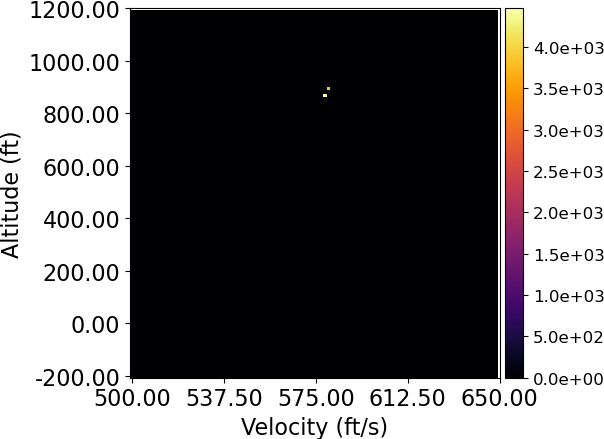} \hfill
\caption{Kernel density}
\end{subfigure}
\begin{subfigure}[b]{0.190\textwidth}
\includegraphics[width=1.0\textwidth]{supple/figs/g0625-095217_images/void_hist.png} \hfill
\caption{Histogram}
\end{subfigure}
\begin{subfigure}[b]{0.190\textwidth}
\includegraphics[width=1.0\textwidth]{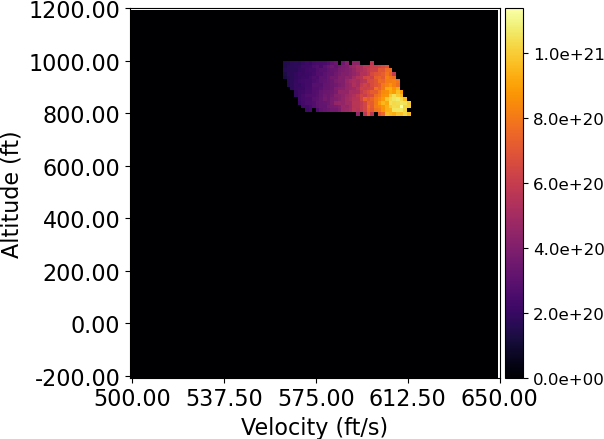} \hfill
\caption{Ours}
\end{subfigure}
\begin{subfigure}[b]{0.190\textwidth}
\includegraphics[width=1.0\textwidth]{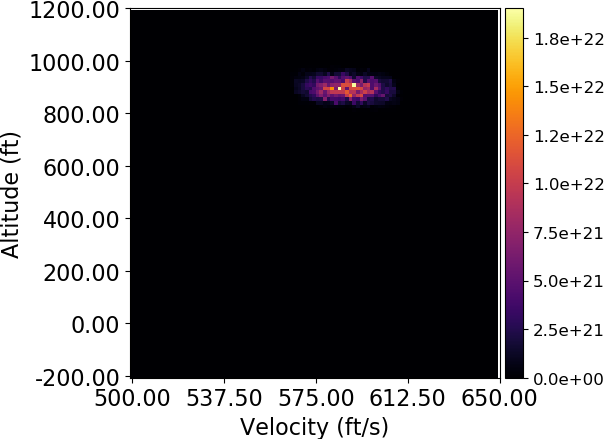} \hfill
\caption{SGPD}
\end{subfigure}
\centering
\caption{Comparison of density prediction accuracies (Ground collision avoidance system, t=20)}
\label{fig:dens_e08_gcas_020}
\end{figure}

\begin{figure}[!htbp]
\begin{subfigure}[b]{0.190\textwidth}
\includegraphics[width=1.0\textwidth]{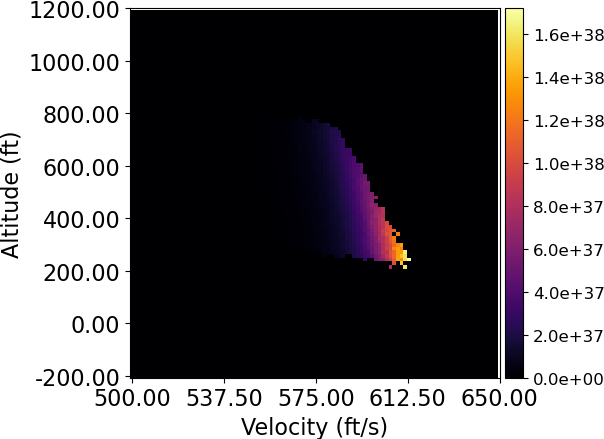} \hfill
\caption{Groundtruth}
\end{subfigure}
\begin{subfigure}[b]{0.190\textwidth}
\includegraphics[width=1.0\textwidth]{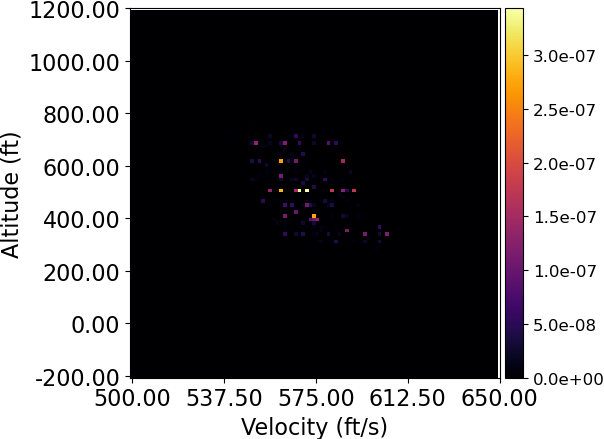} \hfill
\caption{Kernel density}
\end{subfigure}
\begin{subfigure}[b]{0.190\textwidth}
\includegraphics[width=1.0\textwidth]{supple/figs/g0625-095217_images/void_hist.png} \hfill
\caption{Histogram}
\end{subfigure}
\begin{subfigure}[b]{0.190\textwidth}
\includegraphics[width=1.0\textwidth]{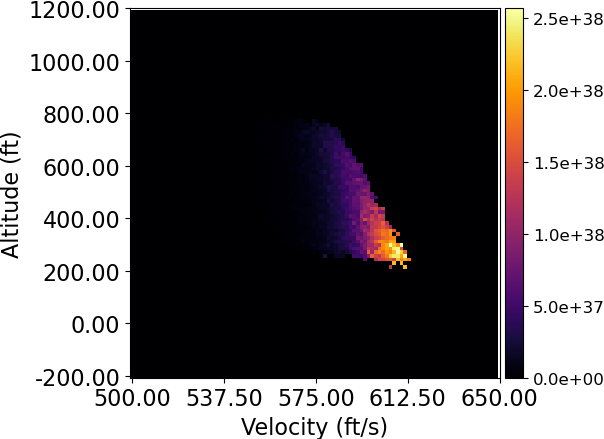} \hfill
\caption{Ours}
\end{subfigure}
\begin{subfigure}[b]{0.190\textwidth}
\includegraphics[width=1.0\textwidth]{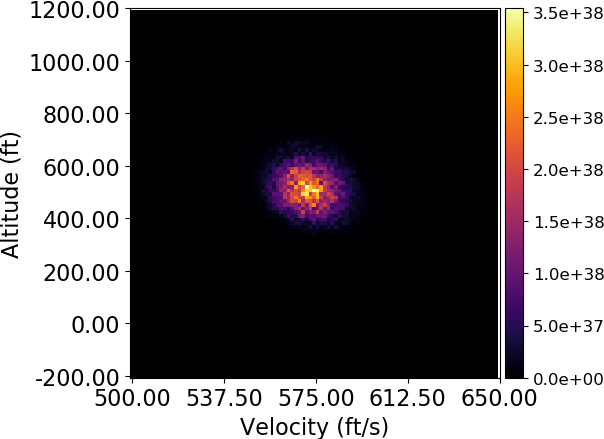} \hfill
\caption{SGPD}
\end{subfigure}
\centering
\caption{Comparison of density prediction accuracies (Ground collision avoidance system, t=59)}
\label{fig:dens_e08_gcas_059}
\end{figure}

\begin{figure}[!htbp]
\begin{subfigure}[b]{0.190\textwidth}
\includegraphics[width=1.0\textwidth]{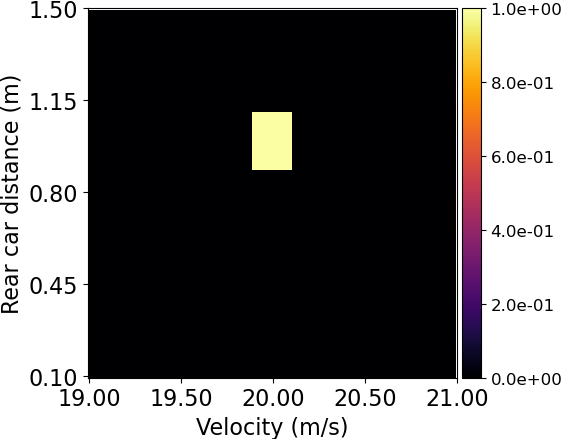} \hfill
\caption{Groundtruth}
\end{subfigure}
\begin{subfigure}[b]{0.190\textwidth}
\includegraphics[width=1.0\textwidth]{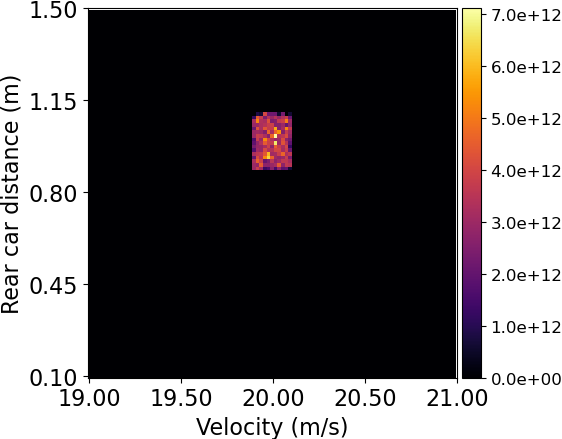} \hfill
\caption{Kernel density}
\end{subfigure}
\begin{subfigure}[b]{0.190\textwidth}
\includegraphics[width=1.0\textwidth]{supple/figs/g0625-095217_images/void_hist.png} \hfill
\caption{Histogram}
\end{subfigure}
\begin{subfigure}[b]{0.190\textwidth}
\includegraphics[width=1.0\textwidth]{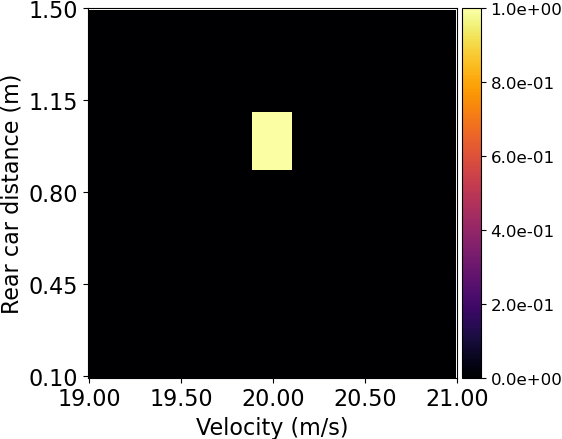} \hfill
\caption{Ours}
\end{subfigure}
\begin{subfigure}[b]{0.190\textwidth}
\includegraphics[width=1.0\textwidth]{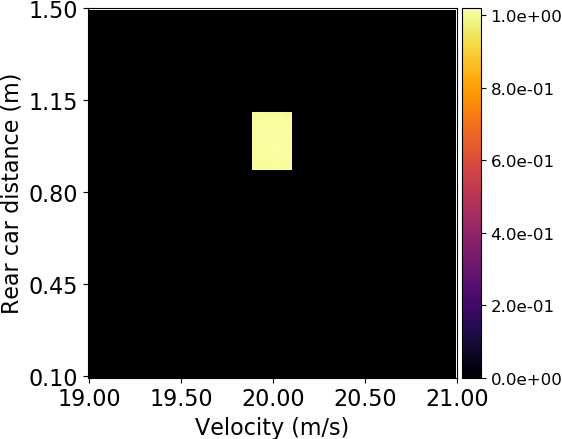} \hfill
\caption{SGPD}
\end{subfigure}
\centering
\caption{Comparison of density prediction accuracies (8-Car platoon system, t=0)}
\label{fig:dens_e09_toon_000}
\end{figure}

\begin{figure}[!htbp]
\begin{subfigure}[b]{0.190\textwidth}
\includegraphics[width=1.0\textwidth]{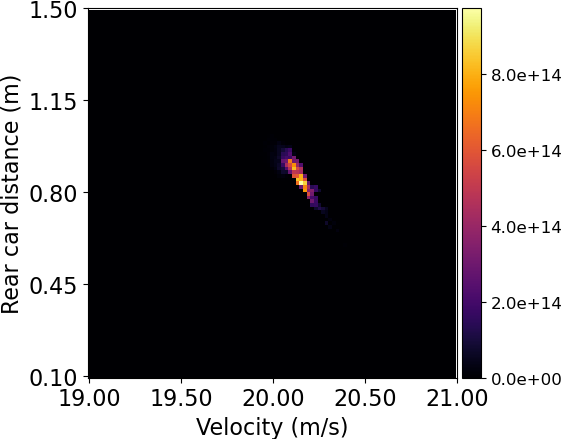} \hfill
\caption{Groundtruth}
\end{subfigure}
\begin{subfigure}[b]{0.190\textwidth}
\includegraphics[width=1.0\textwidth]{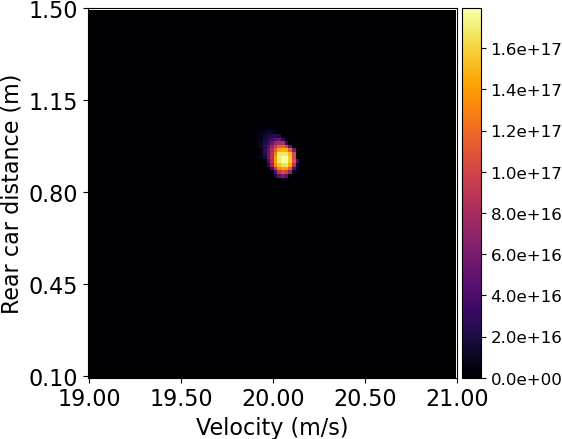} \hfill
\caption{Kernel density}
\end{subfigure}
\begin{subfigure}[b]{0.190\textwidth}
\includegraphics[width=1.0\textwidth]{supple/figs/g0625-095217_images/void_hist.png} \hfill
\caption{Histogram}
\end{subfigure}
\begin{subfigure}[b]{0.190\textwidth}
\includegraphics[width=1.0\textwidth]{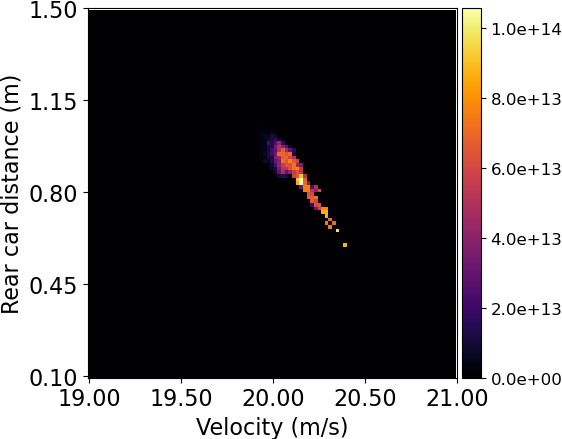} \hfill
\caption{Ours}
\end{subfigure}
\begin{subfigure}[b]{0.190\textwidth}
\includegraphics[width=1.0\textwidth]{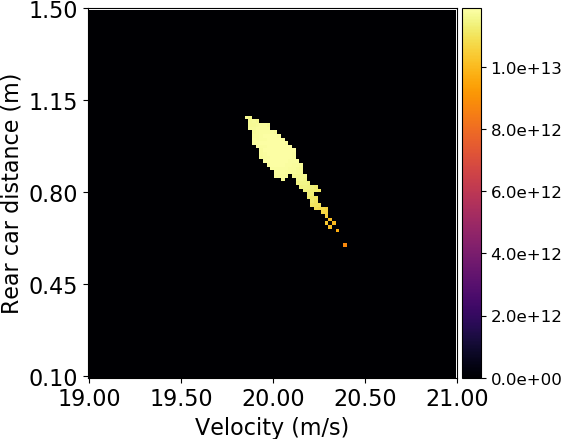} \hfill
\caption{SGPD}
\end{subfigure}
\centering
\caption{Comparison of density prediction accuracies (8-Car platoon system, t=20)}
\label{fig:dens_e09_toon_020}
\end{figure}

\begin{figure}[!htbp]
\begin{subfigure}[b]{0.190\textwidth}
\includegraphics[width=1.0\textwidth]{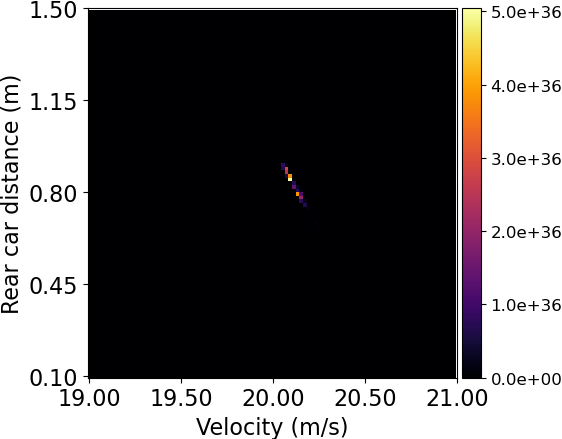} \hfill
\caption{Groundtruth}
\end{subfigure}
\begin{subfigure}[b]{0.190\textwidth}
\includegraphics[width=1.0\textwidth]{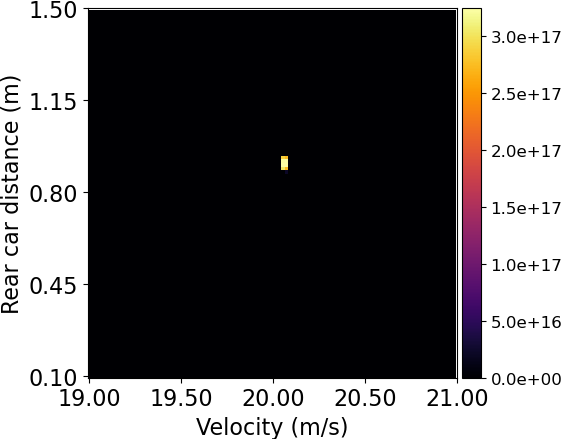} \hfill
\caption{Kernel density}
\end{subfigure}
\begin{subfigure}[b]{0.190\textwidth}
\includegraphics[width=1.0\textwidth]{supple/figs/g0625-095217_images/void_hist.png} \hfill
\caption{Histogram}
\end{subfigure}
\begin{subfigure}[b]{0.190\textwidth}
\includegraphics[width=1.0\textwidth]{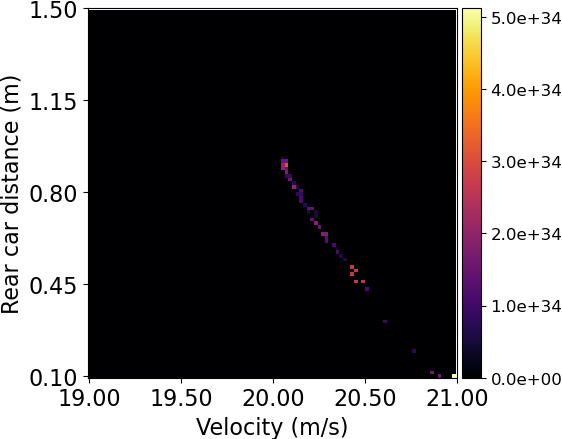} \hfill
\caption{Ours}
\end{subfigure}
\begin{subfigure}[b]{0.190\textwidth}
\includegraphics[width=1.0\textwidth]{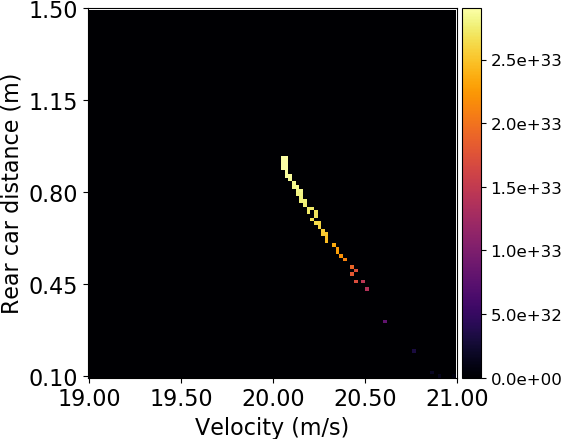} \hfill
\caption{SGPD}
\end{subfigure}
\centering
\caption{Comparison of density prediction accuracies (8-Car platoon system, t=49)}
\label{fig:dens_e09_toon_049}
\end{figure}

\newpage
\subsection{Comparison of reachable set computation among different tools}
\label{sec:f2}

\begin{figure}[!htbp]
\begin{subfigure}[b]{0.222\textwidth}
\includegraphics[width=1.0\textwidth]{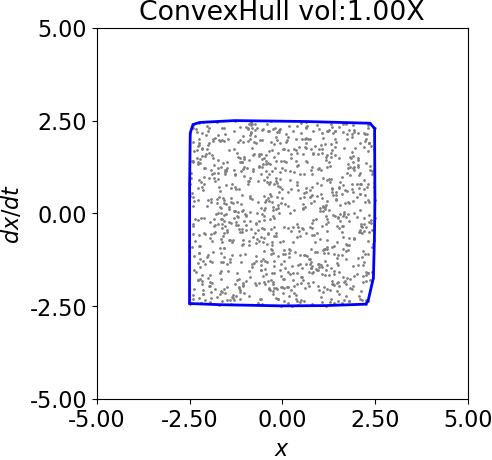} \hfill
\caption{Convex Hull}
\end{subfigure}
\begin{subfigure}[b]{0.222\textwidth}
\includegraphics[width=1.0\textwidth]{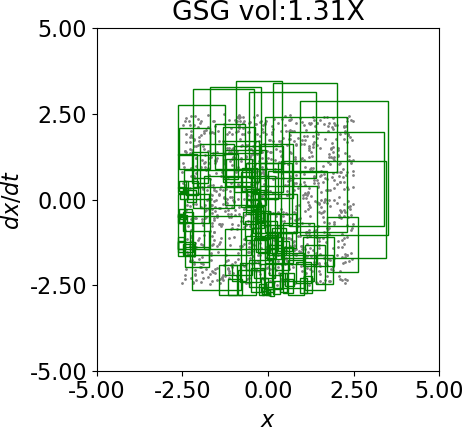} \hfill
\caption{GSG}
\end{subfigure}
\begin{subfigure}[b]{0.222\textwidth}
\includegraphics[width=1.0\textwidth]{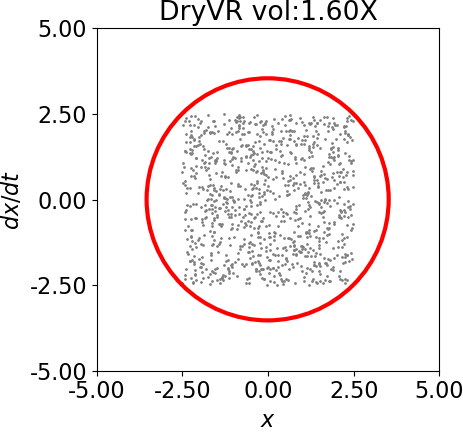} \hfill
\caption{DryVR}
\end{subfigure}
\begin{subfigure}[b]{0.253\textwidth}
\includegraphics[width=1.0\textwidth]{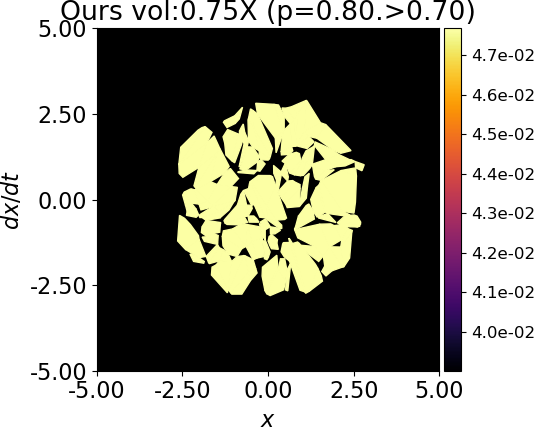} \hfill
\caption{Ours}
\end{subfigure}
\centering
\caption{Comparison of reachable set computation among different tools (Van der Pol, t=0)\CAPRBT}
\label{fig:reach_e00_vdp_000}
\end{figure}

\begin{figure}[!htbp]
\begin{subfigure}[b]{0.222\textwidth}
\includegraphics[width=1.0\textwidth]{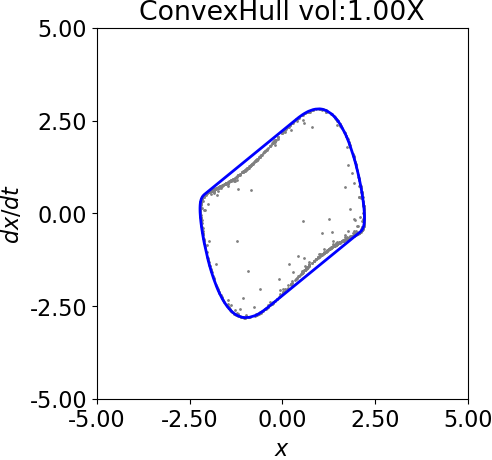} \hfill
\caption{Convex Hull}
\end{subfigure}
\begin{subfigure}[b]{0.222\textwidth}
\includegraphics[width=1.0\textwidth]{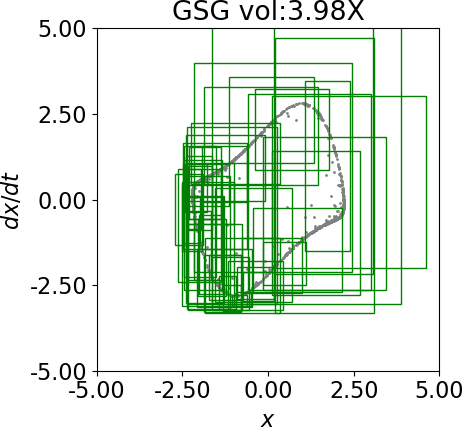} \hfill
\caption{GSG}
\end{subfigure}
\begin{subfigure}[b]{0.222\textwidth}
\includegraphics[width=1.0\textwidth]{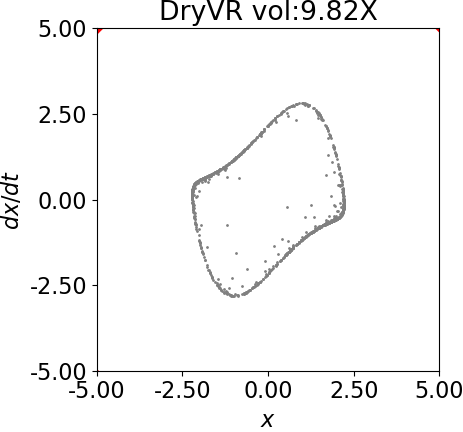} \hfill
\caption{DryVR}
\end{subfigure}
\begin{subfigure}[b]{0.253\textwidth}
\includegraphics[width=1.0\textwidth]{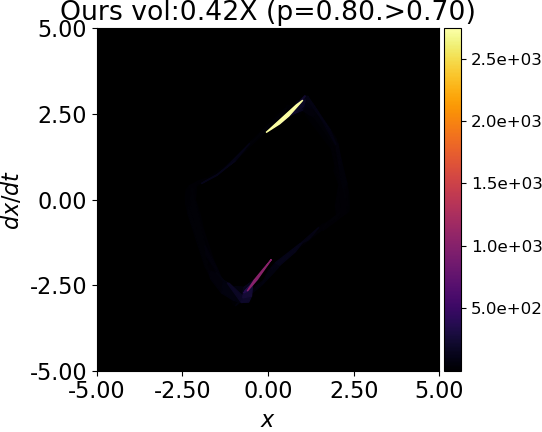} \hfill
\caption{Ours}
\end{subfigure}
\centering
\caption{Comparison of reachable set computation among different tools (Van der Pol, t=40)\CAPRBT}
\label{fig:reach_e00_vdp_040}
\end{figure}

\begin{figure}[!htbp]
\begin{subfigure}[b]{0.222\textwidth}
\includegraphics[width=1.0\textwidth]{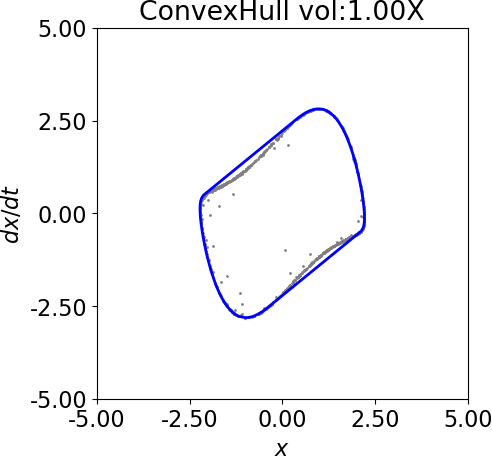} \hfill
\caption{Convex Hull}
\end{subfigure}
\begin{subfigure}[b]{0.222\textwidth}
\includegraphics[width=1.0\textwidth]{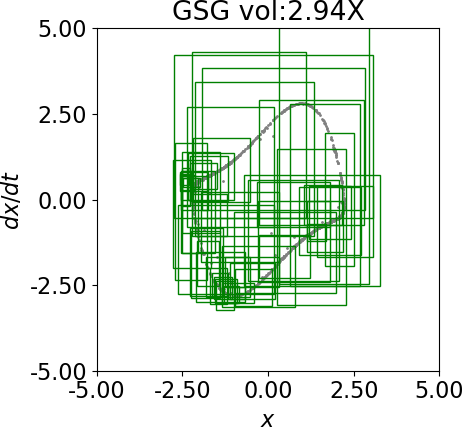} \hfill
\caption{GSG}
\end{subfigure}
\begin{subfigure}[b]{0.222\textwidth}
\includegraphics[width=1.0\textwidth]{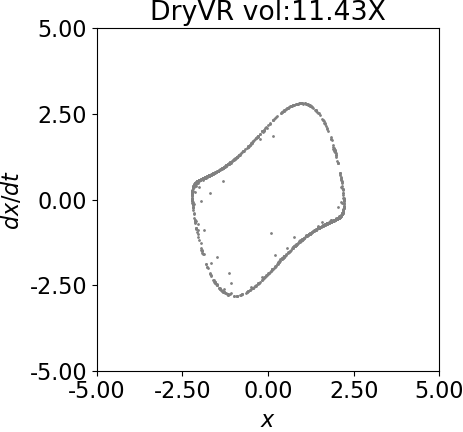} \hfill
\caption{DryVR}
\end{subfigure}
\begin{subfigure}[b]{0.253\textwidth}
\includegraphics[width=1.0\textwidth]{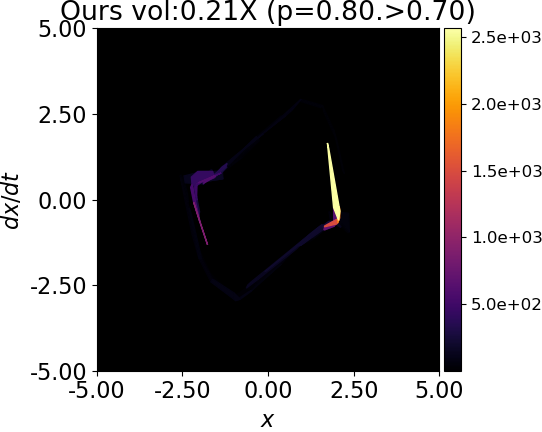} \hfill
\caption{Ours}
\end{subfigure}
\centering
\caption{Comparison of reachable set computation among different tools (Van der Pol, t=49)\CAPRBT}
\label{fig:reach_e00_vdp_049}
\end{figure}

\begin{figure}[!htbp]
\begin{subfigure}[b]{0.222\textwidth}
\includegraphics[width=1.0\textwidth]{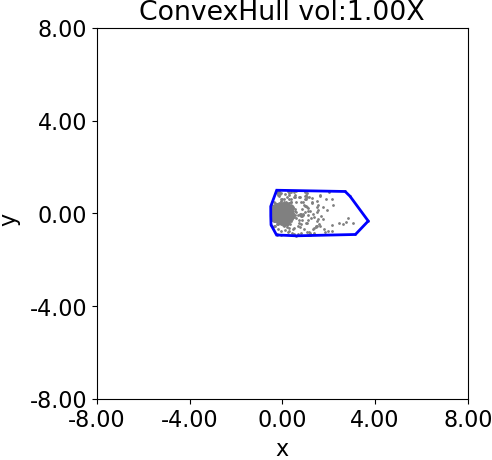} \hfill
\caption{Convex Hull}
\end{subfigure}
\begin{subfigure}[b]{0.222\textwidth}
\includegraphics[width=1.0\textwidth]{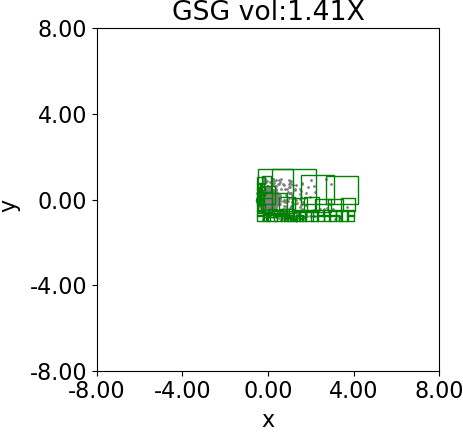} \hfill
\caption{GSG}
\end{subfigure}
\begin{subfigure}[b]{0.222\textwidth}
\includegraphics[width=1.0\textwidth]{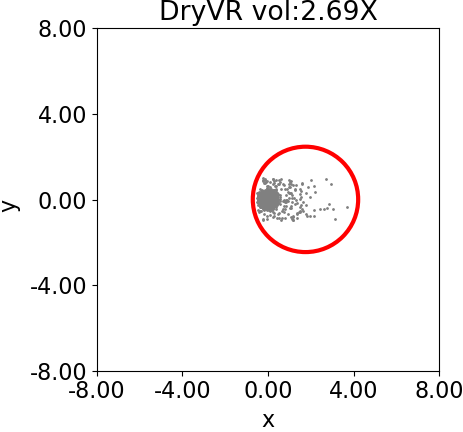} \hfill
\caption{DryVR}
\end{subfigure}
\begin{subfigure}[b]{0.253\textwidth}
\includegraphics[width=1.0\textwidth]{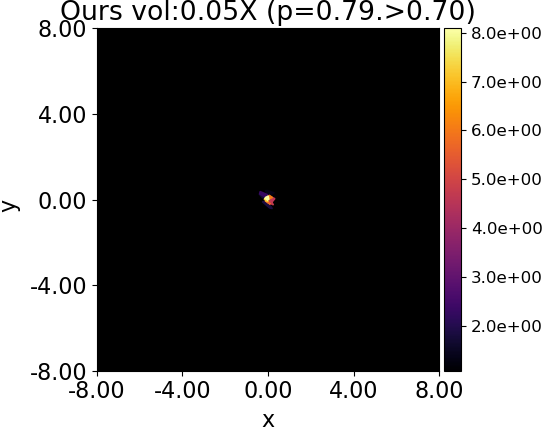} \hfill
\caption{Ours}
\end{subfigure}
\centering
\caption{Comparison of reachable set computation among different tools (Double integrator, t=0)\CAPRBT}
\label{fig:reach_e01_dint_000}
\end{figure}

\begin{figure}[!htbp]
\begin{subfigure}[b]{0.222\textwidth}
\includegraphics[width=1.0\textwidth]{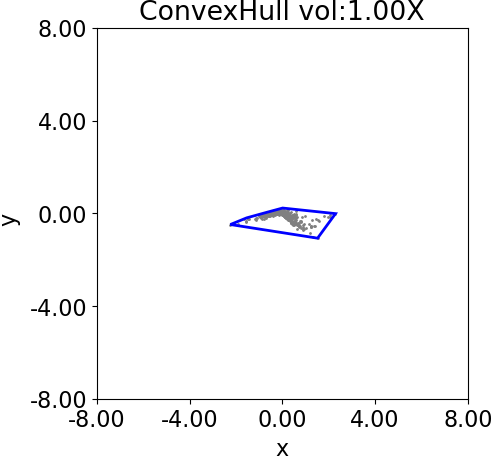} \hfill
\caption{Convex Hull}
\end{subfigure}
\begin{subfigure}[b]{0.222\textwidth}
\includegraphics[width=1.0\textwidth]{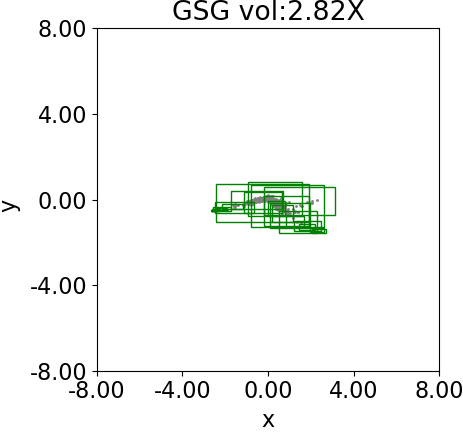} \hfill
\caption{GSG}
\end{subfigure}
\begin{subfigure}[b]{0.222\textwidth}
\includegraphics[width=1.0\textwidth]{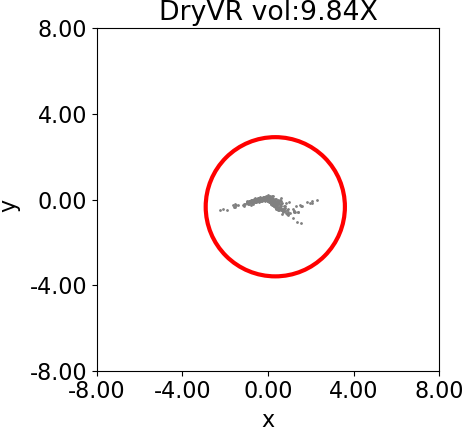} \hfill
\caption{DryVR}
\end{subfigure}
\begin{subfigure}[b]{0.253\textwidth}
\includegraphics[width=1.0\textwidth]{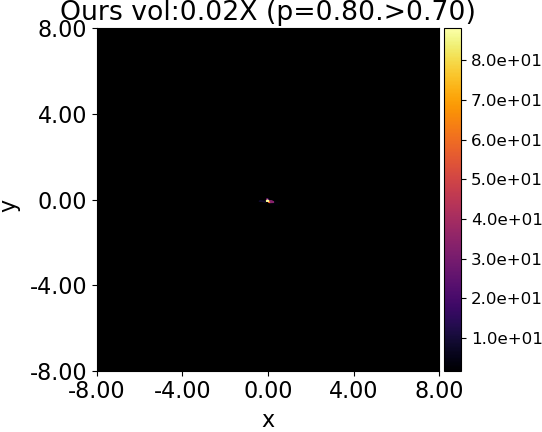} \hfill
\caption{Ours}
\end{subfigure}
\centering
\caption{Comparison of reachable set computation among different tools (Double integrator, t=3)\CAPRBT}
\label{fig:reach_e01_dint_003}
\end{figure}

\begin{figure}[!htbp]
\begin{subfigure}[b]{0.222\textwidth}
\includegraphics[width=1.0\textwidth]{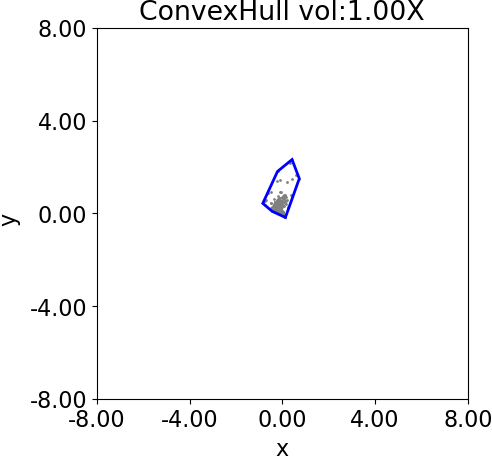} \hfill
\caption{Convex Hull}
\end{subfigure}
\begin{subfigure}[b]{0.222\textwidth}
\includegraphics[width=1.0\textwidth]{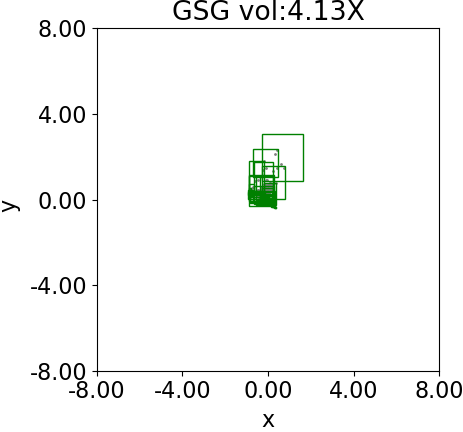} \hfill
\caption{GSG}
\end{subfigure}
\begin{subfigure}[b]{0.222\textwidth}
\includegraphics[width=1.0\textwidth]{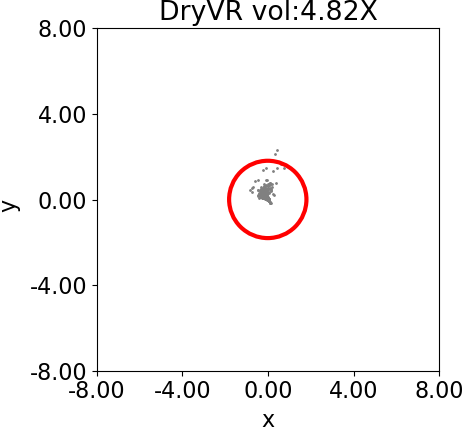} \hfill
\caption{DryVR}
\end{subfigure}
\begin{subfigure}[b]{0.253\textwidth}
\includegraphics[width=1.0\textwidth]{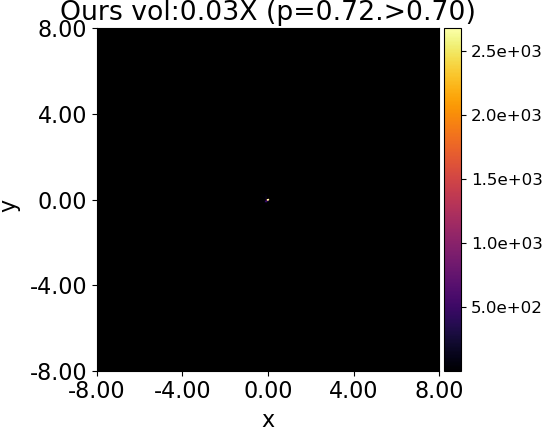} \hfill
\caption{Ours}
\end{subfigure}
\centering
\caption{Comparison of reachable set computation among different tools (Double integrator, t=7)\CAPRBT}
\label{fig:reach_e01_dint_007}
\end{figure}

\begin{figure}[!htbp]
\begin{subfigure}[b]{0.222\textwidth}
\includegraphics[width=1.0\textwidth]{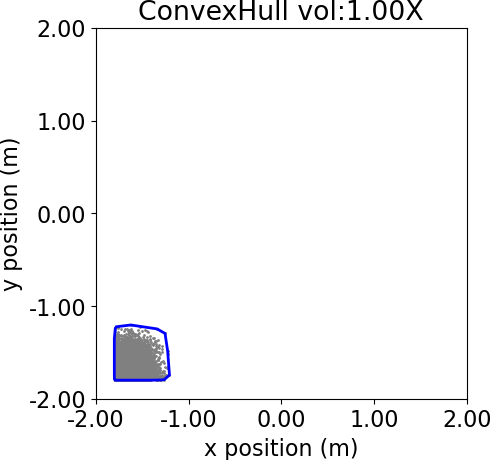} \hfill
\caption{Convex Hull}
\end{subfigure}
\begin{subfigure}[b]{0.222\textwidth}
\includegraphics[width=1.0\textwidth]{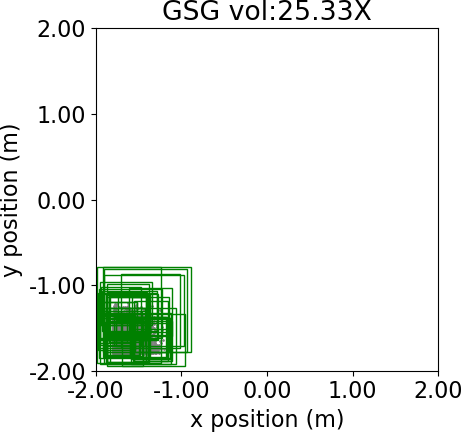} \hfill
\caption{GSG}
\end{subfigure}
\begin{subfigure}[b]{0.222\textwidth}
\includegraphics[width=1.0\textwidth]{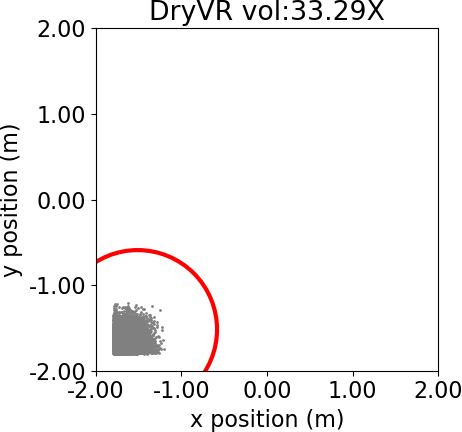} \hfill
\caption{DryVR}
\end{subfigure}
\begin{subfigure}[b]{0.253\textwidth}
\includegraphics[width=1.0\textwidth]{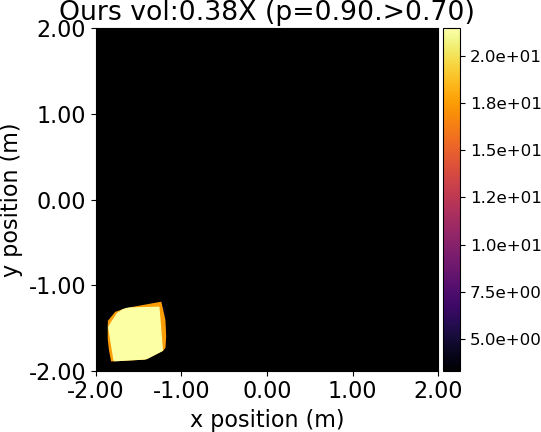} \hfill
\caption{Ours}
\end{subfigure}
\centering
\caption{Comparison of reachable set computation among different tools (Ground robot navigation, t=0)\CAPRBT}
\label{fig:reach_e04_robot_000}
\end{figure}

\begin{figure}[!htbp]
\begin{subfigure}[b]{0.222\textwidth}
\includegraphics[width=1.0\textwidth]{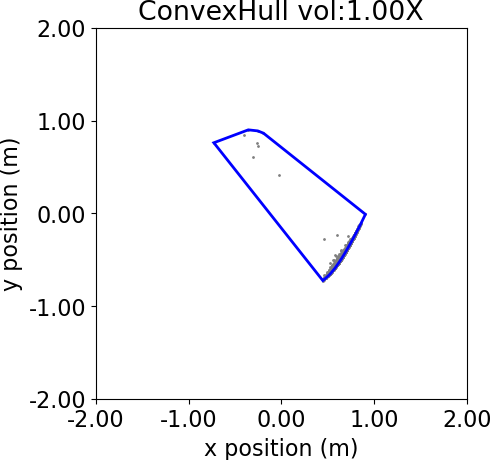} \hfill
\caption{Convex Hull}
\end{subfigure}
\begin{subfigure}[b]{0.222\textwidth}
\includegraphics[width=1.0\textwidth]{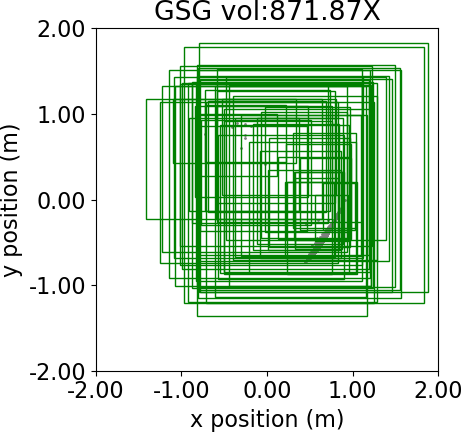} \hfill
\caption{GSG}
\end{subfigure}
\begin{subfigure}[b]{0.222\textwidth}
\includegraphics[width=1.0\textwidth]{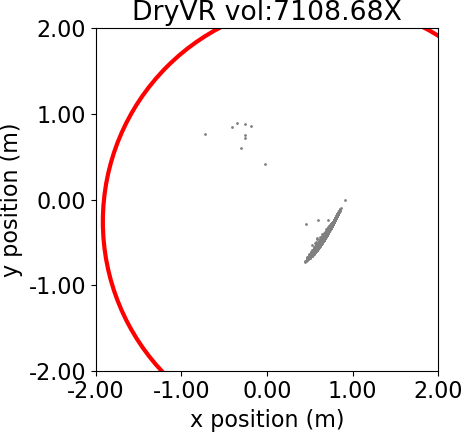} \hfill
\caption{DryVR}
\end{subfigure}
\begin{subfigure}[b]{0.253\textwidth}
\includegraphics[width=1.0\textwidth]{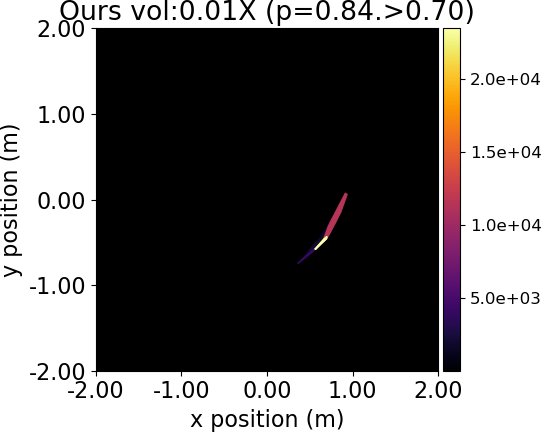} \hfill
\caption{Ours}
\end{subfigure}
\centering
\caption{Comparison of reachable set computation among different tools (Ground robot navigation, t=20)\CAPRBT}
\label{fig:reach_e04_robot_020}
\end{figure}

\begin{figure}[!htbp]
\begin{subfigure}[b]{0.222\textwidth}
\includegraphics[width=1.0\textwidth]{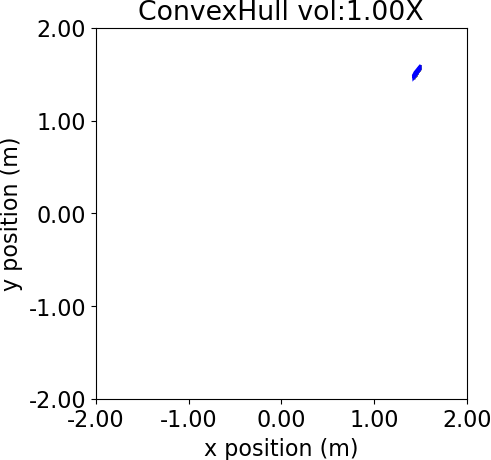} \hfill
\caption{Convex Hull}
\end{subfigure}
\begin{subfigure}[b]{0.222\textwidth}
\includegraphics[width=1.0\textwidth]{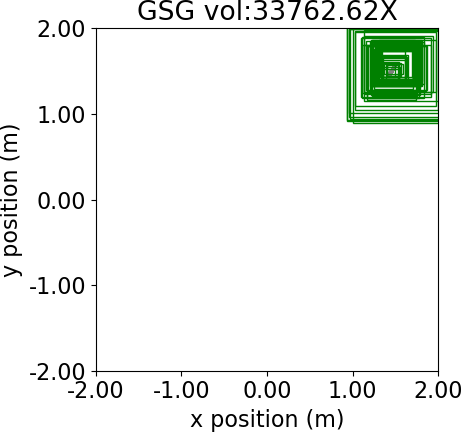} \hfill
\caption{GSG}
\end{subfigure}
\begin{subfigure}[b]{0.222\textwidth}
\includegraphics[width=1.0\textwidth]{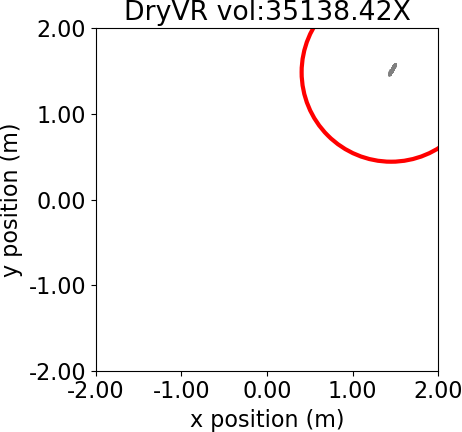} \hfill
\caption{DryVR}
\end{subfigure}
\begin{subfigure}[b]{0.253\textwidth}
\includegraphics[width=1.0\textwidth]{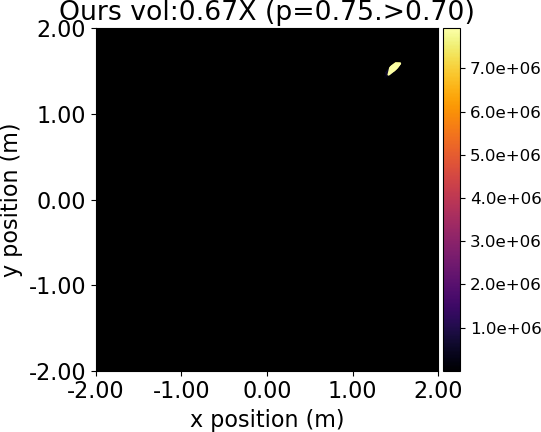} \hfill
\caption{Ours}
\end{subfigure}
\centering
\caption{Comparison of reachable set computation among different tools (Ground robot navigation, t=40)\CAPRBT}
\label{fig:reach_e04_robot_040}
\end{figure}

\begin{figure}[!htbp]
\begin{subfigure}[b]{0.222\textwidth}
\includegraphics[width=1.0\textwidth]{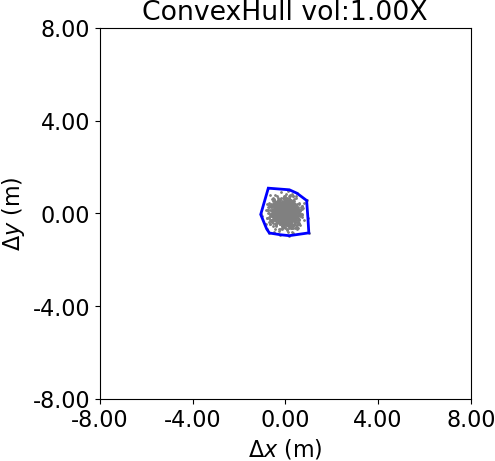} \hfill
\caption{Convex Hull}
\end{subfigure}
\begin{subfigure}[b]{0.222\textwidth}
\includegraphics[width=1.0\textwidth]{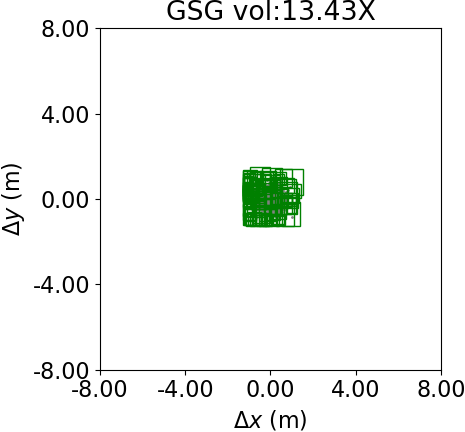} \hfill
\caption{GSG}
\end{subfigure}
\begin{subfigure}[b]{0.222\textwidth}
\includegraphics[width=1.0\textwidth]{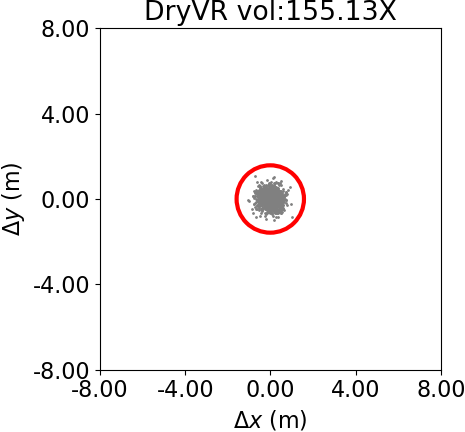} \hfill
\caption{DryVR}
\end{subfigure}
\begin{subfigure}[b]{0.253\textwidth}
\includegraphics[width=1.0\textwidth]{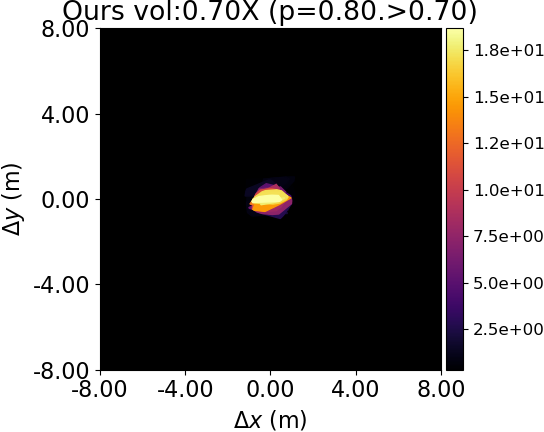} \hfill
\caption{Ours}
\end{subfigure}
\centering
\caption{Comparison of reachable set computation among different tools (FACTEST car model, t=0)\CAPRBT}
\label{fig:reach_e05_car_000}
\end{figure}

\begin{figure}[!htbp]
\begin{subfigure}[b]{0.222\textwidth}
\includegraphics[width=1.0\textwidth]{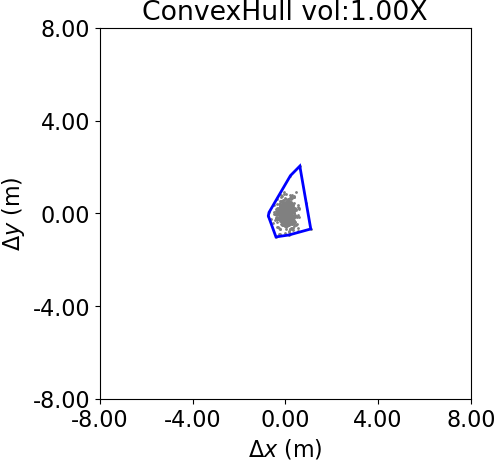} \hfill
\caption{Convex Hull}
\end{subfigure}
\begin{subfigure}[b]{0.222\textwidth}
\includegraphics[width=1.0\textwidth]{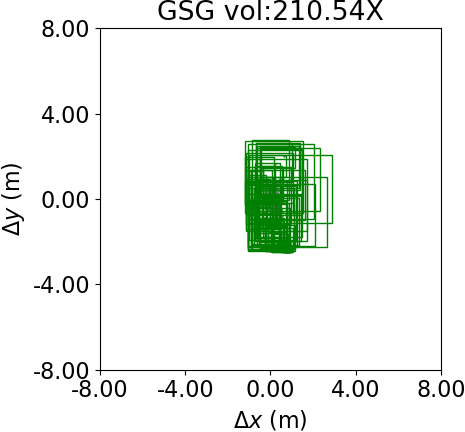} \hfill
\caption{GSG}
\end{subfigure}
\begin{subfigure}[b]{0.222\textwidth}
\includegraphics[width=1.0\textwidth]{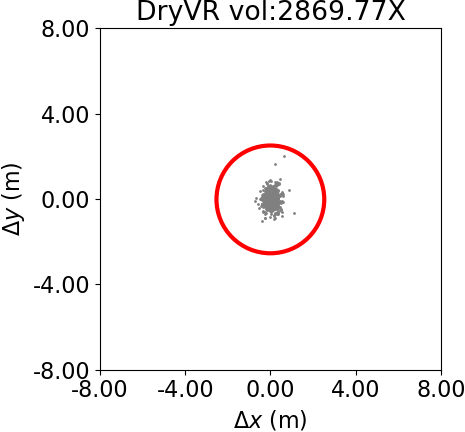} \hfill
\caption{DryVR}
\end{subfigure}
\begin{subfigure}[b]{0.253\textwidth}
\includegraphics[width=1.0\textwidth]{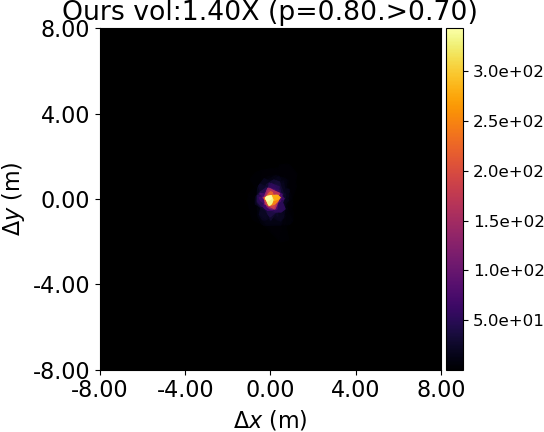} \hfill
\caption{Ours}
\end{subfigure}
\centering
\caption{Comparison of reachable set computation among different tools (FACTEST car model, t=20)\CAPRBT}
\label{fig:reach_e05_car_020}
\end{figure}

\begin{figure}[!htbp]
\begin{subfigure}[b]{0.222\textwidth}
\includegraphics[width=1.0\textwidth]{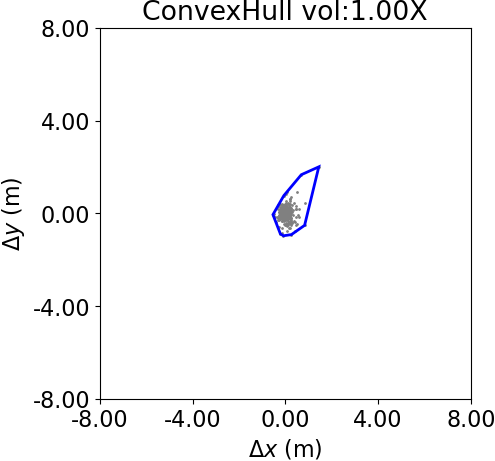} \hfill
\caption{Convex Hull}
\end{subfigure}
\begin{subfigure}[b]{0.222\textwidth}
\includegraphics[width=1.0\textwidth]{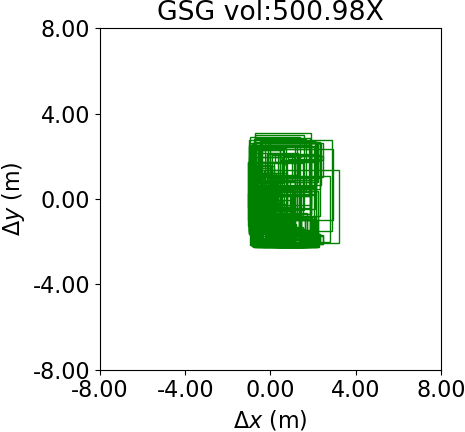} \hfill
\caption{GSG}
\end{subfigure}
\begin{subfigure}[b]{0.222\textwidth}
\includegraphics[width=1.0\textwidth]{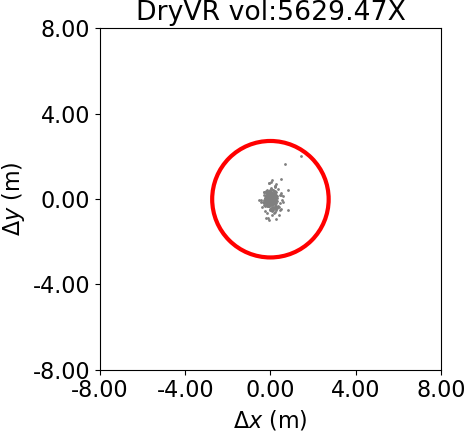} \hfill
\caption{DryVR}
\end{subfigure}
\begin{subfigure}[b]{0.253\textwidth}
\includegraphics[width=1.0\textwidth]{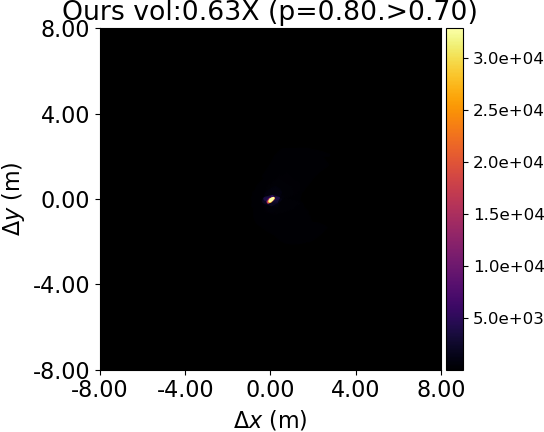} \hfill
\caption{Ours}
\end{subfigure}
\centering
\caption{Comparison of reachable set computation among different tools (FACTEST car model, t=49)\CAPRBT}
\label{fig:reach_e05_car_049}
\end{figure}

\newpage
\DRAFT{\section{Comparison between histogram-based and Liouville-based approaches}}
\label{sec:g}

\DRAFT{The advantage of learning density distribution by solving Liouville ODE is that it requires less training samples than histogram-based approaches, hence has the potential to generalize to high-dimension cases. To show its advantages in training efficiency and testing accuracy, we compare the histogram-based approach and Liouville-based approach's density estimation for the following system, under different number of training samples. To make sure we can compare to the ``groundtruth" density, we manually design the system such that the state density distribution at each time step has a closed form solution.}

\DRAFT{Consider a 1-d system: $\dot{x}=-x^2$ with initial states ranged from $[0, 1]$. Under uniformly distributed initialization, the system dynamics $x(t)$ and density distribution $\beta(x,t)$ (here $\beta(x,t)$ denotes the density at time $t$ at location $x$) can be directly written out in the closed form:}
\begin{equation}
\begin{cases}
    x(t)=1/(C+t)\\
    \beta(x, t)=1/(1-x\cdot t)^2
\end{cases}
\end{equation}
where the parameter $C$ can be derived from the initial condition $x(0)=x_0=1/C$.

\begin{figure}[!htbp]
\centering
\includegraphics[width=0.6\textwidth]{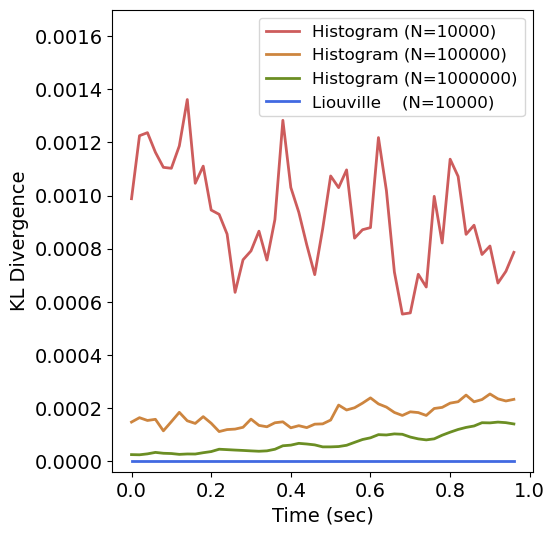} \hfill
\caption{KL Divergence (comparing to groundtruth state density) for histogram-based approach and Liouville-based approach in different numbers of training samples. The groundtruth state density is computed analytically in closed form. We use 10000, 100000, 1000000 training samples for the histogram-based approach and use only 10000 training samples for the Liouville-based approach. The KL divergence is computed on a separate testing set (200 samples). Comparing to the histogram-based approach, the Liouville-based approach can achieve a smaller KL divergence while using 0.01X number of training samples.}
\label{fig:rbt_1d_sys}
\end{figure}

\DRAFT{We then use histogram-based and Liouville-based approach to estimate the state density for this system. We uniformly sample initial states and generate 1000000 trajectories using ODE45 solver. We use 10000,  100000 and 1000000 training samples for the histogram-based approach and use only 10000 training samples for the Liouville-based approach, then we estimate the density on a separate testing set of trajectories using nearest neighbor interpolation. At each time step, we measure the estimation accuracy on the test set by computing the KL divergence to the groundtruth density. As shown in Fig.~\ref{fig:rbt_1d_sys}, histogram-based approach needs lots of samples to accurately approximate a good distribution (the KL divergence converges to zero at each time step as the number of samples increases), where our approach can learn the density distribution with the lowest KL divergence using just 0.01X of the sampled trajectories. This shows the advantage of solving Liouville ODE to estimate the state density.} 

\DRAFT{\section{Comparison with state-of-the-art worst-case reachability approaches}}
\label{sec:h}

\DRAFT{We compare our approach with three state-of-the-art worst-case reachability methods: Sherlock~\citep{dutta2017output}, Verisig~\citep{ivanov2019verisig} and ReachNN~\citep{fan2020reachnn}. We use the official implementation of Verisig and ReachNN which focus on reachable set computation for neural network control systems (NNCS), and use the re-implementation of Sherlock from \citep{liu2019algorithms}, which is for neural network verification.}

\DRAFT{To make a fair comparison, we set a timeout limit of six hours for all approaches. Among all the four datasets that our method has computed, Sherlock can solve for the reachable sets for the datasets ``Double integrator", ``Ground robot navigation" and ``FACTEST car tracking system", and Verisig and ReachNN can only calculate for the ``Double integrator" dataset - Verisig encounters numerical issue on this dataset at first due to the large initial set, and we have to divide the initial set to smaller sets and run the program multiple times in parallel to compute for the reachable sets. Similar in Sec.~\ref{sec:reachable_set_analysis}, we measure the reachable sets by computing the volume of the reachable sets relative to the volume of the convex hull of the sampled points.}

\DRAFT{We use different networks when doing reachability analysis, because all those methods have different requirements for the analyzed system: }
\begin{enumerate}[label=(\alph*)]
    \item \DRAFT{ The RPM used in our approach is doing reachability analysis for ReLU-based NNs. For the “Double Integrator” system, the controller is another ReLU-based NN that has a clip function at the output (to rectify the control output between $[-1,1]$) }
    \item \DRAFT{The Sherlock approach we used in~\cite{liu2019algorithms} can only work with ReLU-based NN (not  NNCS). Thus we used the same NN used in (a) and conducted the experiments. Since we only compute for the reachable set, we just collect the flow map estimation $\Phi_{\omega}$ part of this NN (i.e. we did not need to use the density estimator part of the NN).}
    \item \DRAFT{Verisig can only work with a Neural Network Controlled System (NNCS) with Sigmoid/Tanh-based NN controllers. Thus we re-trained a Tanh-based NN controller (using the same number of hidden layers and hidden units) to reproduce the output of the original controller in (a) and use this new controller to do reachability analysis. We verified that the L2 error between the Tanh-based NN controller and the original controller is less than 0.001 on the testing set, and we also inspected the trajectories generated using these two controllers and cannot find a substantial difference.
    }
    \item \DRAFT{ReachNN can work with NNCS that has Sigmoid/Tanh/ReLU-based NN controllers. However, it cannot directly process the controller we had in (a) because the controller in (a) has a clip function at the output to rectify the control output between $[-1,1]$. Therefore, we trained another ReLU-based NN controller that does not have that clip function to reproduce the output of the original controller in (a). We use this newly trained controller to do reachability analysis in ReachNN.}
\end{enumerate}

\begin{figure}[!htbp]
\begin{subfigure}[b]{0.23\textwidth}  
\includegraphics[width=1.0\textwidth]{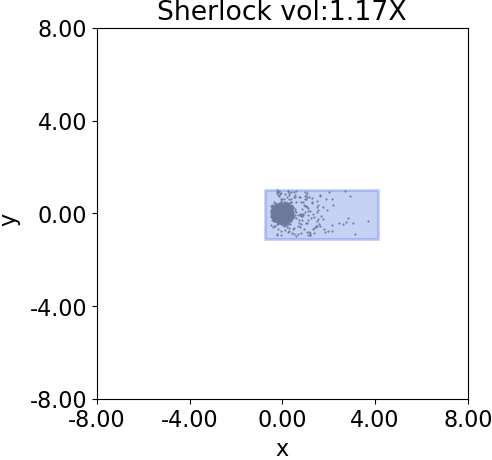} \hfill
\caption{Sherlock}
\end{subfigure}
\begin{subfigure}[b]{0.23\textwidth}
\includegraphics[width=1.0\textwidth]{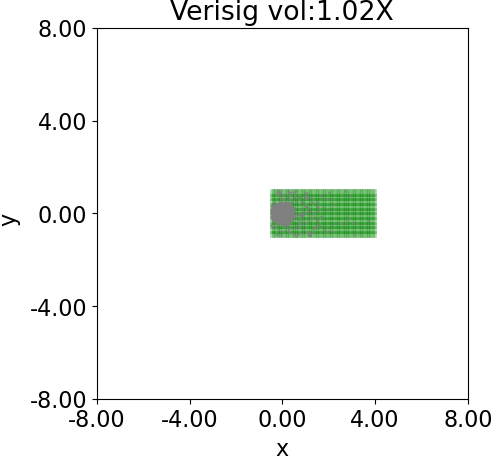} \hfill
\caption{Verisig}
\end{subfigure}
\begin{subfigure}[b]{0.23\textwidth}
\includegraphics[width=1.0\textwidth]{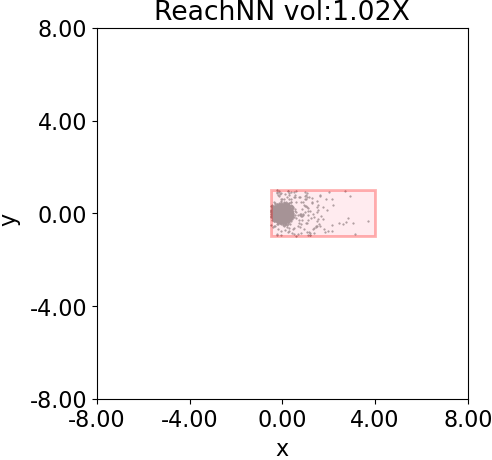} \hfill
\caption{ReachNN}
\end{subfigure}
\begin{subfigure}[b]{0.27\textwidth}  
\includegraphics[width=1.0\textwidth]{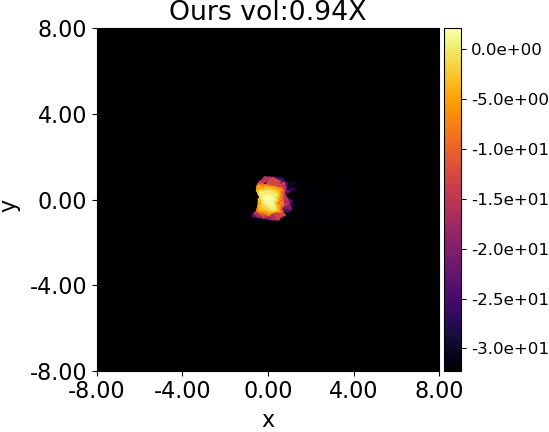} \hfill
\caption{Ours}
\end{subfigure}
\centering
\caption{Comparison of reachable sets (Double integrator, t=0)}
\label{fig:rbt_dint_000}
\end{figure}

\DRAFT{As shown in Fig.~\ref{fig:rbt_dint_000} $\sim$ Fig.~\ref{fig:rbt_dint_009}, in the ``Double integrator" experiment, all of the three worst-case reachability analysis methods can only over-approximate the reachable sets of the system, with the reachable volume increasing over time. The approximation error for Versig and ReachNN will severely accumulate, hence the corresponding reachable sets gradually occupy the whole figure (where the growths is 32.24X for Verisig and 67.96X for ReachNN respectively), whereas our approach estimated reachable sets have volume less than the convex hull volume, and can reflect the convergence of the majority of the system states owing to the ability to predict the state density. For higher dimension benchmarks like ``Ground robot navigation" and ``FACTEST car tracking system" (as shown in Fig.~\ref{fig:rbt_sherlock_robot}  $\sim$ Fig.~\ref{fig:rbt_rpm_car}), only our approach and Sherlock are able to compute the reachable set under the timeout limit. Due to the high dimensionality, Sherlock's estimated volume grows dramatically over time (16.51X for the ``Ground robot navigation", and 42.60X for the ``FACTEST car model"), while our approach still gives more compact reachable sets. These observations illustrate the advantages of our approach in precisely estimating the system reachable sets as well as the state density distribution. One advantage of Sherlock over ours is that it can also solve for other benchmarks listed in Table.~\ref{table:1}, where our approach cannot solve due to the numerical issues in RPM. Another limitation is that our approach only solves for NN with ReLU activations, which is again a restriction inherited from RPM. We believe combining our learning framework with a more advanced exact reachability tool will resolve this issue in the future.}

\begin{figure}[!htbp]
\begin{subfigure}[b]{0.23\textwidth}  
\includegraphics[width=1.0\textwidth]{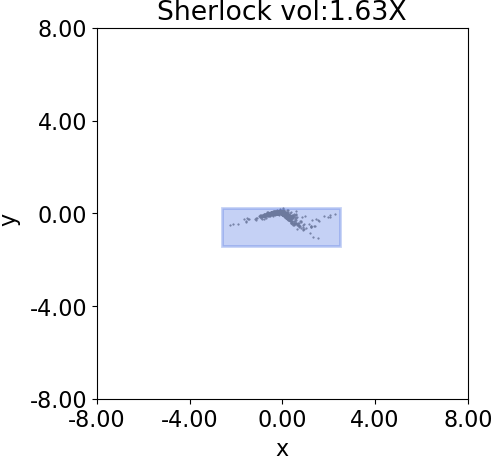} \hfill
\caption{Sherlock}
\end{subfigure}
\begin{subfigure}[b]{0.23\textwidth}
\includegraphics[width=1.0\textwidth]{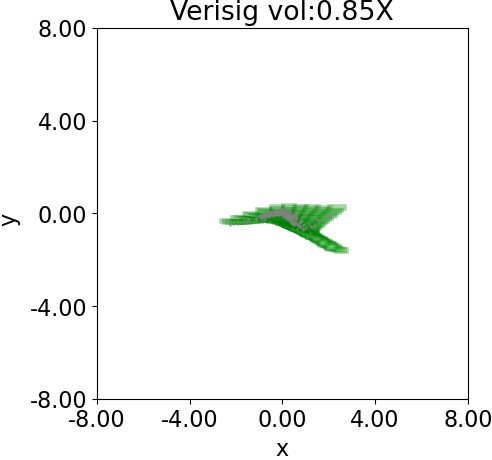} \hfill
\caption{Verisig}
\end{subfigure}
\begin{subfigure}[b]{0.23\textwidth}
\includegraphics[width=1.0\textwidth]{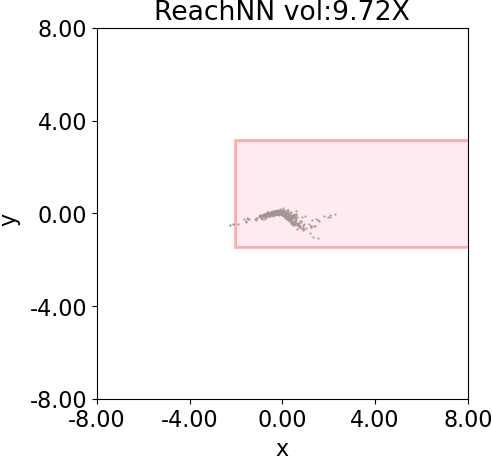} \hfill
\caption{ReachNN}
\end{subfigure}
\begin{subfigure}[b]{0.27\textwidth}  
\includegraphics[width=1.0\textwidth]{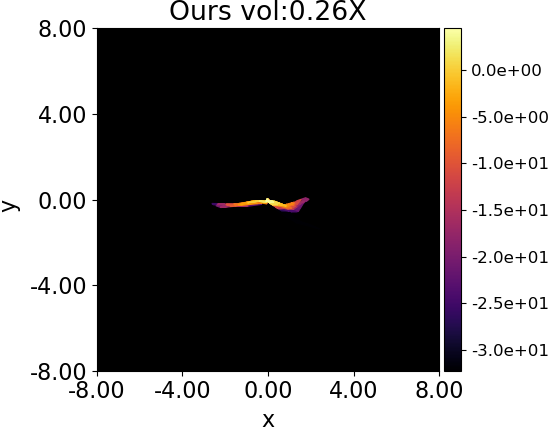} \hfill
\caption{Ours}
\end{subfigure}
\centering
\caption{Comparison of reachable sets (Double integrator, t=3)}
\label{fig:rbt_dint_003}
\end{figure}

\begin{figure}[!htbp]
\begin{subfigure}[b]{0.23\textwidth}  
\includegraphics[width=1.0\textwidth]{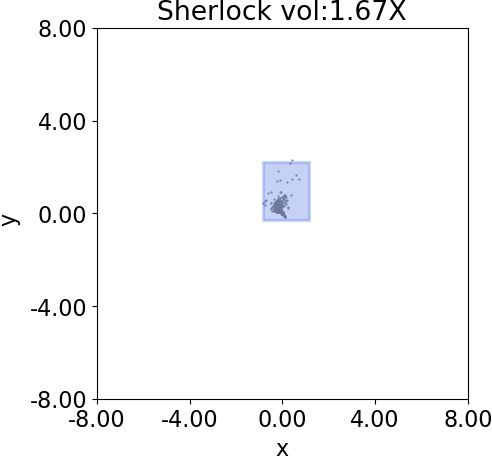} \hfill
\caption{Sherlock}
\end{subfigure}
\begin{subfigure}[b]{0.23\textwidth}
\includegraphics[width=1.0\textwidth]{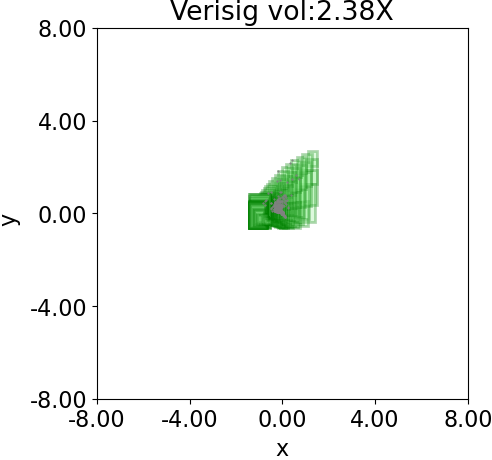} \hfill
\caption{Verisig}
\end{subfigure}
\begin{subfigure}[b]{0.23\textwidth}
\includegraphics[width=1.0\textwidth]{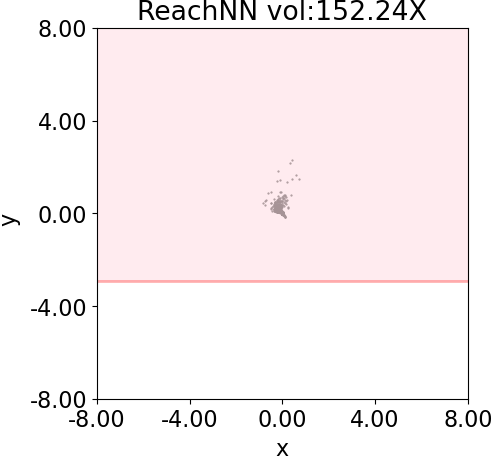} \hfill
\caption{ReachNN}
\end{subfigure}
\begin{subfigure}[b]{0.27\textwidth}  
\includegraphics[width=1.0\textwidth]{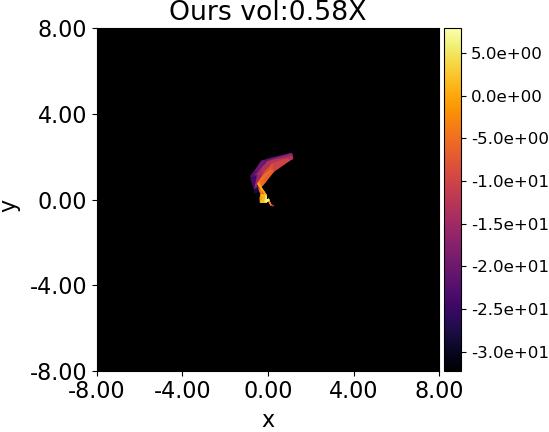} \hfill
\caption{Ours}
\end{subfigure}
\centering
\caption{Comparison of reachable sets (Double integrator, t=7)}
\label{fig:rbt_dint_007}
\end{figure}

\begin{figure}[!htbp]
\begin{subfigure}[b]{0.23\textwidth}  
\includegraphics[width=1.0\textwidth]{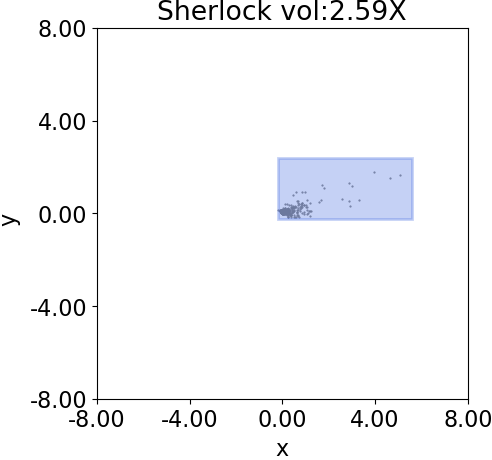} \hfill
\caption{Sherlock}
\end{subfigure}
\begin{subfigure}[b]{0.23\textwidth}
\includegraphics[width=1.0\textwidth]{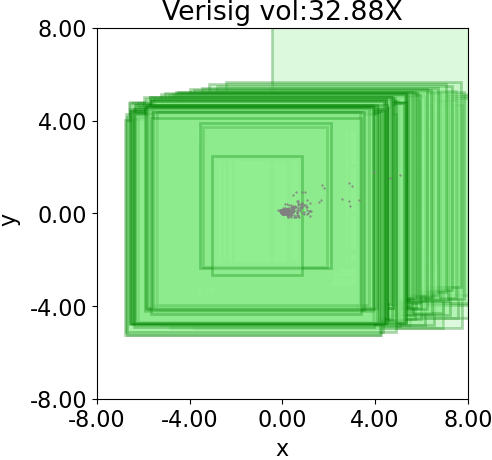} \hfill
\caption{Verisig}
\end{subfigure}
\begin{subfigure}[b]{0.23\textwidth}
\includegraphics[width=1.0\textwidth]{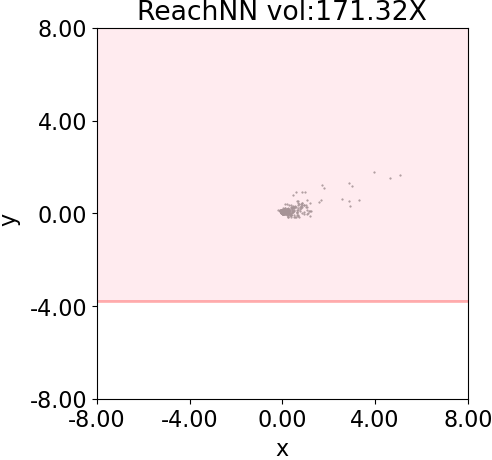} \hfill
\caption{ReachNN}
\end{subfigure}
\begin{subfigure}[b]{0.27\textwidth}  
\includegraphics[width=1.0\textwidth]{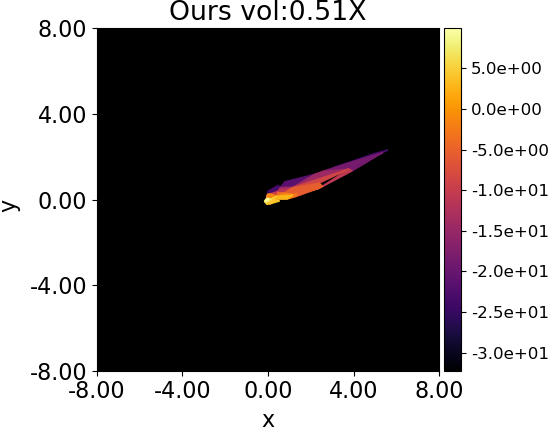} \hfill
\caption{Ours}
\end{subfigure}
\centering
\caption{Comparison of reachable sets (Double integrator, t=9)}
\label{fig:rbt_dint_009}
\end{figure}

\begin{figure}[!htbp]
\begin{subfigure}[b]{0.210\textwidth}
\includegraphics[width=1.0\textwidth]{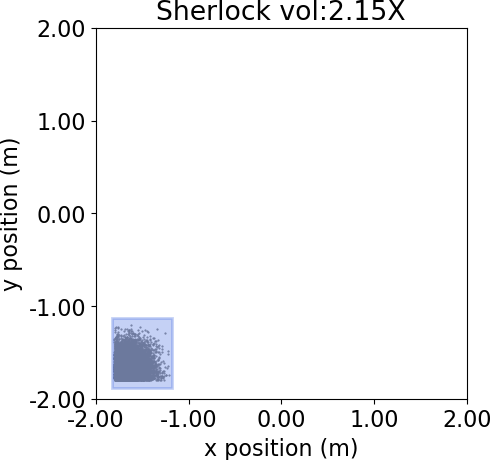} \hfill
\caption{Sherlock, t=0}
\end{subfigure}
\begin{subfigure}[b]{0.210\textwidth}
\includegraphics[width=1.0\textwidth]{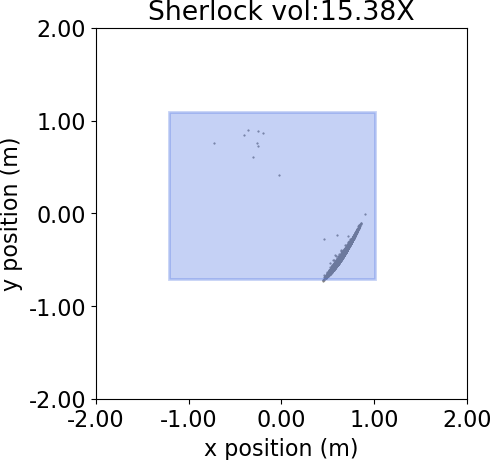} \hfill
\caption{Sherlock, t=20}
\end{subfigure}
\begin{subfigure}[b]{0.210\textwidth}
\includegraphics[width=1.0\textwidth]{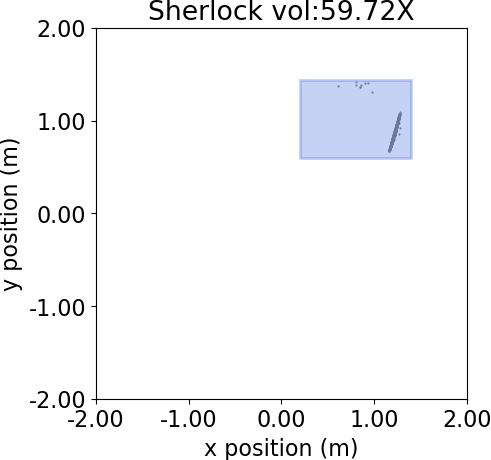} \hfill
\caption{Sherlock, t=30}
\end{subfigure}
\begin{subfigure}[b]{0.210\textwidth}
\includegraphics[width=1.0\textwidth]{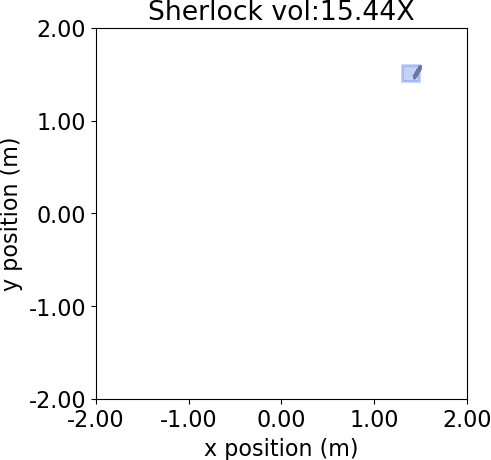} \hfill
\caption{Sherlock, t=40}
\end{subfigure}
\centering
\caption{Sherlock results (Ground robot navigation, t=0, 20, 30, 40)}
\label{fig:rbt_sherlock_robot}
\end{figure}


\begin{figure}[!htbp]
\begin{subfigure}[b]{0.220\textwidth}
\includegraphics[width=1.0\textwidth]{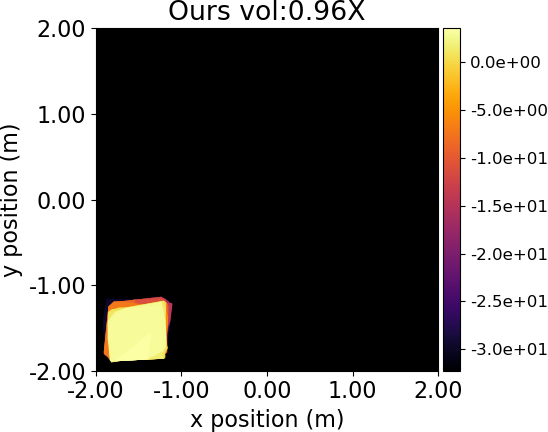} \hfill
\caption{Ours, t=0}
\end{subfigure}
\begin{subfigure}[b]{0.220\textwidth}
\includegraphics[width=1.0\textwidth]{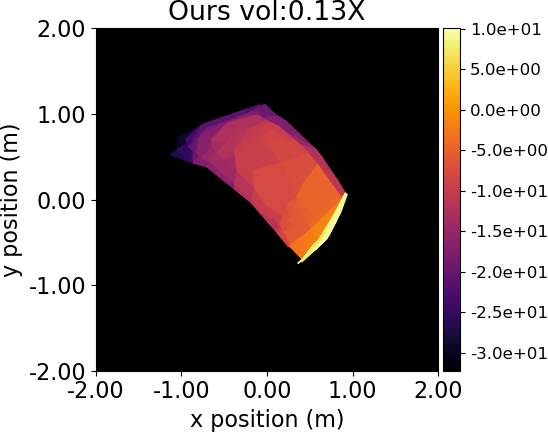} \hfill
\caption{Ours, t=20}
\end{subfigure}
\begin{subfigure}[b]{0.220\textwidth}
\includegraphics[width=1.0\textwidth]{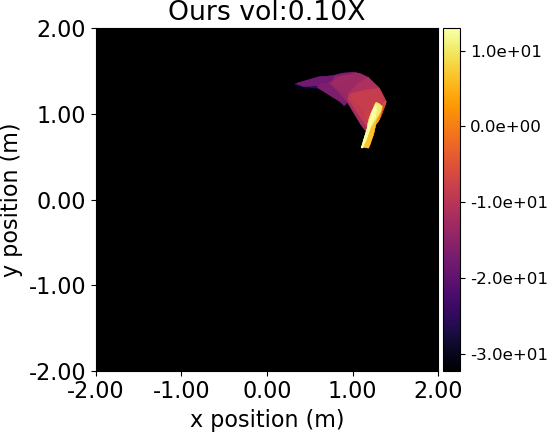} \hfill
\caption{Ours, t=30}
\end{subfigure}
\begin{subfigure}[b]{0.220\textwidth}
\includegraphics[width=1.0\textwidth]{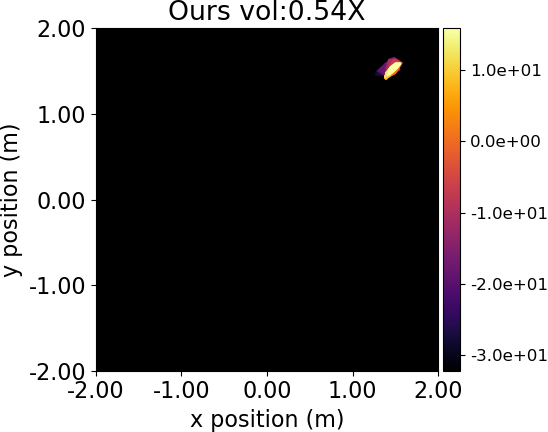} \hfill
\caption{Ours, t=40}
\end{subfigure}
\centering
\caption{Our results (Ground robot navigation, t=0, 20, 30, 40)}
\label{fig:rbt_rpm_robot}
\end{figure}

\begin{figure}[!htbp]
\begin{subfigure}[b]{0.210\textwidth}
\includegraphics[width=1.0\textwidth]{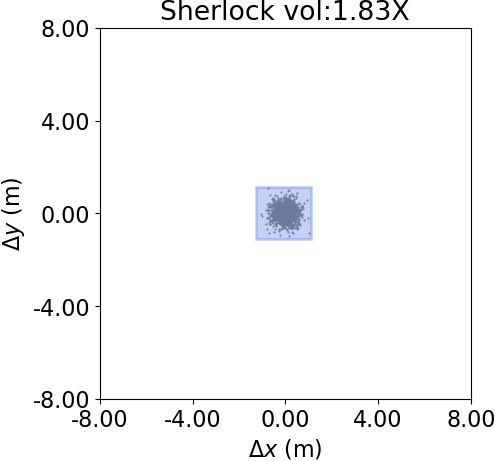} \hfill
\caption{Sherlock, t=0}
\end{subfigure}
\begin{subfigure}[b]{0.210\textwidth}
\includegraphics[width=1.0\textwidth]{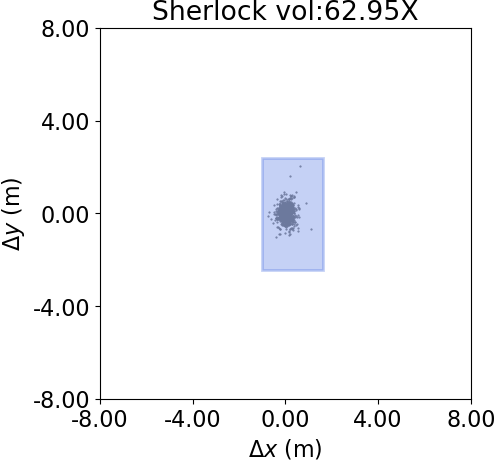} \hfill
\caption{Sherlock, t=20}
\end{subfigure}
\begin{subfigure}[b]{0.210\textwidth}
\includegraphics[width=1.0\textwidth]{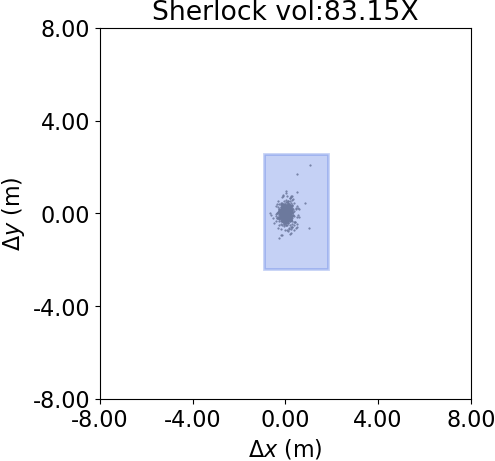} \hfill
\caption{Sherlock, t=30}
\end{subfigure}
\begin{subfigure}[b]{0.210\textwidth}
\includegraphics[width=1.0\textwidth]{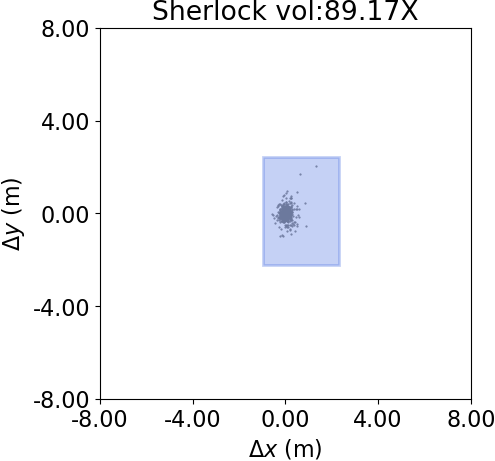} \hfill
\caption{Sherlock, t=40}
\end{subfigure}
\centering
\caption{Sherlock results (FACTEST car model, t=0, 20, 30, 40)}
\label{fig:rbt_sherlock_car}
\end{figure}


\begin{figure}[!htbp]
\begin{subfigure}[b]{0.220\textwidth}
\includegraphics[width=1.0\textwidth]{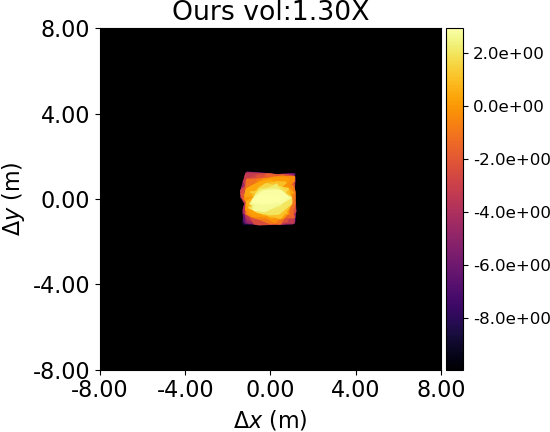} \hfill
\caption{Ours, t=0}
\end{subfigure}
\begin{subfigure}[b]{0.220\textwidth}
\includegraphics[width=1.0\textwidth]{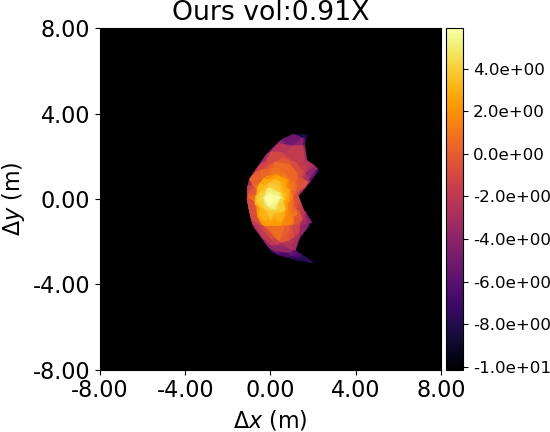} \hfill
\caption{Ours, t=20}
\end{subfigure}
\begin{subfigure}[b]{0.220\textwidth}
\includegraphics[width=1.0\textwidth]{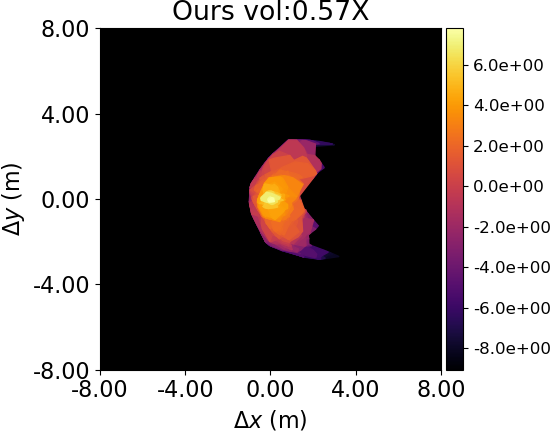} \hfill
\caption{Ours, t=30}
\end{subfigure}
\begin{subfigure}[b]{0.220\textwidth}
\includegraphics[width=1.0\textwidth]{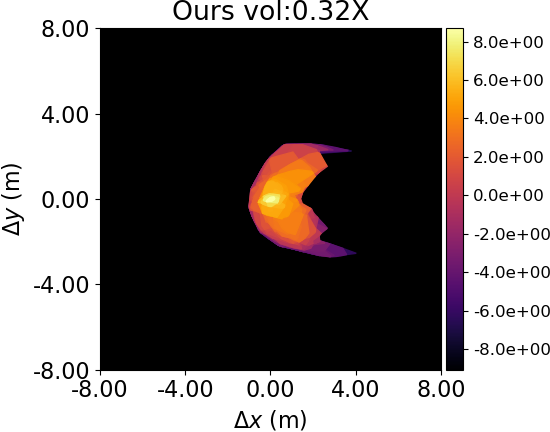} \hfill
\caption{Ours, t=40}
\end{subfigure}
\centering
\caption{Our results (FACTEST car model, t=0, 20, 30, 40)}
\label{fig:rbt_rpm_car}
\end{figure}

\newpage
\end{document}